\documentclass[10pt,twocolumn,letterpaper]{article}

\usepackage[pagenumbers]{cvpr}              %

\usepackage{graphicx}
\usepackage{amsmath}
\usepackage{amssymb}
\usepackage{booktabs}

\usepackage{xr}
\makeatletter

\usepackage[dvipsnames]{xcolor}
\definecolor{cvprblue}{rgb}{0.21,0.49,0.74}
\usepackage[pagebackref,breaklinks,colorlinks,citecolor=cvprblue]{hyperref}

\usepackage[capitalize]{cleveref} 
\crefname{section}{Sec.}{Secs.}
\Crefname{section}{Section}{Sections}
\Crefname{table}{Table}{Tables}
\crefname{table}{Tab.}{Tabs.}

\def\paperID{*****} %
\def\confName{CVPR}
\def\confYear{2024}

\usepackage{glossaries}

\newacronym{ai}{AI}{Artificial Intelligence}
\newacronym{dl}{DL}{Deep Learning}
\newacronym{dnn}{DNN}{Deep Neural Network}
\newacronym{ood}{OOD}{Out Of Distribution}
\newacronym{lrp}{LRP}{Layer-wise Relevance Propagation}
\newacronym{xai}{XAI}{eXplainable Artificial Intelligence}
\newacronym{crp}{CRP}{Concept Relevance Propagation}
\newacronym{crv}{CRV}{Concept Relevance Vector}
\newacronym{cav}{CAV}{Concept Activation Vector}
\newacronym{pcrv}{PCRV}{Prototypical Concept Relevance Vector}
\newacronym{ml}{ML}{Machine Learning}
\newacronym{gmm}{GMM}{Gaussian Mixture Model}
\newacronym{auc}{AUC}{Area Under the Curve}
\newacronym{se}{SE}{Standard Error}
\newacronym{ours}{PCX}{Prototypical Concept-based Explanations}
\newcommand{\x}{{\mathbf{x}}\xspace}
\newcommand{\argmax}{{\text{argmax}}\xspace} 
\newcommand{\argmin}{{\text{argmin}}\xspace} 
\newcommand{\bt}[1]{\hspace{-0.11em}\mathbf{{#1}}\hspace{-0.11em}} %

\def\cavbold{\mathbf{u}}
\def\mean{\boldsymbol{\mu}}
\def\sig{\boldsymbol{\Sigma}}

\begin{document}

\title{Understanding the (Extra-)Ordinary: Validating Deep Model Decisions with Prototypical Concept-based Explanations}

\author{
Maximilian Dreyer$^{1}$,
Reduan Achtibat$^{1}$,
Wojciech Samek$^{1,2,3,\dagger}$,
Sebastian Lapuschkin$^{1,\dagger}$\\
$^1$ Fraunhofer Heinrich Hertz Institute,
$^2$ Technical University of Berlin, \\
$^3$ BIFOLD – Berlin Institute for the Foundations of Learning and Data\\
$^\dagger${\small corresponding authors:}
{\tt\small  \{wojciech.samek\,|\,sebastian.lapuschkin\}@hhi.fraunhofer.de}
}
\maketitle

\begin{abstract}

    Ensuring both transparency and safety is critical when deploying Deep Neural Networks (DNNs) in high-risk applications, such as medicine.
    The field of explainable AI (XAI) has proposed various methods to comprehend the decision-making processes of opaque DNNs.
    However,
    only few XAI methods are suitable of ensuring safety in practice as they heavily rely on repeated labor-intensive and possibly biased human assessment.
    In this work,
    we present a novel post-hoc concept-based XAI framework that conveys besides instance-wise (local) also class-wise (global) decision-making strategies via prototypes.
    What sets our approach apart is the combination of local and global strategies,
    enabling a clearer understanding of the \mbox{(dis-)similarities} in model decisions compared to the expected (prototypical) concept use,
    ultimately reducing the dependence on human long-term assessment.
    Quantifying the deviation from prototypical behavior
    not only allows to associate predictions with specific model sub-strategies but also to detect outlier behavior.
    As such, our approach constitutes an intuitive and explainable tool for model validation.
    We demonstrate the effectiveness of our approach in identifying out-of-distribution samples, spurious model behavior and data quality issues across three datasets (ImageNet, CUB-200, and CIFAR-10) utilizing VGG, ResNet, and EfficientNet architectures.
    Code is available at \url{https://github.com/maxdreyer/pcx}.

\end{abstract}

\section{Introduction}

    \glspl{dnn} showcase remarkable performance in tasks such as medical diagnosis~\cite{brinker2019deep} and autonomous driving~\cite{grigorescu2020survey}.
    The significance of understanding and validating \gls{ml} models becomes particularly pronounced in such safety-critical applications.
    Notably, \glspl{dnn} have been shown to learn shortcuts that stem from spurious data artifacts,
    such as watermarks \cite{lapuschkin2019unmasking}.
    They further provide unreliable predictions when faced with samples from unrelated data domains, commonly referred to as \gls{ood} samples.
    In both scenarios, model predictions may be rooted in incorrect reasoning,
    potentially leading to severe consequences when these models are deployed in real-world applications.

    The field of \gls{xai} has emerged to demystify the inner workings of black-box models and offer insights into their decision-making processes. 
    \gls{xai} methods generally fall into the categories of \emph{global} and \emph{local}.
    While global techniques study model behavior on the class-wise or dataset-wise level,
    local \gls{xai} renders explanations at the instance level,
    facilitating an understanding of input feature relevance for specific prediction outcomes.
    
    When deploying \gls{ml} models,
    the importance of transparency and safety for \emph{single} decisions often takes precedence,
    rendering global \gls{xai} insufficient on its own~\cite{jia2022role}.
    While local \gls{xai} techniques offer the potential for model prediction validation,
    they often rely heavily on human assessment to understand and interpret model behavior.
    The labor-intensive process of human assessment,
    coupled with the potential for human bias~\cite{du2019techniques},
    hinders the practical implementation of these systems in critical applications. 
    Therefore,
    a need for a more efficient and reliable means of understanding and validating safety of \gls{ml} models remains. 
    
    In this work,
    we address this challenge and propose a novel concept-based \gls{xai} framework named \gls{ours},
    \begin{figure*}[t]
                    \centering
                    \includegraphics[width=0.99\linewidth]{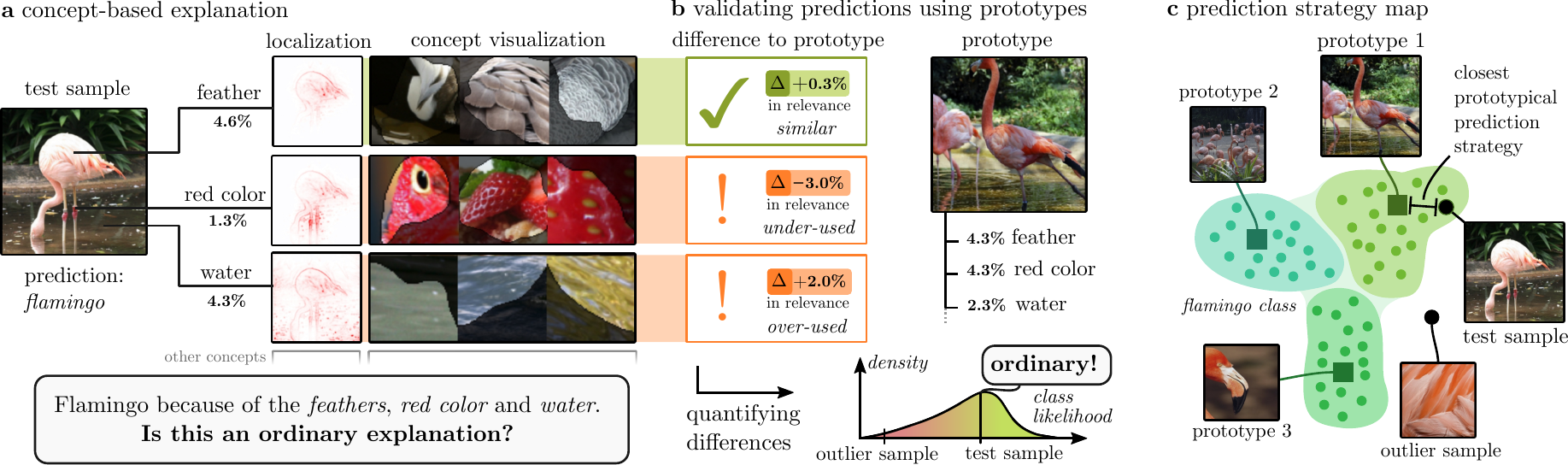}
                    \captionof{figure}{
                    Using the \gls{ours} framework:
                    By contrasting a prediction with the prototypical prediction strategy, the stakeholder can understand how (un-)ordinarly the model behaves. 
                    (\textbf{a}): A flamingo prediction is based on concepts like ``feather'', ``red color'' and ``water''.
                    While recent concept-based \gls{xai} methods provide relevance scores, localization heatmaps, and visualizations for each concept, it remains unclear whether such composition of used concepts is expected.
                    (\textbf{b}): Comparing against prototypes enables to understand to what extend concepts are similar (\eg, ``feather''), underused (\eg, ``red color''), or overused (\eg, ``water'').
                    These differences can be quantitatively measured to assess the degree of an outlier prediction.
                    (\textbf{c}): \gls{ours} allows to automatically identify outliers, or, alternatively, the closest prototypical prediction strategy.
                    Prototypes are hereby automatically discovered, summarizing the global model behavior in condensed fashion.
                    }
                    \label{fig:introduction:local_to_glocal}
    \end{figure*}
    that signals and reveals deviations from expected model behavior by
    providing meaningful and more objective explanations, 
    reducing the need for (possibly biased) human interpretation.
    Concretely,
    for any prediction,
    \gls{ours} communicates the differences and similarities to the expected model behavior via (automatically discovered) prototypes.
    Here,
    prototypes are representative predictions, that summarize the global model behavior in condensed fashion.
    To guarantee high interpretability throughout,
    we build upon the latest progress made in concept-based \gls{xai},
    offering explanations in terms of human-understandable concepts~\cite{achtibat2023attribution,fel2023craft},
    applicable to any \gls{dnn} architecture in a post-hoc manner.
    Notably,
    \gls{ours} can hereby not only highlight which concepts are used, but also which ones are \emph{not} used.
    For the flamingo in \cref{fig:introduction:local_to_glocal}a, \eg,
    we learn that the red color concept, typically present for the prototype in \cref{fig:introduction:local_to_glocal}b, is underrepresented.
    Quantifying differences in latent feature use allows for an objective means to identify typical prediction strategies, \eg, for positively validating predictions, issuing warnings otherwise.
    \paragraph{Contributions}

        This work introduces \gls{ours},
        an intrinsically explainable framework for validating \gls{dnn} predictions that combines both class-wise \emph{and} instance-wise prediction strategies.
        We show how \gls{ours} allows to
        \begin{enumerate}
            \item \emph{locally} study predictions on the concept-level by leveraging state-of-the-art concept-based \gls{xai} techniques. 
            \item \emph{globally} understand (dis-)similarities in prediction strategies within and across classes via prototypes.
            We further validate prototypes \wrt metrics such as faithfulness, stability and sparseness,
            demonstrating the superiority of concept relevance scores over activations.
            \item \emph{glocally} quantify and understand (un-)usual concept use by a model for individual predictions by comparing these to prototypes.
            We showcase \gls{ours} for detecting spurious model behavior, data quality issues and \gls{ood} samples.
        \end{enumerate}
\section{Related Work}

    We now present an overview of related work in concept-based \gls{xai}, prototypical explanations and \gls{ood} detection.
    
    \subsection{Concept-based Explanations}
    
    Contrary to traditional local feature attribution methods that investigate the importance of input-level features,
    concept-based \gls{xai} methods study the function (concept) of latent representations in a specific layer of a \gls{dnn}.
    Here,
    either single neurons~\cite{bau2017network}, directions, \ie, \glspl{cav}~\cite{kim2018interpretability}, or feature subspaces~\cite{vielhaben2023multi} are investigated.
    
    Early \gls{xai} works study how these concepts are used for \emph{global} decision making, \eg, the concepts most relevant for an output class~\cite{kim2018interpretability}.
    Recent works also generalize local feature attribution methods to compute importance scores of concepts for \emph{individual} predictions~\cite{achtibat2023attribution,fel2023craft},
    bringing concept-based explanations to the instance level.
    
    Whereas instance-wise concept-based explanations enable new levels of insight,
    they can be overwhelming and complex for a stakeholder to process, 
    as hundreds of concepts might exist and need to be studied for each instance~\cite{achtibat2023attribution}. Hence, other works~\cite{chan2020melody, fel2023holistic} illustrate the advantage of visualizing local decisions in a global embedding, as also shown in \cref{fig:introduction:local_to_glocal}c.
    With \gls{ours}, we extend this idea by introducing prototypes. This further reduces the need for human interpretation as it allows to compare individual (local) prediction strategies with prototypical (global) ones.
    
    \subsection{Prototypes for Explanations}
    
    Prototypes represent example predictions that summarize the global model behavior in condensed fashion,
    rendering them especially valuable for large and complex datasets.
    While there is a large group of works focusing on using prototypes for (robust) classification, \eg, ~\cite{liu2001evaluation,yang2018robust},
    few works use prototypes for \gls{xai}. 
    The works of ProtoAttend~\cite{arik2019protoattend} and \cite{chong2021toward}  increase interpretability of \glspl{dnn} by anchoring decisions on prototypical samples.
    Whereas both require modification of the \gls{dnn} architecture as well as additional training,
    \cite{chong2021toward} highlights similar features between test sample and prototype.
    In contrast,
    \gls{ours} is \emph{post-hoc} applicable and communicates similarities as well as \emph{differences} via human-understandable concepts.
    
    Prototypical parts are widely used in interpretable models~\cite{chen2019looks,nauta2021neural,nauta2023pip}.
    For example, ProtoPNet~\cite{chen2019looks}, one of the pioneering works, dissects images into prototypical parts for each class,
    subsequently classifying images by consolidating evidence from prototypes.
    It is important to note,
    that whereas these works study prototypical \emph{parts},
    we define a prototype as a \emph{prediction strategy}.

    \subsection{Out-of-Distribution Detection}

    A popular safety task in \gls{dnn} deployment is \gls{ood} detection,
    catching data samples that are not of the training distribution.
    One line of research focuses on the fact that \gls{ood} samples often result in uncertain predictions, rendering the softmax output values already highly informative~\cite{hendrycks2016baseline,liu2020energy}.
    Other works leverage, \eg, the latent activations and measure divergence from typical patterns \cite{lee2018simple}.
    Notably,
    \gls{ood} methods are not intrinsically interpretable.
    Hence, first works introduce post-hoc concept-based explanations for \gls{ood} detectors~\cite{choi2023concept,sevyeri2023transparent}.
    As \gls{ours} is rooted in concept-based \gls{xai},
    it inherently provides interpretable \gls{ood} detection.

\section{Methods}

    \gls{ours} is based on recent concept-based \gls{xai} techniques, which are introduced in \cref{sec:methods:concept-based-xai}.
    Using concept-based explanations,
    we define and compute prototypical explanations, 
    discussed in \cref{sec:methods:prototypes}, to which we then compare individual predictions as described in \cref{sec:methods:dissimilarities}.
    
    \subsection{Concept-based Explanations}
    \label{sec:methods:concept-based-xai}
        A model’s prediction is the result of successive layer-wise feature operations, where the intermediate latent features of each layer are described by the activations of its neurons.
        Given a sample $\x$ of dataset $\mathcal{X}$, 
        the latent activations $\mathbf{a}(\x) \in \mathbb{R}^n$ in a specific layer with $n$ neurons can be viewed as a point in a vector space (activation space) that is spanned by $n$ canonical basis vectors (one for each neuron).  
        
        Then,
        we can assign a concept $c$ to each neuron, or more generally, also to a superposition of neurons describing a direction in latent space
        via a \gls{cav} $\cavbold^c \in \mathbb{R}^n$.
        How the latent activations $\mathbf{a}(\x)$ are decomposed into a linear combination of $m$ (chosen) \glspl{cav} summarized by $\mathbf{U} = (\cavbold^1, \dots, \cavbold^m) \in \mathbb{R}^{n\times m}$
        is described by the transformation
        \begin{equation}
        \label{eq:methods:act2concept}
            \mathbf{a}(\x) = \mathbf{U} \boldsymbol \nu^{\text{act}}(\x)
        \end{equation}
        from concept space to activation space.
        Here, 
        $\boldsymbol \nu^{\text{act}}(\x) \in \mathbb{R}^m$
        summarizes the activation (contribution) of each of the $m$ concepts.
        Depending on the choice of the set of \glspl{cav},
        the decomposition might only be approximated, \eg, when using non-negative matrix factorization \cite{fel2023holistic}.
        For simplicity and to ensure exact reconstruction as in \cref{eq:methods:act2concept},
        we study the concepts of individual neurons in this work, \ie, choose $\cavbold^c = \mathbf{e}^c$ with canonical basis vectors $\mathbf{e}^c_i = \delta_{ci}$ with Kronecker delta $\delta$ leading to a direct mapping $\boldsymbol \nu^{\text{act}}(\x) = \mathbf{a}(\x)$.

        \paragraph{Concept Relevance Scores}
        In order to attain an understanding of how concepts are \emph{used} for individual samples $\x$ and prediction outcomes $y_k$, 
        we require relevance scores $ \nu_c^{\text{rel}}(\x|y_k)$ of each concept $c$.
        Concept relevance scores can be computed using various established feature attribution methods such as Input$\times$Gradient~\cite{shrikumar2017learning} or \gls{lrp}~\cite{bach2015pixel} (see~\cite{fel2023holistic} for an overview). 
        When studying each neuron's concept,
        concept relevances $\nu_c^{\text{rel}}(\x|y_k)$ are directly given by applying a feature attribution method and aggregating the relevances in the latent space instead of the input space. 

        \paragraph{Concept Localizations}
        Furthermore,
        we localize individual concepts in the input via heatmaps
        as shown in \cref{fig:introduction:local_to_glocal}a.
        Specifically,
        we leverage the CRP framework~\cite{achtibat2023attribution} that enables concept-specific heatmaps by restricting the backward pass of feature attribution methods (with \gls{lrp} by default).
        
        \paragraph{Concept Visualizations}
        Several works have proposed techniques to visualize concepts of latent representations \cite{nguyen2019understanding}.
        We adhere to the recently proposed Relevance Maximization approach~\cite{achtibat2023attribution}.
        This technique explains concepts by exemplifying them, selecting reference samples that most accurately represent the functionality of a neuron.
        These reference samples highlight the input components most \emph{relevant} to a specific concept, as shown in \cref{fig:introduction:local_to_glocal}a.
    \subsection{Finding Prototypes}
    \label{sec:methods:prototypes}
        During training,
        a model learns to extract and use features from the input data to fulfill its training task.
        If we are to collect the \emph{presence} of such features (given by latent activations),
        we can measure feature distributions that are characteristic for specific classes.
        The works of \cite{lee2018simple,yuksekgonul2023beyond} model such distribution via multivariate Gaussian distributions.
        
\begin{figure}
            \centering
                \includegraphics[width=0.80\linewidth]{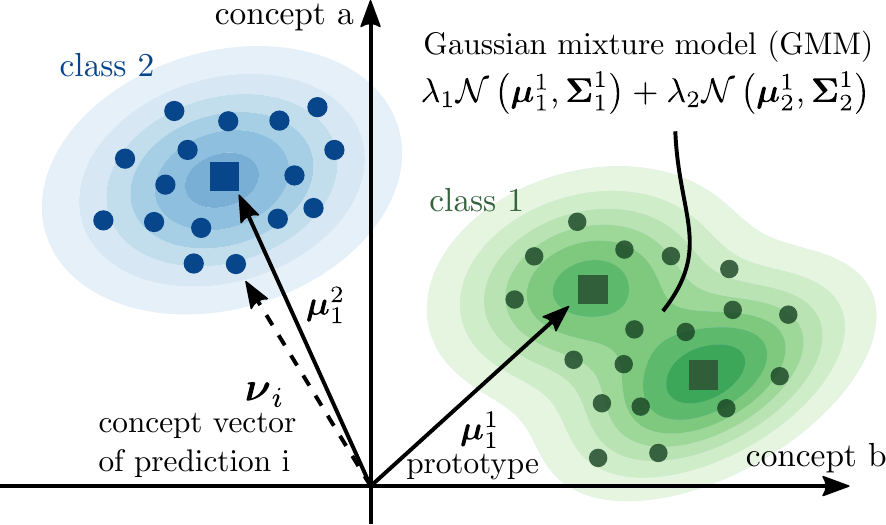}
                \includegraphics[width=0.92\linewidth]{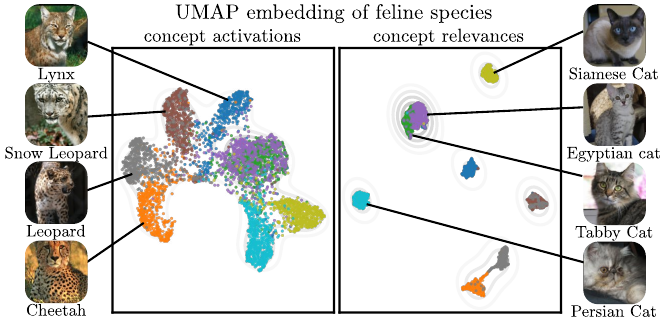}
                \captionof{figure}{Intuition behind modeling prototypes: 
                (\emph{top}): In concept space, each dimension represents the relevance or activation of a concept.
                We assume,
                that concept vectors $\boldsymbol{\nu}$ of a specific class are forming distinct clusters that can be approximated by a mixture of Gaussian distributions (\gls{gmm}).
                (\emph{bottom}):
                Concept relevances (\gls{lrp} $\varepsilon$-rule) result in more disentangled UMAP embeddings compared to activations.
                Shown are eight feline ImageNet classes (differently color-coded) for the VGG-16's last convolutional layer.
                }
                \label{fig:methods:intuition}
        \end{figure}

        In our framework,
        we collect \emph{relevances} instead of activations of features, which describe how features are utilized by the model in making specific predictions. 
        Thus, relevances naturally filter out irrelevant activations and amplify useful features for the model's class prediction. 
        As a result, relevances provide more precise and specific information regarding the encoded classes, as also illustrated in \cref{fig:methods:intuition} (\emph{bottom}) using UMAP embeddings for eight ImageNet class of feline species.
        Further, instead of assuming a single Gaussian distribution for each class,
        we model a class distribution via a multivariate \gls{gmm}.
        This is motivated by the fact, 
        that multiple sub-strategies can exist, \eg, flamingos photographed from different perspectives or distances as shown in \cref{fig:introduction:local_to_glocal}c.

        Concretely,
        as illustrated in \cref{fig:methods:intuition} (\emph{top}),
        we model the distribution $p$ of concept attributions for each class $k$ as 
        \begin{equation}
        \label{eq:app:methods:gmm}
            p^{k} = \sum_i \lambda_i^{k} p_i^k = \sum_i \lambda_i^{k} \mathcal{N}(\mean_i^{k}, \sig_i^{k})
        \end{equation} 
        with $\lambda_i^k \geq 0$ and $\sum_i \lambda_i^k = 1$ to ensure that all probabilities add up to one.
        Here,
        $\mean_i^{k}$ and $\sig_i^{k}$ correspond to the means and covariance matrices of each Gaussian.
        
        \begin{figure}[t]
                    \centering
                    \includegraphics[width=0.8\linewidth]{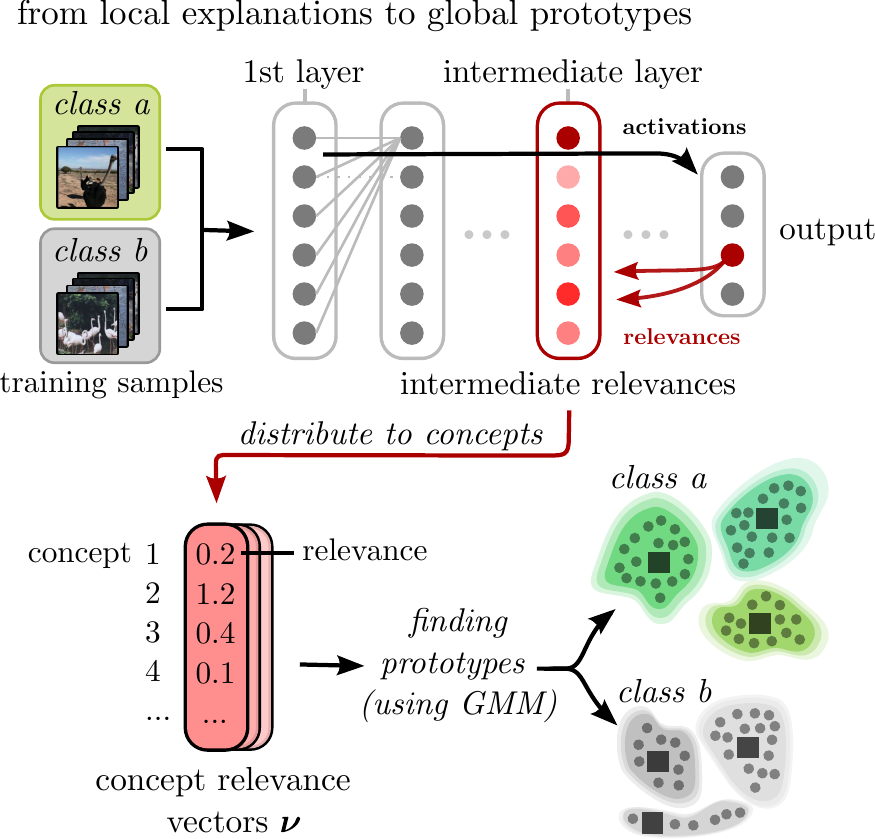}
                    \captionof{figure}{Pre-processing pipeline of \gls{ours} : 
                    \gls{dnn} predictions are generated over training samples of a specific class.
                    We further compute concept relevance scores for each prediction, representing prediction strategies.
                    By fitting \glspl{gmm} on the concept relevance vectors,
                    we find prototypical prediction strategies.
                    }
                    \label{fig:methods:framework}
        \end{figure}
        The probability density function of each Gaussian $p_i^k$ is further given as
        \begin{equation}
        \label{eq:methods:pdf}
            p_i^k(\boldsymbol{\nu}) = \frac{1}{(2\pi)^\frac{n}{2} \det( \sig^k_i)^\frac{1}{2}}e^{-\frac{1}{2} \left(\boldsymbol{\nu} - \mean^k_i\right)^\top \left(\sig^k_i\right)^{-1} \left(\boldsymbol{\nu} - \mean^k_i\right)}\,.
        \end{equation}
        Having specified the number of Gaussians,
        \glspl{gmm} then naturally provide prototypes, \ie, one for each Gaussian as in~\cite{jiang2021mixture,pereira2008prototype}.
        In fact,
        we receive a mean and covariance matrix for each prototype,
        which with \cref{eq:methods:pdf} allows to measure how likely a new point belongs to a prototype,
        further detailed in the following section.
        Note, that as the mean does not necessarily correspond to a training sample,
        we show the closest sample for illustration purposes. 
        
        To summarize, 
        \gls{ours} requires a pre-processing step to find prototypes, as outlined in \cref{fig:methods:framework}:
        For the training samples of each class,
        we compute concept relevance vectors
        on which a \gls{gmm} is fitted to provide the prototypes.

        \subsection{Quantifying the (Extra-)Ordinary}
        \label{sec:methods:dissimilarities}

        Having modeled class prediction strategies via \glspl{gmm}
        allows to compute the likeliness of new sample (prediction) to correspond to class $k$ directly via the log-likelihood $L^k$ 
        \begin{equation}
        \label{eq:methods:log_likelihood}
            L^k (\boldsymbol{\nu}) = \log p^k(\boldsymbol{\nu})\,.
        \end{equation}
        Then, a prediction with concept relevances $\boldsymbol{\nu}$ corresponds most likely to the class given as $\argmax_k L^k(\boldsymbol{\nu})$.

        Analogously,
        we can also assign a test prediction to the most likely prototypical prediction strategy $\rho^*$
        given by 
        \begin{equation}
        \label{eq:methods:assigning_proto}
            \rho^* (\boldsymbol{\nu}) = \argmax_{k,i} \log p^k_i(\boldsymbol{\nu})\,.
        \end{equation}
        Other popular metrics that can be used to assign samples to prototypical predictions
        include the Mahalanobis distance and Euclidean distance,
        as further discussed in \cref{app:metrics}.

        \paragraph{Understanding (Dis-)Similarities}
            A high likelihood as in \cref{eq:methods:log_likelihood,eq:methods:assigning_proto} directly results from small deviations
            in concept relevance scores between test and prototype prediction,
            given by the difference of concept relevance vectors
            \begin{equation}
                \boldsymbol\Delta^k_i(\boldsymbol{\nu}) = \boldsymbol{\nu} - \mean^k_i\,.
            \end{equation}
            Thus,
            we can understand, which concepts are over- and underused, corresponding to high and low entries in $\boldsymbol\Delta^k_i(\boldsymbol{\nu})$, respectively,
            or similar (with small entries in $|\boldsymbol\Delta^k_i(\boldsymbol{\nu})|$).
            It is to note that it is also possible to include information from the covariance matrices as in \cref{eq:methods:pdf}, allowing for an understanding of which concept \emph{combinations} are (un-)usual,
            further described in \cref{app:add_proto:covar}.

\section{Experiments}
    We address the following research questions:
    \begin{enumerate}
        \item \textbf{(Q1)} What global insights can we gain with prototypes?
        \item \textbf{(Q2)} How can we evaluate prototypes?
        \item \textbf{(Q3)} How can we use prototypes to validate predictions and ensure safety?
    \end{enumerate}
    \paragraph{Experimental Setting}
    We use ResNet-18 \cite{he2016deep}, VGG-16 \cite{simonyan2015very} and EfficientNet-B0 \cite{tan2019efficientnet} architectures on ImageNet~\cite{russakovsky2015imagenet}, CUB-200 \cite{welinder2010caltech} and CIFAR-10 \cite{krizhevsky2009learning}.
    Whereas models on ImageNet are pre-trained from the PyTorch model zoo,
    we train models on CUB-200 and CIFAR-10.
    Details on datasets and training are given in \cref{app:datasets_models}.
    
    \subsection{(Q1) Prototypical Concept-based Explanations}

    Class prototypes 
    allow us to inspect and understand the global prediction strategies of our model.
    Concretely,
    we can study the concepts most relevant for each prototype,
    with concept visualizations and localizations available for easier understanding
    of the concepts, as shown in \cref{fig:introduction:local_to_glocal}a.

    \paragraph{Comparing Class Strategies}

\begin{figure}
            \centering
                \includegraphics[width=0.94\linewidth]{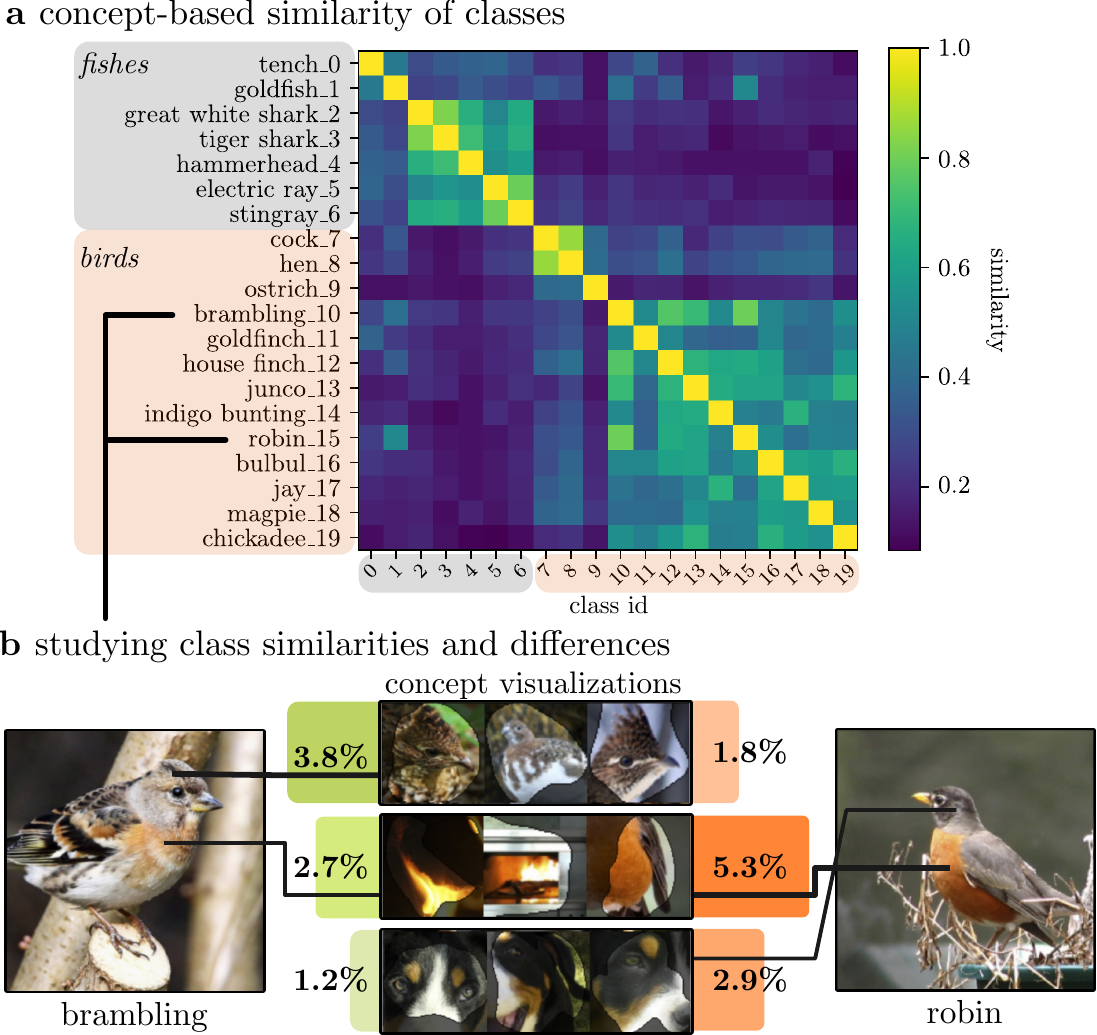}
                \captionof{figure}{
                Prototypes allow for a global understanding of class prediction (dis-)similarities.
                (\textbf{a}) Similarity matrix of the first 20 ImageNet class prototypes. We can identify distinct clusters for fishes and bird species.
                (\textbf{b}) Unraveling the (dis-)similarities of the Brambling and Robin prototype:
                Whereas both are similar in terms of orange-brown color in parts,
                they differ , \eg, in a ``gray-white spotted'' texture (indication for Brambling).
                }
                \label{fig:exp:class_prototypes:similarity_of_classes}
        \end{figure}

        To gain a global understanding at one glance,
        we visualize the similarity in prediction strategies between all classes 
        via a similarity matrix in \cref{fig:exp:class_prototypes:similarity_of_classes}a.
        Here, 
        we compute the cosine similarity between class prototypes (one per class).
        Concretely,
        the prototypes of the VGG-16 model for the first 20 ImageNet classes are shown
        when using \gls{lrp}-$\varepsilon$ for concept relevances in layer \texttt{features.28}.
        There are apparent clusters with similar prediction strategies,
        \eg, for fish and bird species.

        At this point,
        we can compare individual class strategies, such as the Brambling and Robin bird species (more examples in \cref{app:add_proto}).
        Both seem to have similar concepts, as indicated by a similarity of 80\,\%.
        Whereas both show orange-brown color in parts,
        they differ in a ``gray-white spotted'' texture (indicating Brambling) and the combination of brown, white and black patches (indicating Robin).

    \paragraph{Prototypes for Data Quality and Annotation}
    \label{sec:exp:data_quality}
     When examining \emph{multiple} prototypes per class,
     we gain a post-hoc understanding of a model's sub-strategies for decision-making,
     including, \eg, prototypes for different types of hens, or habitats for the ice bear class.
     This in turn shows promise for large-scale data annotation by assigning samples to the respective prototypes.
     While studying prototypes, 
     we encounter a multitude of problems within the ImageNet~\cite{russakovsky2015imagenet} (train) dataset such as wrong labels, poor data quality, and correlating features (shortcuts).
     All following and additional examples are provided in \cref{app:data_quality}.

     \textit{Wrong Labels:} We identify prototypes for objects not corresponding to the actual label,
     likely due to a similar naming, including tigers for ``Tiger Cat'', buses for ``passenger car'' and Leopard Lacewing butterflies for ``lacewing''.

     \textit{Correlating Features:} We reveal various correlations that the model exploits for shortcut-learning, distinctly visible in prototypes, including cats in cartons or buckets, dogs with tennis balls, and white wolves or lynxes behind fences.

    \textit{Poor Data Quality:} We further observe data quality issues, such as large numbers of low-resolution images that result in dedicated ``blur" concepts relevant for, \eg, detecting milk cans, red-breasted merganser or ties.
    Also, objects are cropped out of images in some cases due to, \eg, data augmentation, as for the ``pickelhaube'' class,
    stressing the need to inspect data after applying augmentation.
    
        \begin{table*}[t]
            \centering
                \caption{Evaluating different attribution methods for concept relevance scores used for prototypes. 
                We show results on ImageNet for 20 classes using (VGG$\,|\,$ResNet$\,|\,$EfficientNet) architectures averaged over all layers, where higher ($\uparrow$) values are better and best are bold.}
            \begin{tabular}{@{\hspace{0.2em}}lccccc@{\hspace{0.2em}}}
\toprule
 &  {Faithfulness} ($\uparrow$)&  {Stability} ($\uparrow$) &  {Sparseness} ($\uparrow$) & {Coverage} ($\uparrow$) & {Outlier Detection} ($\uparrow$)\\
 \midrule
 \gls{lrp} ($\varepsilon$-rule)~\cite{bach2015pixel}    & $     12.2\,|\,\bt{14.2}\,|\,     7.4$    & $     91.7\,|\,     90.6\,|\,96.1$ & $\bt{37.1}\,|\,\bt{36.6}\,|\,\bt{37.0}$  & $\bt{56.4}\,|\,\bt{66.5}\,|\,\bt{71.3}$ & $\bt{70.9}\,|\,\bt{78.8}\,|\,\bt{82.8}$ \\
 Input$\times$Gradient~\cite{shrikumar2017learning}     & $     12.2\,|\,\bt{14.2}\,|\,     6.7$    & $     91.8\,|\,     90.7\,|\,84.1$ & $\bt{37.1}\,|\,\bt{36.6}\,|\,     35.5$  & $     56.2\,|\,    66.3\,|\,     50.8$ & $     70.4\,|\,    78.3\,|\,     72.9$ \\
 \gls{lrp} (composite)~\cite{montavon2019layer}         & $\bt{12.6}\,|\,     13.6\,|\,\bt{7.5}$    & $     98.1\,|\,     99.0\,|\,99.0$& $     21.0\,|\,     22.8\,|\,     14.0$  & $     42.3\,|\,     56.5\,|\,     50.2$ & $     65.6\,|\,    73.3\,|\,     68.2$ \\
 GuidedBackProp~\cite{springenberg2014striving}         &$     12.0\,|\,     13.0\,|\,     6.0$     & $     98.7\,|\,\bt{99.3}\,|\,85.9$ & $     31.1\,|\,     30.9\,|\,     31.8$  & $     43.6\,|\,     59.3\,|\,     53.2$ & $     66.2\,|\,    74.8\,|\,     73.5$ \\
 Activation (max)                                       & $     11.9\,|\,     12.5\,|\,     6.3$    & $ \bt{99.3}\,|\,     99.2\,|\,\bt{99.1}$ & $     7.1\,|\,  \hspace{.24em}   4.9\, \hspace{.24em}|\,     9.8$ & $     27.5\,|\,     39.5\,|\,     36.1$ & $     54.3\,|\,    57.7\,|\,     57.3$ \\
 Activation (mean)                                      & $     11.1\,|\,     13.1\,|\,     5.9$    & $     98.7\,|\,      98.8\,|\, 92.8$ & $     11.4\,|\,     12.2\,|\,     24.0$  & $     24.8\,|\,     41.6\,|\,     36.1$ & $     55.8\,|\,    60.8\,|\,     60.2$ \\
         \bottomrule
            \end{tabular}
        
            \label{tab:exp:evaluation}
        \end{table*}
    \subsection{(Q2) Evaluating Prototypes}
    \label{sec:exp:evaluation}

        To evaluate prototypical explanations,
        three established \gls{xai} evaluation criteria in literature~\cite{hedstrom2023quantus,fel2023holistic} are applied, namely \emph{faithfulness}, \emph{stability} and \emph{sparseness}. We further introduce a \emph{coverage} measure and perform outlier detection.

        \textit{\textbf{Faithfulness:}}
            Local \gls{xai} methods compute importance scores of features, which can either be input features or latent features.
            One of the most popular evaluation methods to check an explanation's faithfulness to the model is to perform feature deletion~\cite{nauta2023anecdotal}.
            Concretely,
            the most important features (according to a chosen feature attribution method) are removed successively, \ie, set to a baseline value, and the change in model confidence is measured.
            A faithful explanation is assumed to result in a strong confidence drop, when the most important features are removed.
            In our case,
            the most relevant concepts according to the nearest class prototype are removed (\ie, set to zero activation),
            and the change in the class output logit measured.
            To receive a final score,
            the \gls{auc} is computed.

        \textit{\textbf{Stability:}}
            To evaluate stability, we compute five prototypes on $k$-fold subsets of the data ($k=10$ as default).
            We then map prototypes together using a Hungarian loss function~\cite{kuhn1955hungarian} and measure the cosine similarity between vectors.
        
            \textit{\textbf{Sparseness:}}
            We compute the cosine similarity between the absolute value of the prototype vector (\ie, centroid $\boldsymbol{\mu}$) and the unit vector, which represents a uniform distribution of concept attributions.
            The less similar a prototype is to the unit vector,
            the more sparse, and easier to interpret.
        
        \textit{\textbf{Coverage:}}
            To measure
            how well prototypes model the underlying distributions and are suitable to assign correct prediction strategies,
            we introduce the \emph{coverage} metric.
            The task is to correctly assign sample predictions from a hold-out set correctly to known sub-strategies using \cref{eq:methods:assigning_proto}.
            We therefore compute eight prototypes on concept attributions from eight (animal) classes of the same family.
            Such a setting is illustrated in \cref{fig:methods:intuition} (\emph{bottom}) for feline species. 
            Details on the groups of classes are given in \cref{app:evaluation}.

        \textit{\textbf{Outlier Detection:}}
            \glspl{gmm} do not only allow to assign samples to prototypes, but also to detect outliers.
            We adhere to the same setting as for the \emph{coverage} metric,
            but now measure how well we can detect outliers (from other classes of the same family) using \cref{eq:methods:log_likelihood}.
            Concretely, 
            the \gls{auc} is measured when plotting the true positive rate over false positive rate under a varying detection threshold.
            
    In the following,
    we evaluate the influence of various degrees of freedom of our approach.
    This includes the choice of the underlying attribution method to compute concept relevance scores,
    and the number of prototypes used to fit the \gls{gmm}.
    Note,
    that the concept basis $\mathbf{U}$, as in \cref{eq:methods:act2concept}, is also variable.
    We refer to \cite{fel2023holistic} for a thorough comparison of various techniques to compute concept bases.
    Further note, that we average the evaluation scores computed over multiple model layers (layers are detailed in \cref{app:datasets_models}).

    \subsubsection{Evaluating Concept Attribution Methods}

    We compare (modified-)gradient-based attribution methods to compute concept \emph{relevances}, including \gls{lrp} variants~\cite{bach2015pixel,montavon2019layer}, Input$\times$Gradient~\cite{shrikumar2017learning}, GuidedBackProp~\cite{springenberg2014striving} and \emph{activation} with max- and mean-pooling (details in \cref{app:evaluation}). 
    Notably,
    we refrain from using other popular methods such as SHAP and GradCAM due to their inefficiency or inapplicability, as discussed in \cref{app:evaluation}.

    When comparing the results in \cref{tab:exp:evaluation} (standard errors reported in \cref{app:evaluation}) for models on ImageNet,
    it is apparent that relevance scores (computed via local \gls{xai} methods) are not only more faithful than activation values, but also more sparse, similarly observed in \cite{dreyer2023revealing}.
    Further,
    relevances lead to better coverage and outlier detection scores,
    indicating higher disentanglement of distributions,
    as also observed in \cref{fig:methods:intuition} (\emph{bottom}). 
    Generally,
    higher-level layers result in better scores, as shown in \cref{app:add_proto}.
    
    Overall,
    \gls{lrp} ($\varepsilon$-rule) relevances result in high faithfulness, sparseness, coverage and outlier detection scores, by still providing stable prototypes.
    Thus,
    in the following,
    we use \gls{lrp} ($\varepsilon$-rule) relevances.

    \subsubsection{Varying the Number of Prototypes}
    \label{sec:exp:number_of_prototypes}

    Increasing the number of prototypes allows for a more fine-grained, but also more complex understanding. 
    Interestingly,
    \emph{faithfulness} does not significantly increase with the number of prototypes, as shown in \cref{fig:app:evaluation:num_prototypes} of \cref{app:evaluation}.
    It could be expected,
    that the closer a prototype to the actual sample (which is more probable for a larger number of prototypes),
    the higher the faithfulness score.
    Apparently,
    this is true when removing the first concepts, but for later stages
    the summarizing (global) effect of few prototypes seems to be favorable, as further detailed in \cref{app:evaluation}.

    There are, however, clear trends regarding \emph{stability}, that is decreasing, and \emph{sparseness}, which is increasing.
    Further,
    as can be expected,
    \emph{coverage} and outlier detection scores are improving as well.
    All trends are depicted in detail in \cref{fig:app:evaluation:num_prototypes} of \cref{app:evaluation}.
    We also provide qualitative examples for different prototype numbers in \cref{app:add_proto:prot_number}.

    \subsubsection{Using GMMs for Improved Clustering}
    \label{sec:exp:evaluation:clustering}
    The k-means clustering algorithm represents a simpler alternative to \glspl{gmm} for finding prototypes
    that is not based on covariance estimation.
    In fact, k-means is commonly used as a starting point to fit \glspl{gmm}~\cite{scikit-learn}.
    Compared to k-means,
    \glspl{gmm} lead to improved coverage and outlier detection scores, as further shown and discussed in \cref{sec:app:evaluation:clustering}.
    Notably,
    the covariance information is especially improving outlier detection for small numbers of prototypes.

    \subsection{(Q3) Validating Predictions}
    \label{sec:exp:prediction_validation}

        In this section,
        we leverage prototypes to validate predictions, reveal spurious model behavior, and identify \gls{ood} samples in a human-interpretable, yet automated, manner.
        
        To achieve these goals, we employ a two-step approach:
        First, we compute the class likelihood as in \cref{eq:methods:pdf}, 
        which provides a quantitative measure of how unusual a sample is to the model.
        This score allows to objectively assess spurious predictions
        and \gls{ood} samples in \cref{sec:exp:spurious,sec:exp:outlier_detection}, respectively.
        To provide detailed human-understandable information, 
        we secondly proceed to compute the difference between concepts relevance values with the prototypes,
        allowing to
        identify which features are over- or underused in the context of the given sample.

        \textit{Spotting Differences:}
        In \cref{fig:introduction:local_to_glocal}a, a sample is predicted as ``flamingo" with a class likelihood slightly below the ordinary.
        We begin by studying the deviation of relevance values in \cref{fig:introduction:local_to_glocal}b between sample and prototype.
        The sample shows strong relevance on the water concept, suggesting an unusual amount of water in the background.
        Additionally,
        the comparatively lower relevance on the redness concept indicates that the depicted flamingo lacks the expected level of redness.
        Notably, we can also analyze deviations to prototypes of other classes, offering counterfactual insights.

        \textit{Aligning to Prototypical Strategies:}
        The sample closely aligns with prototype 1, representing flamingos standing in water.
        By understanding the underlying prototypes we can thus not only understand the data (or domain) as per the model's perception,
        but also to identify similar data instances,
        \eg, more flamingos in water,
        useful for data annotation purposes.
        We further want to remark the idea of tracking prototypes during training,
        possibly giving insights into challenges such as (detecting) data drift~\cite{hoens2012learning}.    

            \begin{figure}[t] 
                \centering
                    \includegraphics[width=0.9\linewidth]{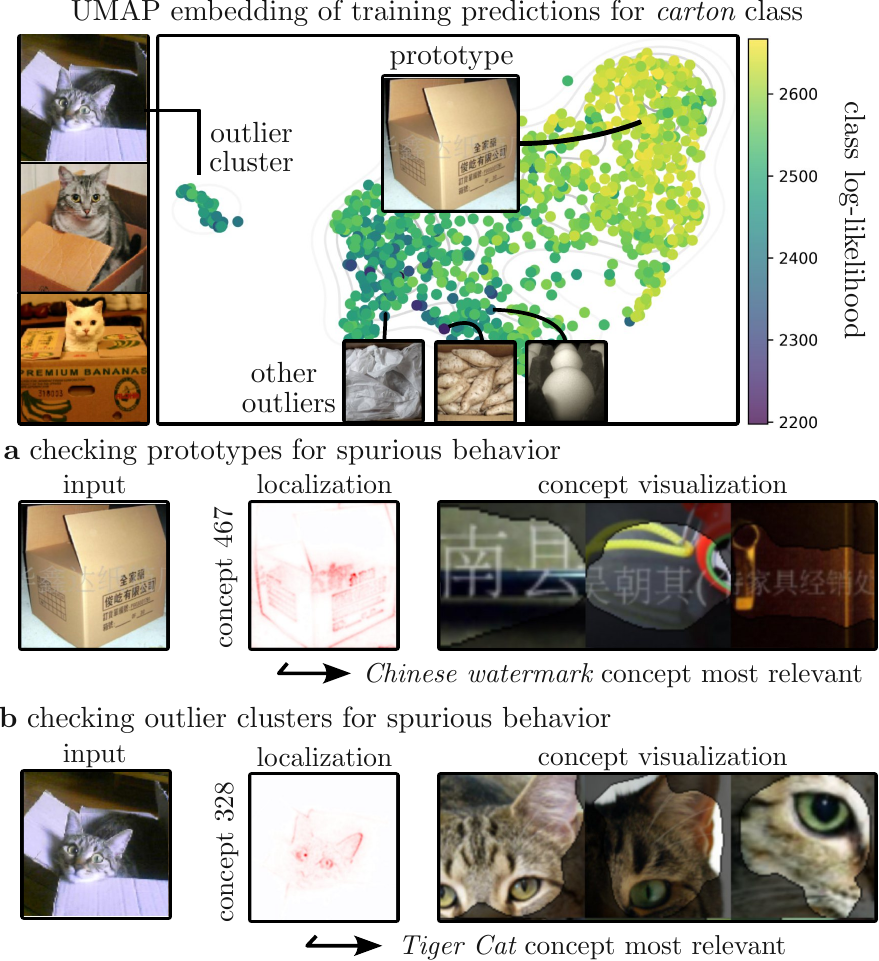}
                    \captionof{figure}{
                    Revealing spurious model behavior with \gls{ours}:
                    (\textbf{a}) Firstly, we examine the characteristic concepts of each prototype to find spurious concepts.
                    As shown, a spurious Chinese watermark concept is most relevant for the prototype of the ``carton'' class.
                    (\textbf{b}) Secondly, clusters of training predictions that deviate strongly from prototypes can be studied for spurious behavior. 
                    For the ``carton'' class,
                    we reveal a cluster of Tiger Cats in cartons,
                    that lead to the model using cat features to predict the carton class.
                    }
                    \label{fig:exp:sanity:use_case:spurious}
            \end{figure}
        \begin{table*}[t] 
        \centering
        \caption{\gls{ood} detection results for (VGG$|$ResNet$|$EfficientNet) models trained on CUB-200. Higher \gls{auc} scores are better with best bold.}
        \begin{tabular}{@{\hspace{1em}}l@{\hspace{1.5em}}c@{\hspace{2em}}c@{\hspace{1.5em}}c@{\hspace{1.5em}}c@{\hspace{1.5em}}c@{\hspace{1em}}}
        \toprule
        {} &                 LSUN &                 Places365 &             Textures &                 ImageNet &  Average \\
        \midrule
MSP~\cite{hendrycks2016baseline}    &   $\hspace{.47em}99.2\,|\,98.8\,|\,98.9$                &  $91.2\,|\,93.9\,|\,88.4$ &  $89.2\,|\,91.6\,|\,89.7$     &    $85.3\,|\,90.7\,|\,87.0$   &  $92.0$ \\
Energy~\cite{liu2020energy}          &  $\bt{100.0}\,|\,99.8\,|\,47.7$ &  $97.3\,|\,96.1\,|\,84.4$ &  $94.7\,|\,95.0\,|\,63.9$     &   $89.7\,|\,93.2\,|\,87.2$   &  $87.4$ \\
Mahalanobis~\cite{lee2018simple}     &   $\hspace{.47em}16.9\,|\,74.4\,|\,53.1$               &  $80.3\,|\,95.9\,|\,90.0$ &  $92.1\,|\,96.9\,|\,95.6$     &     $89.4\,|\,95.7\,|\,89.6$   &  $80.8$ \\
\gls{ours}-E (ours)                  &   $\hspace{.47em}99.9\,|\,99.8\,|\,99.8$               &  $95.9\,|\,97.4\,|\,94.9$ &  $98.7\,|\,98.9\,|\,98.6$     &   $93.2\,|\,95.7\,|\,92.7$   &  $97.1$ \\
\gls{ours}-GMM (ours)                &  $\bt{100.0}\,|\,\bt{99.9}\,|\,\bt{99.9}$ &  $\bt{97.9}\,|\,\bt{98.5}\,|\,\bt{96.1}$ &  $\bt{99.3}\,|\,\bt{99.3}\,|\,\bt{98.8}$     &  $\bt{95.9}\,|\,\bt{97.2}\,|\,\bt{93.6}$        &  $\mathbf{98.0}$ \\ 
        \bottomrule
        \end{tabular}
        \label{tab:exp:ood}
        \end{table*} 

        \subsubsection{Use Case: Spurious Model Behavior}
        \label{sec:exp:spurious} 
            Several works in the field of \gls{xai} have tackled the problem of revealing spurious model behavior.
            The usual approach is to study outliers, \eg, in local explanations~\cite{lapuschkin2019unmasking,anders2022finding,schramowski2020making} or latent representations~\cite{bykov2023dora,bykov2023labeling,wu2023discover}. 
            Whereas \gls{ours} allows to find and study outlier predictions,
            we also want to highlight the study of ordinary, \ie, prototypical predictions.

            An illustrative example is given by the ImageNet ``carton'' class, as shown in \cref{fig:exp:sanity:use_case:spurious}a.
            Here,
            we reveal that most characteristic for the class prototype is a gray ``Chinese watermark'' concept which is overlaid over the entire image (best observed in digital print). 

            Furthermore, our analysis extends beyond individual outlier samples to include the study of entire outlier clusters within the training set. These outlier clusters represent instances that significantly deviate from the norm and often reveal surprising and unexpected model strategies. 
            In \cref{fig:exp:sanity:use_case:spurious}b, we discover an outlier cluster consisting of samples depicting cartons with cats inside, leading the model to utilize cat-related features to increase its prediction confidence.
            Notably,
            the cat cluster receives its own prototype as we increase the number of prototypes (see \cref{app:add_proto}).

            This example underlines the value of examining prototypes to ensure safety.
            Notably, once examined, they allow for each new prediction to be automatically validated and understood when assigned to a prototype (and no outlier).

        \subsubsection{Use Case: Out-of-Distribution Detection}
        \label{sec:exp:outlier_detection}
        In \cref{sec:exp:evaluation}, we use \gls{ours} to detect outlier predictions that deviate from prototypes.
        In the following, we investigate the effectiveness
        of our approach for detecting \gls{ood} samples,
        and compare against established and dedicated methods in literature,
        namely MSP~\cite{hendrycks2016baseline} based on softmax probabilities,
        Energy~\cite{liu2020energy} and Mahalanobis~\cite{lee2018simple}.
        The task is to detect samples from unrelated datasets such as LSUN~\cite{yu2015lsun}, iSUN \cite{xu2015turkergaze}, Textures~\cite{cimpoi14describing}, SVHN~\cite{yuval2011reading} and Places365~\cite{zhou2014learning}. 
        For \gls{ours},
        we perform \gls{ood} detection by measuring the likelihood for the predicted class using \cref{eq:methods:log_likelihood} (\gls{ours}-\gls{gmm}), and alternatively also compute the Euclidean distance to the closest prototype of the predicted class (\gls{ours}-E).
        To evaluate \gls{ood} detection performance, we report the \gls{auc} when plotting the true positive rate over false positive rate under a varying detection threshold.
        \gls{ours} and Mahalanobis are hereby based on features of the last convolutional layer.
        
        \begin{figure}[t]
            \centering
                \includegraphics[width=0.99\linewidth]{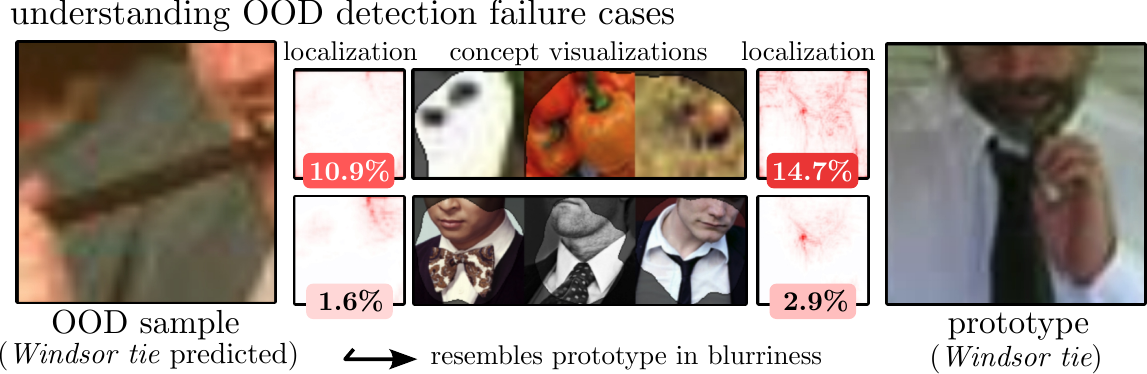}
                \captionof{figure}{Understanding why an \gls{ood} detection is classified as in-distribution: For the model,
                the blurry \gls{ood} sample is similar to a ``Windsor tie'' prototype due to the high relevance of blurring (top concept). This suggests a potential flaw in the model as it relies on a blur concept rather than the actual ``tie-like'' concept.
                }
                \label{fig:exp:ood_explanation} 
        \end{figure}
        
        For the models trained on CUB-200, \gls{ours} is most effective for \gls{ood} detection (results given in \cref{tab:exp:ood}).
        Note that we exclude bird species from ImageNet here.
        For CIFAR-10 and ImageNet models, the Energy method performs slightly better on average, as shown in \cref{app:ood_detection}.

        Importantly, \gls{ours} is intrinsically explainable (contrary to other dedicated \gls{ood} detection methods),
        allowing to understand why \gls{ood} samples are (falsely) classified as in-distribution.
        In \cref{fig:exp:ood_explanation},
        a sample from LSUN (predicted as ``Windsor tie'' by a VGG-16 on ImageNet) is similar to a class prototype,
        because of a ``blur'' concept relevant for both. 
        This reveals that the model has learned to associate blurred images with the ``Windsor tie'' class,
        as many low-resolution training samples exist.
        Thus,
        by understanding \gls{ood} failure cases, 
        we can reveal flaws of the model itself.

\section{Limitations and Future Work}
    \gls{ours} relies on estimating covariances for \glspl{gmm} which becomes unstable with few data points.
    To automatically specify the number of prototypes, studying approaches as in \cite{allen2019infinite,nauta2023pip} is of interest.
    Further, how to choose a concept basis (\ie, $\mathbf{U}$) with optimal human-interpretability is still an open question in concept-based \gls{xai} literature. 
    Improving concepts will also further increase the usefulness of \gls{ours}.
\section{Conclusion}

     \gls{ours} is a novel concept-based \gls{xai} framework that brings prototypes to local explanations,
     providing more objective and informative explanations of \glspl{dnn} in a post-hoc manner.
     Concept-based prototypes hereby enable to study the model behavior on the whole training data efficiently and in great detail,
     allowing to understand sub-strategies and issues with the data.
     As \gls{ours} bases prototype extraction on \glspl{gmm}, we receive effective quantitative measures for in- and outlier detection.
     By assessing the difference to the prototypical model behavior,
     \gls{ours} reduces the reliance on human assessment and allows for a scalable analysis of large sets of predictions.
    We demonstrate the value of our method in detecting spurious behavior by studying not only outliers, but also inliers, \ie, the prototypes. 
    Further,
    \gls{ours} shows not only effective for \gls{ood} detection,
    but is simultaneously interpretable,
    revealing flaws of the model itself through missed \gls{ood} detections.
    This work firstly introduces post-hoc concept-based prototypes,
    showcasing \gls{xai}'s potential for broader applicability in \gls{ml} validation and safety.

\subsection*{Acknowledgements}
    We thank Johanna Vielhaben for her valuable feedback.
    This work was supported by
    the Federal Ministry of Education and Research (BMBF) as grant BIFOLD (01IS18025A, 01IS180371I);
    the German Research Foundation (DFG) as research unit DeSBi (KI-FOR 5363);
    the European Union’s Horizon Europe research and innovation programme (EU Horizon Europe) as grant TEMA (101093003);
    the European Union’s Horizon 2020 research and innovation programme (EU Horizon 2020) as grant iToBoS (965221);
    and the state of Berlin within the innovation support programme ProFIT (IBB) as grant BerDiBa (10174498).
    
\clearpage 
    
    {\small
    \bibliographystyle{ieeenat_fullname}
    \bibliography{bibliography.bib}
    }

\clearpage

\appendix
\renewcommand\thefigure{A.\arabic{figure}}    
\renewcommand\thetable{A.\arabic{table}}
\setcounter{figure}{0}
\setcounter{table}{0}
\renewcommand\theequation{A.\arabic{equation}}
\setcounter{equation}{0}
\section*{Appendix}
    \section{Datasets and Models} 
    \label{app:datasets_models}
    In the following, we present all models, datasets and training procedures relevant for our experiments.

    \subsection{Models}
        We use ResNet-18 \cite{he2016deep},
        VGG-16 \cite{simonyan2015very} and EfficientNet-B0 \cite{tan2019efficientnet} architectures in our experiments.

        \paragraph{ResNet-18}
        The ResNet-18 is a convolutional neural network architecture consisting of four \texttt{BasicBlock} layers and one fully connected layer.
        For all experiments,
        we collect activation and relevance scores after each \texttt{BasicBlock} layer.
        
        \paragraph{VGG-16}
        The VGG-16 is a convolutional neural network architecture consisting of 13 convolutional layers and three fully connected layers.
        The convolutional layers are given by the identifiers
        \texttt{features.0}, \texttt{features.2}, \texttt{features.5}, \texttt{features.7}, \texttt{features.10}, \texttt{features.12}, \texttt{features.14}, \texttt{features.17}, \texttt{features.19}, \texttt{features.21}, \texttt{features.24}, \texttt{features.26}, \texttt{features.28},
        and the dense layers by 
        \texttt{classifier.0}, \texttt{classifier.3}, \texttt{classifier.6}.

        \paragraph{EfficientNet-B0}
        The EfficientNet-B0 is a convolutional neural network architecture consisting of nine \texttt{features} layers.
        For all experiments,
        we collect activation and relevance scores after each \texttt{features} layer.

    \subsection{Datasets}
        For our experiments,
        we include the datasets of ImageNet~\cite{russakovsky2015imagenet}, CUB-200 \cite{welinder2010caltech} and CIFAR-10 \cite{krizhevsky2009learning}.

        \paragraph{ImageNet}
        ImageNet is a dataset for large scale visual recognition, consisting of 1,000 object classes, totaling 14,197,122 images.
        We use the by the authors provided splits for train and test data.
        Regarding data processing, we resize images to a size where the smallest edge is 256\,px wide, with an additional center crop resulting in 224$\times$224\,px image size. 
        Finally,
        images are normalized with mean $(0.485, 0.456, 0.406)$ and standard deviation $(0.229, 0.224, 0.225)$ over the red, green and blue color channel.

        \paragraph{CUB-200}
        CUB-200 is a visual categorization task dataset consisting of 11,788 images of 200 subcategories belonging to birds, 5,994 for training and 5,794 for testing.
        Regarding data processing, we resize images to a size where the smallest edge is 224\,px wide, with an additional center crop resulting in 224$\times$224\,px image size. 
        Finally,
        images are normalized with mean $(0.47473491, 0.48834997, 0.41759949)$ and standard deviation $(0.22798773, 0.22288573, 0.25982403)$ over the red, green and blue color channel. 

        \paragraph{CIFAR-10}
        The CIFAR-10 dataset consists of 60,000 colour images in 10 object classes, with 6,000 images per class.
        There are 50,000 training images and 10,000 test images in total.
        Regarding data processing, we resize images to 32$\times$32\,px without further normalization. 
        
    \subsection{Training}
        Whereas models on ImageNet are pre-trained and taken from the PyTorch model zoo,
        we train models on CUB-200 and CIFAR-10.
        Hereby,
        all models are trained for 100 epochs using the stochastic gradient descent (VGG, ResNet) or ADAM~\cite{diederik2015adam} (EfficientNet) algorithm.
        
        \paragraph{CUB-200}
        For training, we use a batch size of 32.
        The initial learning rates are $10^{-3}$ for the VGG and ResNet architecture, and $5\cdot10^{-4}$ for the EfficientNet.
        For data augmentation, Gaussian noise (zero mean, standard deviation of 0.05), random horizontal flips (probability of 0.5), random rotation (up to 10 degrees), random translation (up to 20\,\% of edge length in all directions) as well as a random scaling (between 80 and 120\,\%) is applied.
        
        \paragraph{CIFAR-10}
        For training, we use a batch size of 512.
        The initial learning rates are $10^{-2}$ for the VGG and ResNet architecture, and $5\cdot10^{-3}$ for the EfficientNet.
        The learning rate is decreased to one tenth after 50 and 75 epochs.
        For data augmentation, random horizontal flips (probability of 0.5), as well as a random crop (to 32$\times$32\,px, probability 0.5) after padding images with four pixels of zeros is applied.

\renewcommand\thefigure{B.\arabic{figure}}    
\renewcommand\thetable{B.\arabic{table}}
\setcounter{figure}{0}
\setcounter{table}{0}
\renewcommand\theequation{B.\arabic{equation}}
\setcounter{equation}{0}
    \section{Alternative Metrics for Sample-to-Prototype Assignment}
    \label{app:metrics}

    In the following, we present details for how to assign a new prediction with concept relevance vector $\boldsymbol{\nu}$ to a prototypical prediction strategy $\mean^k_i$ (for prototype $i$ of class $k$).
    
    \paragraph{\gls{gmm}}
    The probability density function of a Gaussian distribution (from the fitted \gls{gmm}) is given as 
        \begin{equation}
        \label{eq:app:methods:pdf}
            p_i^k(\boldsymbol{\nu}) = \frac{1}{(2\pi)^\frac{n}{2} \det( \sig^k_i)^\frac{1}{2}}e^{-\frac{1}{2} \left(\boldsymbol{\nu} - \mean^k_i\right)^\top \left(\sig^k_i\right)^{-1} \left(\boldsymbol{\nu} - \mean^k_i\right)}
        \end{equation}
          and serves as a direct means to assign predictions to a prototype.
          Concretely,
          we compute the log-likelihood as
              \begin{equation}
        \label{eq:app:methods:log_likelihood}
            L^k_i (\boldsymbol{\nu}) = \log p^k_i(\boldsymbol{\nu})\
        \end{equation}
        and assign predictions to the prototype with highest log-likelihood as
                \begin{equation}
        \label{eq:app:methods:assigning_proto}
            \rho^* (\boldsymbol{\nu}) = \argmax_{k,i} \log p^k_i(\boldsymbol{\nu})\,.
        \end{equation}
        Note,
        when the task is to assign any prediction to a class instead of prototype,
        we use the probability density $p^k$ of the \gls{gmm} as given in Equation~\eqref{eq:app:methods:gmm}.
    
    \paragraph{Mahalanobis Distance}
        Alternatively,
        one can also use the Mahalanobis distance given as
        \begin{equation}
        \label{eq:app:methods:mahalanobis_distance}
            d_{\text{MD}, i}^k(\boldsymbol{\nu}) = \sqrt{\left(\boldsymbol{\nu} - \mean^k_i\right)^\top \left(\sig^k_i\right)^{-1} \left(\boldsymbol{\nu} - \mean^k_i\right)}
        \end{equation}
        to assign predictions to prototypes.
        Concretely,
        we assign a prediction to the prototype with the smallest distance as
        \begin{equation}
        \label{eq:methods:assigning_proto_mahala}
            \rho^* (\boldsymbol{\nu}) = \argmin_{k,i} d_{\text{MD}, i}^k(\boldsymbol{\nu})\,.
        \end{equation}
        For assigning a prediction to a class,
        we assign it to the class of the closest prototype.

    \paragraph{Euclidean Distance}
        As a simple alternative,
        one can also use the Euclidean distance given as
        \begin{equation}
        \label{eq:app:methods:euclidean_distance}
            d_{\text{E}, i}^k(\boldsymbol{\nu}) = \sqrt{\left(\boldsymbol{\nu} - \mean^k_i\right)^\top \left(\boldsymbol{\nu} - \mean^k_i\right)}\,.
        \end{equation}
        to assign predictions to prototypes.
        Concretely,
        we assign a prediction to the prototype with the smallest distance as
        \begin{equation}
        \label{eq:methods:assigning_proto_euclidean}
            \rho^* (\boldsymbol{\nu}) = \argmin_{k,i} d_{\text{E}, i}^k(\boldsymbol{\nu})\,.
        \end{equation}
        For assigning a prediction to a class,
        we assign it to the class of the closest prototype.
        It is to note,
        that whereas log-likelihood and Mahalanobis distance are requiring covariance matrices,
        Euclidean distance does not.
        Thus, Euclidean distance can also be used as a lightweight alternative, when prototypes are not computed via \glspl{gmm} but, \eg, k-means.
        However,
        as experiments show, \eg, Section~\ref{sec:exp:outlier_detection},
        covariance information is beneficial for better modeling of underlying distributions.
        Then, 
        as shown in Figure~\ref{fig:methods:intuition}b, Mahalanobis or log-likelihood are more accurate in assigning predictions to the true class.
        
\renewcommand\thefigure{C.\arabic{figure}}    
\renewcommand\thetable{C.\arabic{table}}
\setcounter{figure}{0}
\setcounter{table}{0}
\renewcommand\theequation{C.\arabic{equation}}
\setcounter{equation}{0}
    \section{Evaluating and Inspecting Prototypes}
    \label{app:evaluation}
    In this section,
    we provide additional details and results for the experiments regarding the evaluation and inspection of prototypes, \eg, Sections~\ref{sec:exp:data_quality} and~\ref{sec:exp:evaluation}.

    \subsection{Choice of the Attribution Method}
    For the results in Table~\ref{tab:exp:evaluation}, where we evaluate prototypes based on concept relevance scores from different attribution methods,
    we provide \gls{se} values in Table~\ref{tab:app:evaluation:standard_error}.
    For faithfulness, 
    we compute the \gls{auc} on eight subsets of the data (totaling 300 samples) to estimate the \gls{se} of the mean.
    Regarding stability and sparseness,
    we collect all individual values and compute the \gls{se} of the mean.
    For coverage,
    we compute the accuracy in assigning test samples to known eight sub-strategies.
    For estimating the sub-strategies, we use half of the training set from ImageNet, and the other half for the test set.
    Then, for each of the seven animal families (listed in \cref{app:evaluation:species}),
    we chose eight random classes (each corresponding to a sub-strategy) seven times.
    Thus, we compute the \gls{se} of the mean over 49 coverage scores.
    Lastly,
    for outlier detection,
    we compute the \gls{auc} for differentiating between predictions of eight known sub-stragies and five other classes of the same family.
    Again,
    for each of the seven animal families,
    we chose eight random classes (each corresponding to a sub-strategy) and five random outlier classes seven times.
    In total, we compute the \gls{se} of the mean over 49 scores.

                    \begin{table*}[t]
            \centering
                \caption{Standard Errors for the results reported in Table~\ref{tab:exp:evaluation} regarding the evaluation of different attribution methods for concept relevance scores used for prototypes.
                We report values for ImageNet with 20 classes using (VGG$\,|\,$ResNet$\,|\,$EfficientNet) architectures averaged over all layers.}
            \begin{tabular}{lccccc}
                \toprule
                 &  {Faithfulness} &  {Stability} &  {Sparseness} & {Coverage} & {Outlier Detection}\\
                 \midrule
                 LRP ($\varepsilon$-rule)~\cite{bach2015pixel} & 
                 $     0.07\,|\,     0.11\,|\,     0.03$ &
                 $     0.01\,|\,     0.01\,|\,     0.01$&
                 $     0.8\,|\,     1.2\,|\,     2.5$ & 
                 $     0.3\,|\,     0.5\,|\,     0.3$ &
                 $     0.2\,|\,     0.4\,|\,     0.2$
                 \\ 
                 Input$\times$Gradient~\cite{shrikumar2017learning}  & 
                  $     0.07\,|\,     0.11\,|\,     0.03$ &
                  $     0.01\,|\,     0.01\,|\,     0.03$& 
                  $     0.8\,|\,     1.2\,|\,     1.9$ & 
                  $     0.3\,|\,     0.5\,|\,     0.3$ &
                  $     0.2\,|\,     0.4\,|\,     0.2$
                  \\
                 LRP (composite)~\cite{montavon2019layer}  & 
                 $     0.07\,|\,     0.12\,|\,     0.03$ &
                 $     0.01\,|\,     0.01\,|\,     0.01$&
                 $     0.6\,|\,     0.2\,|\,     0.1$ & 
                 $     0.2\,|\,     0.5\,|\,     0.3$ &
                 $     0.2\,|\,     0.4\,|\,     0.3$
                 \\ 
                 GuidedBackProp~\cite{springenberg2014striving} & 
                 $     0.07\,|\,     0.11\,|\,     0.02$ &
                 $     0.01\,|\,     0.01\,|\,     0.04$& 
                 $     0.2\,|\,     0.2\,|\,     1.4$ & 
                 $     0.2\,|\,     0.4\,|\,     0.3$ &
                 $     0.2\,|\,     0.4\,|\,     0.2$
                 \\ 
                 Activation (max)   & 
                 $     0.07\,|\,     0.10\,|\,     0.03$ &
                 $     0.01\,|\,     0.01\,|\,     0.01$& 
                 $     0.2\,|\,     0.1\,|\,     0.2$ & 
                 $     0.2\,|\,     0.4\,|\,     0.2$ &
                 $     0.2\,|\,     0.5\,|\,     0.3$
                 \\ 
                 Activation (mean)  & 
                 $     0.06\,|\,     0.11\,|\,     0.03$ &
                 $     0.01\,|\,     0.01\,|\,     0.01$&
                 $     0.2\,|\,     0.2\,|\,     0.4$ & 
                 $     0.2\,|\,     0.5\,|\,     0.3$ &
                 $     0.2\,|\,     0.5\,|\,     0.3$
                 \\  
         \bottomrule
            \end{tabular}
        
            \label{tab:app:evaluation:standard_error}
        \end{table*}

    \subsection{Increasing the Number of Prototypes}

\begin{figure}[t]
            \centering
                \includegraphics[width=0.99\linewidth]{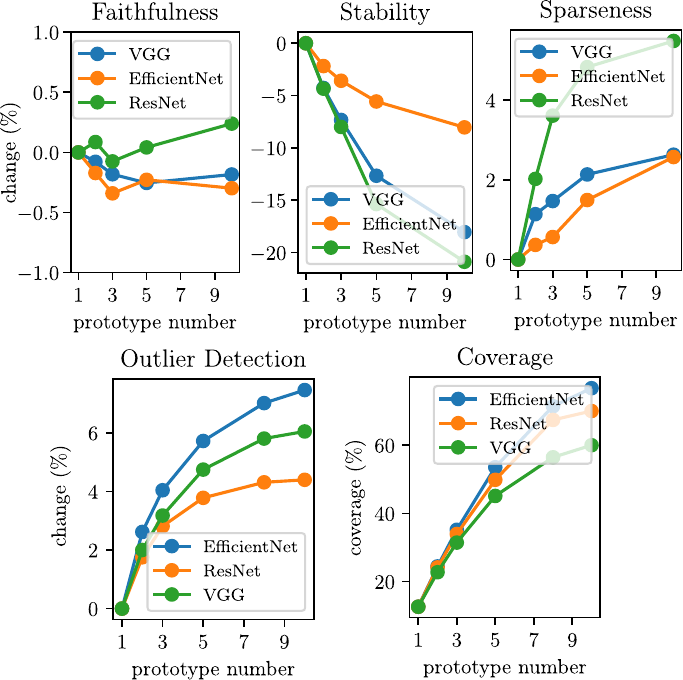}
                \captionof{figure}{
                Effect of increasing the number of prototypes per class on the evaluation metrics. We show the relative change compared to one prototype for all architectures on 20 ImageNet classes using LRP-$\varepsilon$ for attributions. For better understanding, we report the actual values for the coverage metric.
                }
                \label{fig:app:evaluation:num_prototypes}
        \end{figure}

    In the following, we provide more results for the evaluation of the number of prototypes \wrt the metrics defined in Section~\ref{sec:exp:evaluation}.
    The change in metrics when increasing the prototype number is shown in Figure~\ref{fig:app:evaluation:num_prototypes}.
    Here,
    we can see that stability decreases for higher numbers of prototypes.
    Sparseness,
    outlier detection and coverage scores however increase.
    Faithfulness scores do not significantly change, as described in more detail in the following.

\begin{figure}[t] 
                \centering
                    \includegraphics[width=0.9\linewidth]{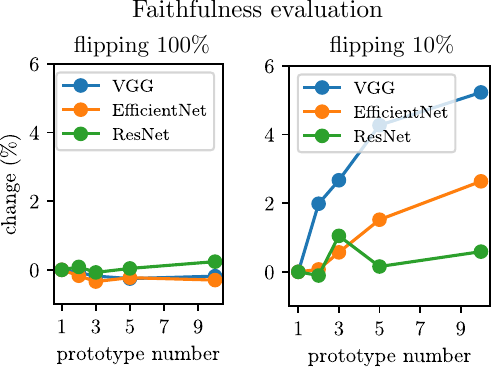}
                    \captionof{figure}{
                    The change in the faithfulness metric when increasing the number of prototypes, as measured in Section~\ref{sec:exp:number_of_prototypes}.
                    (\emph{left}): For evaluating faithfulness, all concepts are removed successively. 
                    No significant effect \wrt the number of prototypes is visible.
                    (\emph{right}): When only 10\,\% of all concepts are removed,
                    we can observe a stronger increase in faithfulness.
                    }
                    \label{fig:app:evaluation:faithfulness}
            \end{figure}

    \paragraph{Faithfulness}
    When increasing the number of prototypes,
    we can not measure a significant increase in faithfulness as reported in Section~\ref{sec:exp:number_of_prototypes} and shown again in Figure~\ref{fig:app:evaluation:faithfulness} (\emph{left}).
    Notably,
    however,
    when we remove only 10\,\% of the most relevant concepts (instead of all),
    we can measure a significant effect on faithfulness when increasing the number of prototypes (faithfulness increases), as shown in Figure~\ref{fig:app:evaluation:faithfulness} (\emph{right}).
    The higher the number of prototypes,
    the more likely it is that a prototype is close to any test prediction point.
    Thus,
    it could be assumed, that removing concepts according to the closest prototype is leading to a higher faithfulness with a higher number of prototypes.
    This is however only true for the first removed concepts, as shown in Figure~\ref{fig:app:evaluation:faithfulness} (\emph{right}).
    Here,
    a low number of prototypes (\eg, more global prototypes) seem to be favorable when a high number of concepts are removed.

    \subsection{Evaluating Clustering Algorithms}
    \label{sec:app:evaluation:clustering}

        In Section~\ref{sec:exp:evaluation:clustering},
        we compare the k-means algorithm against \glspl{gmm} for modeling sub-strategies of a model.
        Concretely,
        using k-means, we first find prototype centroids,
        and thereafter,
        we fit a \gls{gmm} using the k-means centroids as a starting point.
        Compared to k-means,
        \glspl{gmm} are thus based on updated centroids and covariance information. 

        In order to assign test samples,
        we use the Euclidean distance for k-means, and the log-likelihood for \glspl{gmm}.
        Specifically,
        for coverage scores, we assign a prediction to the closest prototype centroid, as in Equation~\eqref{eq:methods:assigning_proto} for \gls{gmm} and use Equation~\eqref{eq:methods:assigning_proto_euclidean} for Euclidean distances.
        For outlier detection,
        we compute the log-likelihood on the class-level using Equation~\eqref{eq:methods:log_likelihood} for \glspl{gmm} and measure the smallest distance to any prototype for Euclidean distance using Equation~\ref{eq:methods:assigning_proto_euclidean}.
        As an additional baseline,
        we use the updated centroids from the \gls{gmm} together with Euclidean distance.
        This way, we can decouple the effects of updated centroids and covariance information.
        
        \begin{table}[t]
            \centering
                \caption{Effect of different clustering algorithms on the coverage and outlier detection scores using \gls{lrp} ($\varepsilon$-rule) concept attributions. We report values for ImageNet with 20 classes using (VGG$\,|\,$ResNet$\,|\,$EfficientNet) architectures averaged over all layers. (Euc.: Euclidean distance is used instead of log-likelihood.)}
            \begin{tabular}{lcc}
                \toprule
                &  {Coverage} & {Outlier Detection}\\
                 \midrule
                 k-means (Euc.)   &  $55.7\,|\,  62.8\,|\,   66.9$  &  $69.0\,|\,  76.7\,|\,   77.6$  \\
                 \gls{gmm} (Euc.) &  $ 56.1\,|\,   63.9\,|\,  68.6$ &  $ 69.0\,|\,   77.0\,|\,  77.7$ \\
                 \gls{gmm}        &  $\bt{56.4}\,|\,   \bt{66.5}\,|\,   \bt{71.3}$ &  $\bt{70.9}\,|\,   \bt{78.8}\,|\,   \bt{82.8}$ \\ 
            \bottomrule
            \end{tabular}
        
            \label{tab:app:evaluation:clustering}
        \end{table}

\begin{figure*}[t] 
        \centering
            \includegraphics[width=0.8\linewidth]{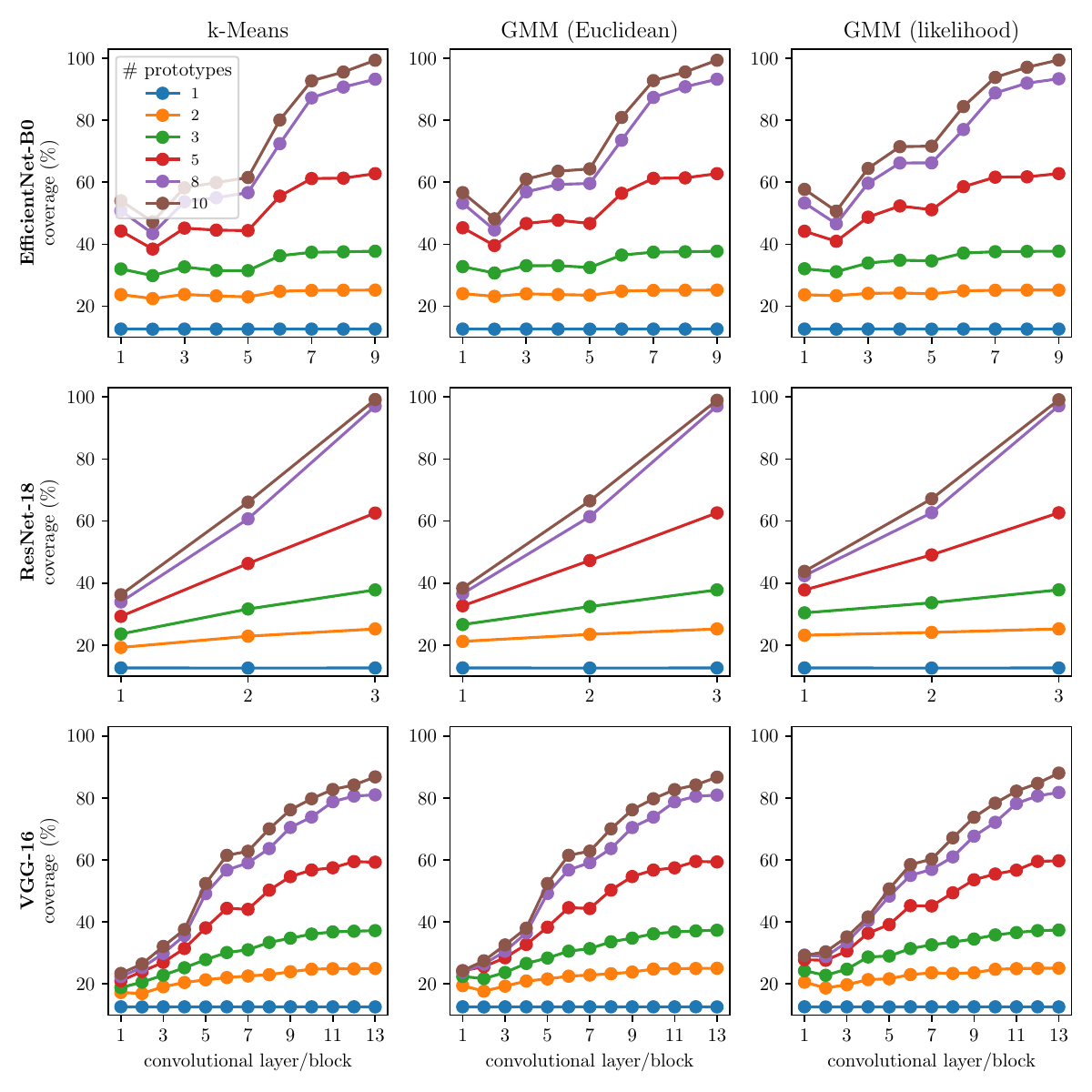}
            \captionof{figure}{
            Effect of different clustering algorithms on the coverage scores using \gls{lrp} ($\varepsilon$-rule) concept attributions over all model layers for different numbers of prototypes.
            }
            \label{fig:app:evaluation:coverage_n}
    \end{figure*}

\begin{figure*}[t] 
        \centering
            \includegraphics[width=0.8\linewidth]{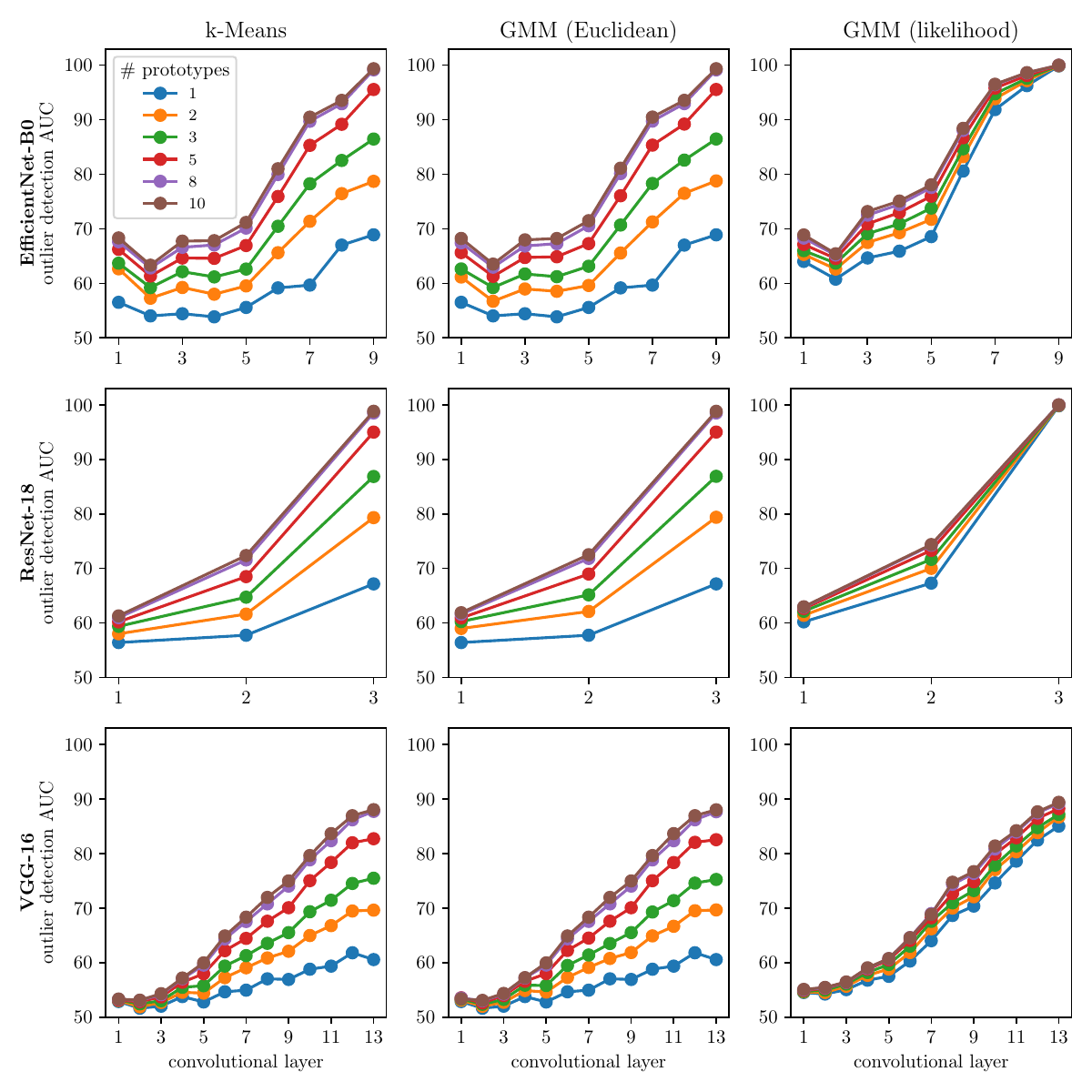}
            \captionof{figure}{
            Effect of different clustering algorithms on the object detection scores using \gls{lrp} ($\varepsilon$-rule) concept attributions over all model layers for different numbers of prototypes.
            }
            \label{fig:app:evaluation:od_n}
    \end{figure*}

        The resulting coverage and outlier detection scores using \gls{lrp} ($\varepsilon$-rule) concept attributions are reported in Table~\ref{tab:app:evaluation:clustering}. 
        Here,
        it becomes apparent, that both the updated centroids and covariance information are beneficial for finding correct sub-strategies and detecting outliers.
        We further show layer-wise scores for all models and for different number of prototypes in Figures~\ref{fig:app:evaluation:coverage_n} (coverage) and \ref{fig:app:evaluation:od_n} (object detection).
        It is visible,
        that covariance information is especially useful for detecting outliers when the prototype number is low.
        

    \subsection{ImageNet species}
    \label{app:evaluation:species}
    For the coverage and outlier detection evaluation,
    we perform experiments over animal classes of the same family.
    The families and classes/species are as follows:
    
        ``\textbf{terrier}'': [``Staffordshire bullterrier'', ``American Staffordshire terrier'', ``Bedlington terrier'', ``Border terrier'', ``Kerry blue terrier'', ``Irish terrier'', ``Norfolk terrier'', ``Norwich terrier'', ``Yorkshire terrier'', ``wire-haired fox terrier'', ``Lakeland terrier'', ``Sealyham terrier'', ``Airedale'', ``cairn'', ``Australian terrier'', ``Dandie Dinmont'', ``Boston bull'', ``miniature schnauzer'', ``giant schnauzer'', ``standard schnauzer'', ``Scotch terrier'', ``Tibetan terrier'', ``silky terrier'', ``soft-coated wheaten terrier'', ``West Highland white terrier'', ``Lhasa'']
        
        ``\textbf{working dog}'': [``kuvasz'', ``schipperke'', ``groenendael'', ``malinois'', ``briard'', ``kelpie'', ``komondor'', ``Old English sheepdog'', ``Shetland sheepdog'', ``collie'', ``Border collie'', ``Bouvier des Flandres'', ``Rottweiler'', ``German shepherd'', ``Doberman'', ``miniature pinscher'', ``Greater Swiss Mountain dog'', ``Bernese mountain dog'', ``Appenzeller'', ``EntleBucher'', ``boxer'', ``bull mastiff'', ``Tibetan mastiff'', ``French bulldog'', ``Great Dane'', ``Saint Bernard'', ``Eskimo dog'', ``malamute'', ``Siberian husky'', ``affenpinscher'']
        
        ``\textbf{bird}'': [``cock'', ``hen'', ``ostrich'', ``brambling'', ``goldfinch'', ``house finch'', ``junco'', ``indigo bunting'', ``robin'', ``bulbul'', ``jay'', ``magpie'', ``chickadee'', ``water ouzel'', ``kite'', ``bald eagle'', ``vulture'', ``great grey owl'']
        
        ``\textbf{fish}'': [``tench'', ``goldfish'', ``great white shark'', ``tiger shark'', ``hammerhead'', ``electric ray'', ``stingray'', ``barracouta'', ``eel'', ``coho'', ``rock beauty'', ``anemone fish'', ``sturgeon'', ``gar'', ``lionfish'', ``puffer'']
        
        ``\textbf{primate}'': [``orangutan'', ``gorilla'', ``chimpanzee'', ``gibbon'', ``siamang'', ``guenon'', ``patas'', ``baboon'', ``macaque'', ``langur'', ``colobus'', ``proboscis monkey'', ``marmoset'', ``capuchin'', ``howler monkey'', ``titi'', ``spider monkey'', ``squirrel monkey'', ``Madagascar cat'', ``indri'']
        
        ``\textbf{feline, felid}'': [``tabby'', ``tiger cat'', ``Persian cat'', ``Siamese cat'', ``Egyptian cat'', ``cougar'', ``lynx'', ``leopard'', ``snow leopard'', ``jaguar'', ``lion'', ``tiger'', ``cheetah'']
        
        ``\textbf{snake, serpent, ophidian}'': [``thunder snake'', ``ringneck snake'', ``hognose snake'', ``green snake'', ``king snake'', ``garter snake'', ``water snake'', ``vine snake'', ``night snake'', ``boa constrictor'', ``rock python'', ``Indian cobra'', ``green mamba'', ``sea snake'', ``horned viper'', ``diamondback'', ``sidewinder'']

    \subsection{Methods for Concept Attribution Computation}
    \label{app:attributions}
    In the following, we provide details on how we compute concept attribution scores using activations and local \gls{xai} methods,
    including \gls{lrp}, Input$\times$Gradient and GuidedBackprop. 
    For simplicity,
    we assume that each neuron in a layer corresponds to a concept.
    Therefore,
    no transformation between latent feature and space and concept space is required, as described by Equation~\eqref{eq:methods:act2concept} in the main manuscript.

    \paragraph{Activation}

        In each convolutional layer (with $n$ channels of spacial dimension $w\times h$),
        we collect the activations $\mathbf{A}\in \mathbb{R}^{n\times w \times h}$.
        We then either perform max- or sum-pooling to compute concept attribution vectors $\boldsymbol{\nu} \in \mathbb{R}^n$, \ie,
        \begin{equation}
            \nu_i = \max_{uv} A_{iuv} \quad \text{or} \quad \nu_i = \sum_{uv} A_{iuv},
        \end{equation}
        respectively.

    \paragraph{Input$\times$Gradient}
        We define the function $g:\mathbb{R}^{n\times w \times h} \rightarrow \mathcal{Y}$, corresponding the part of the model architecture that maps latent activations $\mathbf{A}\in \mathbb{R}^{n\times w \times h}$ to the model output $y_k$ of class $k$.
        Then, concept relevance scores are given by
        \begin{equation}
        \label{eq:app:input_gradient}
            \nu_i = \sum_{uv} A_{iuv} \frac{\partial g_k(\mathbf{A})}{\partial A_{iuv} } ,
        \end{equation}
        multiplying latent gradients with concept activations.

    \paragraph{GuidedBackprop}
        GuidedBackprop is based on the idea of Input$\times$Gradient,
        using Equation~\eqref{eq:app:input_gradient} to compute concept relevance scores.
        However,
        the work of \cite{springenberg2014striving} further proposes to prevent the backward flow of negative gradients through ReLU activation functions by setting negative entries of the top gradient to zero.

    \paragraph{\gls{lrp}}

    Regarding our latent predictor $g$ with $L$ layers
    \begin{equation} \label{eq:lrp_nn}
        g(\mathbf{A})=f_L \circ \dots \circ  f_1(\x)~,
    \end{equation}
    \gls{lrp} follows the flow of activations computed during the forward pass through the model in opposite direction, from the final layer $f_L$ back to the latent layer $f_1$. 
    
    The \gls{lrp} method distributes relevance quantities $R_j^{l+1}$ (corresponding to neuron $j$ in layer $l+1$) towards a lower layer $l$ proportionally to the relative contributions $z_{ij}$ of lower-layer neuron $i$ to the (pre-)activation $z_j$ as
    \begin{equation} \label{eq:lrp_basic}
        R_{i \leftarrow j}^{(l,\: l+1)} = \frac{z_{ij}}{z_j}R_j^{l+1},
    \end{equation}
    where the pre-activation $z_j$ of neuron $j$ for a linear layer operation is given by $z_j = \sum_i z_{ij}$.
    Then, lower neuron relevance is obtained by losslessly aggregating all incoming relevance messages $R_{i \leftarrow j}$ as
    \begin{equation} \label{eq:lrp_aggregation}
        R_i^l = \sum_j R_{i \leftarrow j}^{(l,\: l+1)}.
    \end{equation}
    
    For convolutional layers,
    we have to include the spatial dimensions of latent feature maps resulting in
    \begin{equation}
        R_{(iuv) \leftarrow (pqr)}^{(l,\: l+1)}  =  \frac{z_{(uvi)(pqr)}}{z_{pqr}}R_{pqr}^{l+1}.
    \end{equation}
    
    Then,
    the latent relevance of neuron $i$ is computed as 
    \begin{equation}
        R_i^l = \sum_{uv}\sum_{pqr}R_{(iuv) \leftarrow (pqr)}^{(l,\: l+1)}.
    \end{equation}
    Note,
    that final concept relevance scores are then given when we reach the input level, \ie, $l=1$.
    
    \textit{\gls{lrp} $\varepsilon$-rule:}
    To ensure numerical stability,
    the \gls{lrp} $\varepsilon$-rule is introduced and defined as
    \begin{align}
    R^{(l,\: l+1)}_{i \leftarrow j} & =  \frac{z_{ij}}{z_j + \varepsilon \cdot \text{sign}(z_j)}R_j^{l+1}.
    \end{align}
    Note that for the purpose of numerical stability, the definition of the sign function is altered such that $\text{sign}(0) = 1$.
    The $\varepsilon$-rule is in ReLU networks highly similar to the multiplication of the input times its gradient~\wrt the output~\cite{kohlbrenner2020towards}.
    However, due to its similarity to gradient-based attribution computation, the \gls{lrp} $\varepsilon$-rule may result in noisy attributions in very deep models, where gradient shattering and noisy gradients appear \cite{balduzzi2017shattered}. 
    
    \textit{\gls{lrp} composite:}
    An alternative approach is to combine multiple propagation rules, so-called rule composites, which result in attributions that are less influenced by a noisy gradient: the LRP ${\varepsilon{z^+}}$-rule is an established best practice \cite{kohlbrenner2020towards,montavon2019layer} to keep LRP attribution maps informative, readable and representative while combating gradient shattering effects.
    The LRP ${{z^+}}$-rule operates on the convolutional layers and \gls{lrp} $\varepsilon$-rule is utilized in standard dense layers. 
    The LRP ${{z^+}}$-rule is given as
    \begin{align} \label{eq:appendix:zplus-rule}
    R^{(l,\: l+1)}_{i \leftarrow j} & =  \frac{(z_{ij})^+}{z_j^+}R_j^{l+1}
    \end{align}
    by only taking into account positive contributions $z_j^+ = \sum_i (z_{ij})^+$ with $(\cdot)^+ =\max(0, \cdot)$.
    
    \paragraph{Normalization of Concept Relevance Scores}
    Please note,
    that in order to be interpretable as percentage scores, we normalize concept relevances to an absolute sum of one, \ie, $\sum_i \nu'_i = 1$ with $\nu'_i = \frac{\nu_i}{\sum_j|\nu_j|}$.
    
    \paragraph{Other Attribution Methods}

        We refrain from using other popular attribution methods such as SHAP~\cite{lundberg2017unified}, LIME~\cite{ribeiro2016should}, and GradCAM~\cite{selvaraju2017grad} due to their infeasibility or inapplicability.
        Concretely,
        SHAP and LIME require perturbation of features,
        resulting in a multitude of separate forward passes that need to be performed.
        Further,
        compared to perturbation-based methods, the backpropagation-based feature attribution methods previously presented provide concept relevances for \emph{all} layers in a \emph{single} backward pass.
        GradCAM~\cite{selvaraju2017grad} provides input heatmaps that localize important features by up-sampling latent feature maps.
        As such, GradCAM can only be utilized for spatial localization of concepts, but not to compute latent feature scores.

    \subsection{Investigating Data Quality}
    \label{app:data_quality}

    In the following,
    we present examples for Section~\ref{sec:exp:data_quality} on how prototypes enable an understanding of the data itself (and how the model uses the data).
    
    \subsubsection{Discovering Model (Sub-)Strategies} 
\begin{figure*}[t] 
                \centering
                    \includegraphics[width=0.49\linewidth]{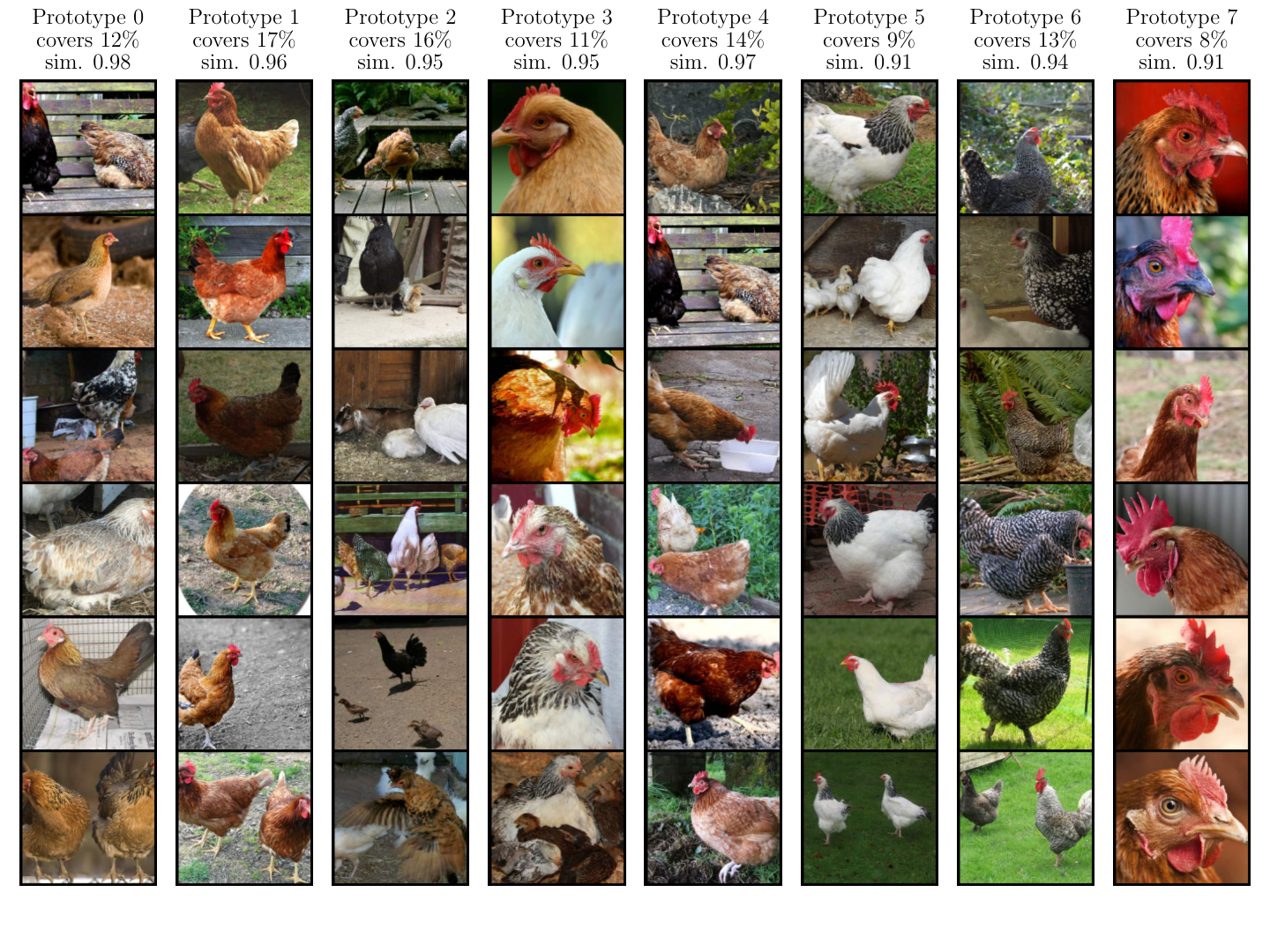}
                    \includegraphics[width=0.49\linewidth]{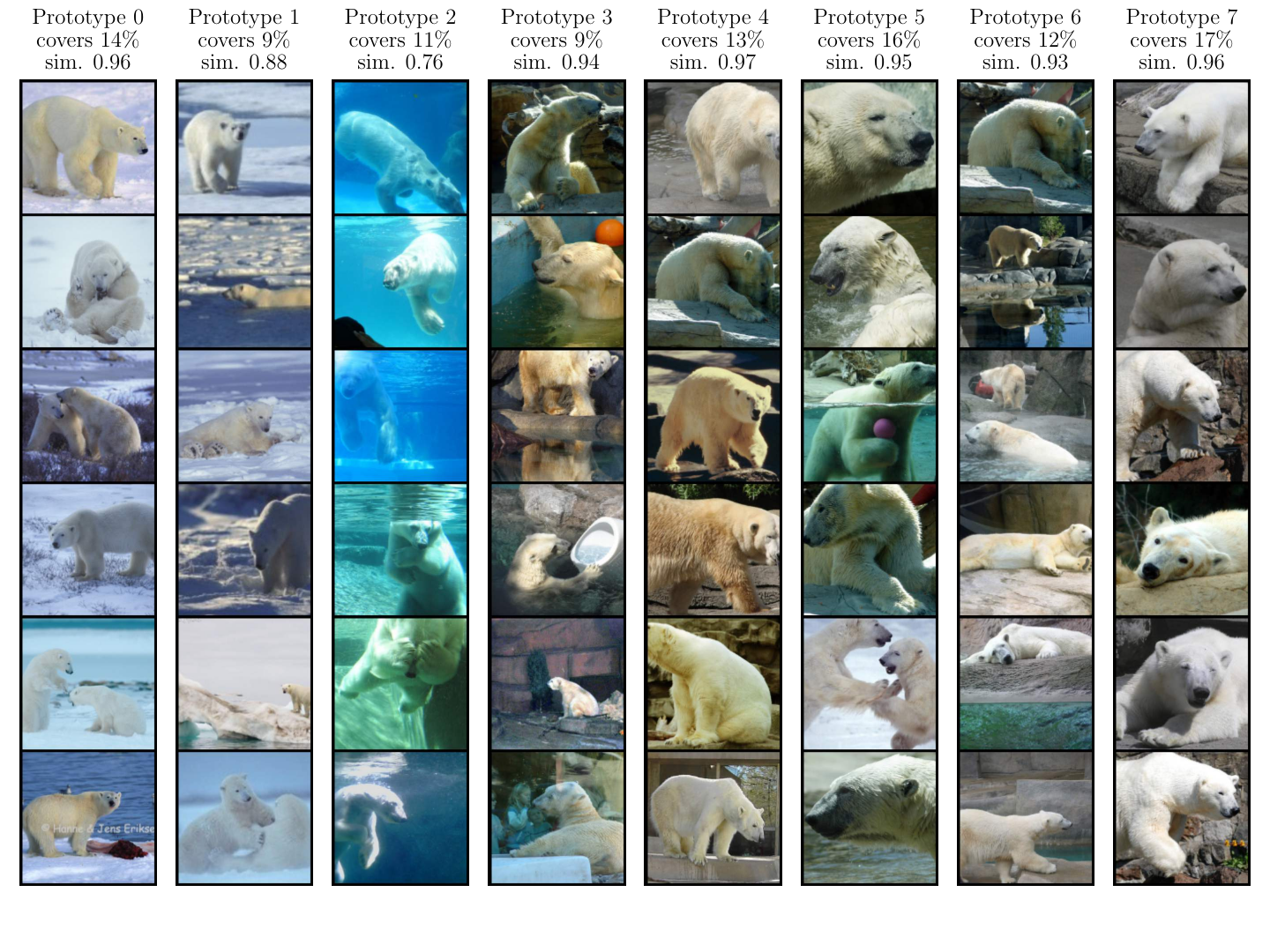}
                    \captionof{figure}{
                    Understanding model sub-strategies using (eight) prototypes.
                    We provide information on how many samples a prototype ``covers'', measured by the number of instances closest to the prototype in training set, and how similar they are to the overall mean in terms of cosine similarity.
                    (\emph{left}): For the ImageNet class ``hen'' (EfficientNet-b0, last \texttt{features} block), the model has learned to perceive hen of different color, \eg, brown, white and black.
                    (\emph{right}): For the ImageNet class ``ice bear'' (VGG-16, layer \texttt{features.28}), the model has learned to perceive ice bears in different environments, \eg, on ice, in water or with gray background.
                    }
                    \label{fig:app:add_proto:data_quality:sub_strategies}
        \end{figure*}

\begin{figure*}[t] 
                \centering
                    \includegraphics[width=0.49\linewidth]{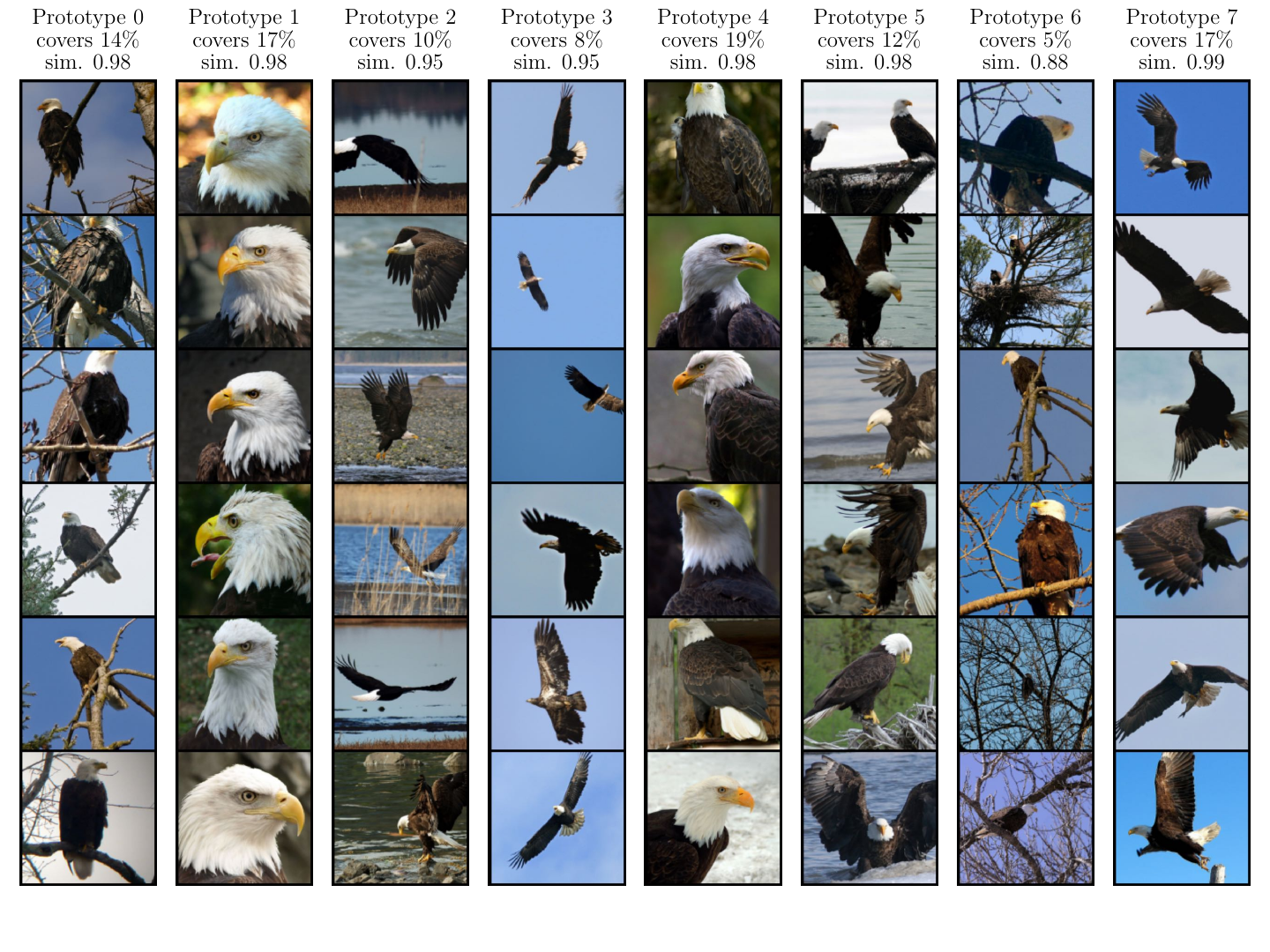}
                    \includegraphics[width=0.49\linewidth]{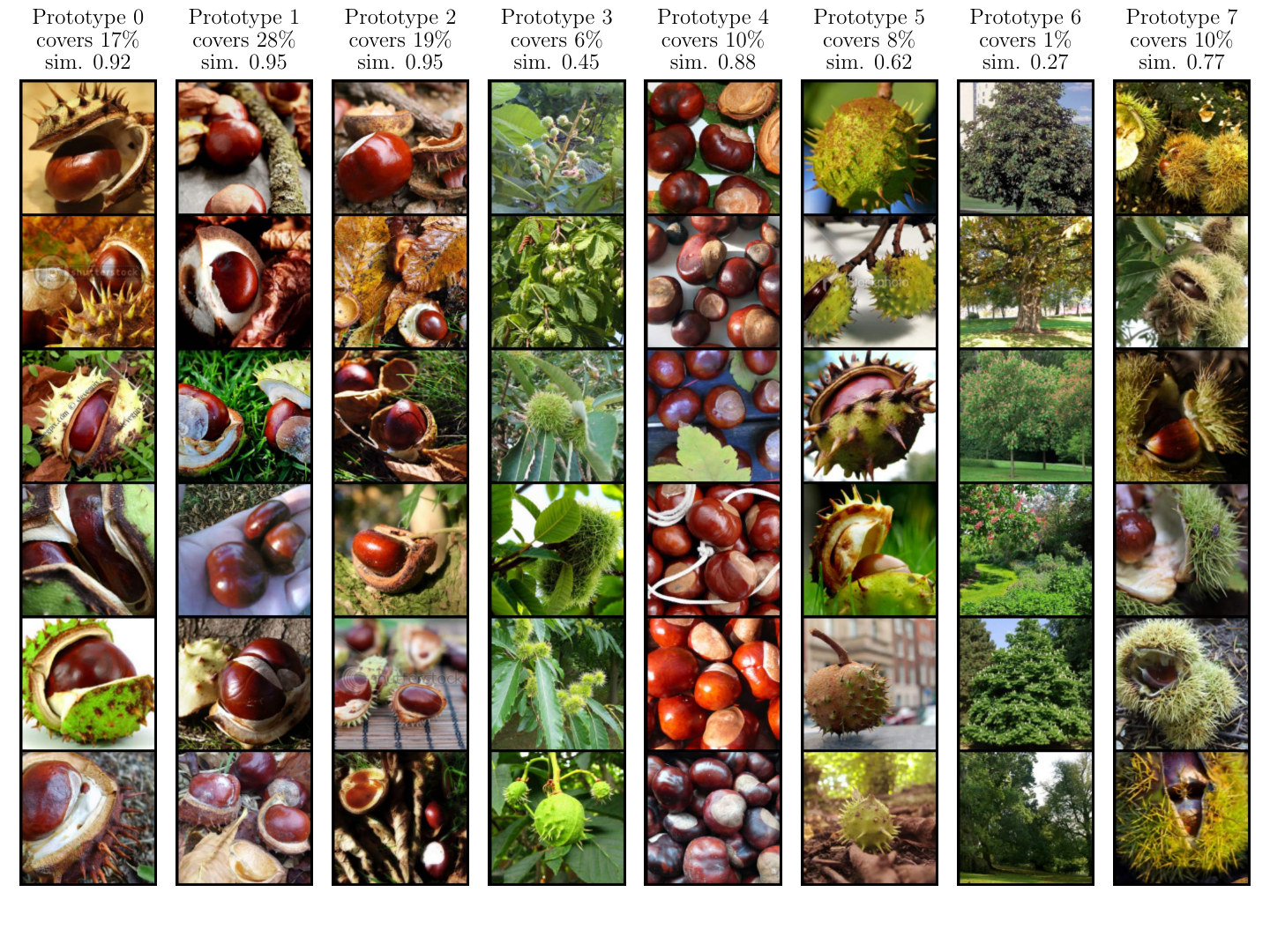}
                    \captionof{figure}{
                    Understanding model sub-strategies using (eight) prototypes.
                    We provide information on how many samples a prototype ``covers'', measured by the number of instances closest to the prototype in the train set, and how similar they are to the overall mean in terms of cosine similarity.
                    (\emph{left}): For the ImageNet class ``bald eagle'' (ResNet-18, last \texttt{BasicBlock} layer), the model has learned to perceive eagles in different environments, \eg, flying over water or sitting on branches.
                    (\emph{right}): For the ImageNet class ``buckeye'' (VGG-16, layer \texttt{features.28}), the model has learned to perceive buckeyes in different age stages, \eg, with green hull or partly visible.
                    Prototype 6 for ``buckeye'' shows not the fruit, but the whole tree, which raises the question whether corresponding samples (not showing the fruit) should be excluded from the training data set. Alternatively, 
                    predicted samples close to the prototype could be labeled with a ``warning'' note.
                    }
                    \label{fig:app:add_proto:data_quality:sub_strategies2}
        \end{figure*}

    We begin with four examples on how prototypes reveal sub-populations in the data (corresponding to learned sub-strategies) in Figures~\ref{fig:app:add_proto:data_quality:sub_strategies} and \ref{fig:app:add_proto:data_quality:sub_strategies2}.
    Here we present for the ImageNet classes ``hen'', ``ice bear'', ``bald eagle'' and ``buckeye'' eight prototypes of different models, with six example images shown corresponding to data points that are closest to a prototype.
    For the class ``hen'', the model (EfficientNet-b0, last \texttt{features} block) has learned to distinguish hen of different color, \eg, brown, white and black.
    Regarding the ``ice bear'' class, the model (VGG-16, layer \texttt{features.28}) has learned to perceive ice bears in different environments, \eg, on ice, in water or with gray background.
    For the ``bald eagle'' class, the model (ResNet-18, last \texttt{BasicBlock} layer) has learned to differentiate between eagles in various environments, \eg, flying over water or sitting on branches.
    Lastly,
    regarding the class of ``buckeye'' (chestnut), the model (VGG-16, layer \texttt{features.28}) has learned to perceive buckeyes in different age stages (including the full tree form), \eg, with green hull or partly visible.
    In the figures, we provide additional information on how many samples a prototype ``covers'', measured by the number of instances closest to the prototype in the training set, and how similar they are to the overall mean (concept relevance vector) in terms of cosine similarity.
    
    \subsubsection{Identifying Mislabeled Instances via Prototypes} 
\begin{figure*}[t] 
            \centering
                \includegraphics[width=0.56\linewidth]{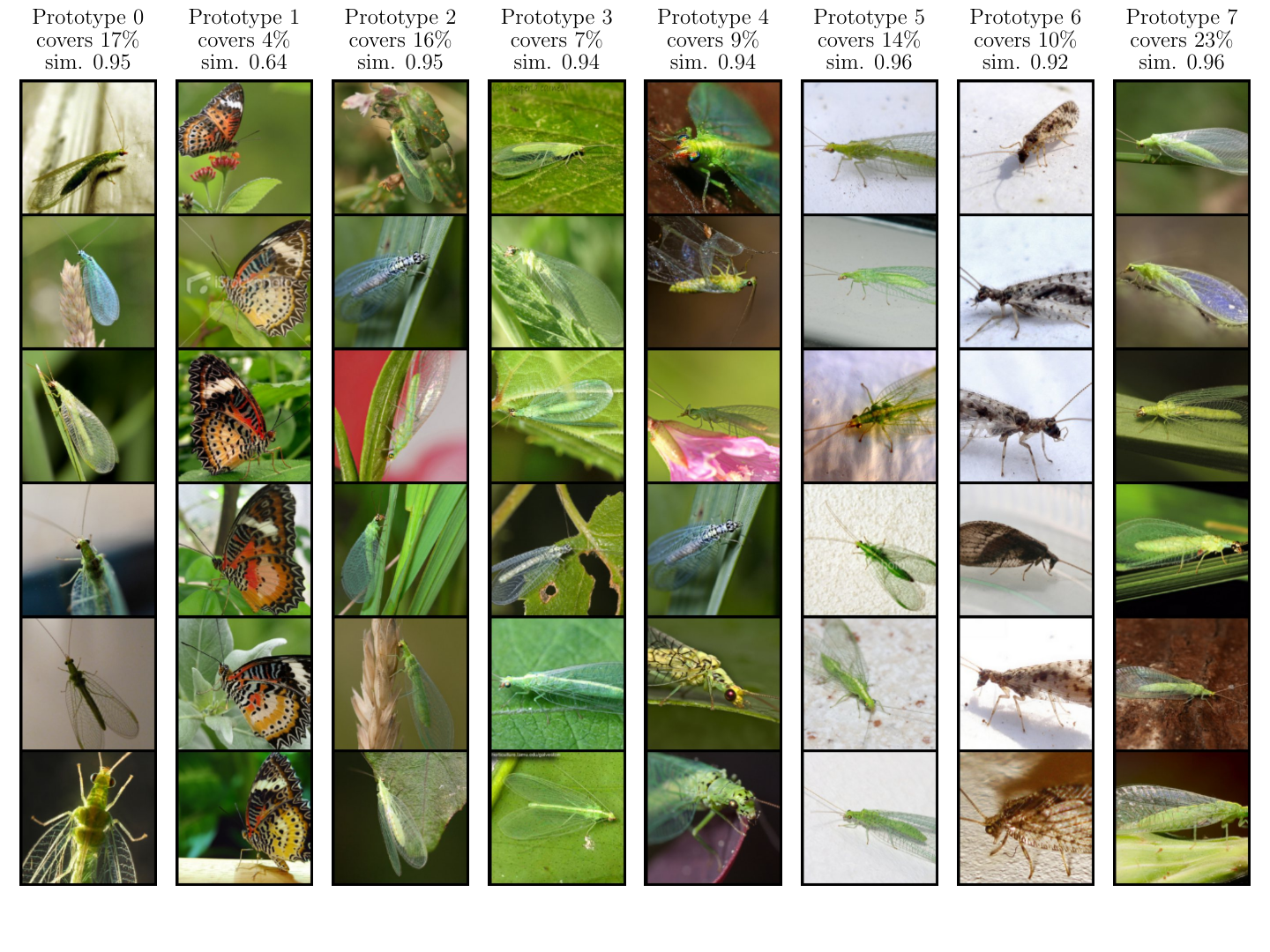}
                \includegraphics[width=0.43\linewidth]{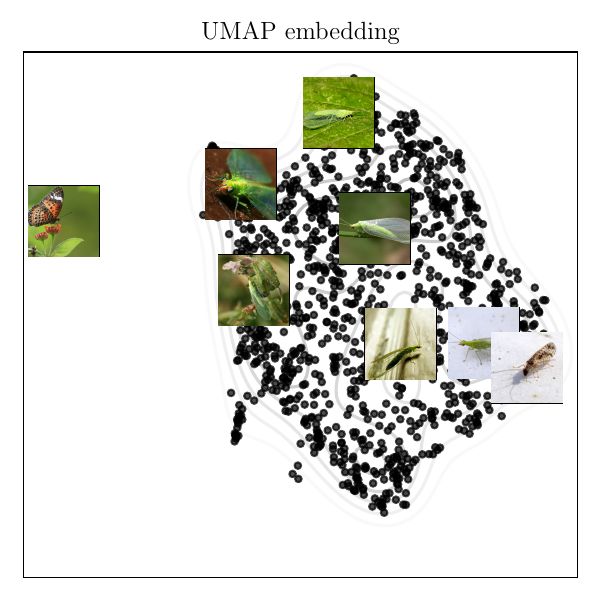}
                \includegraphics[width=0.56\linewidth]{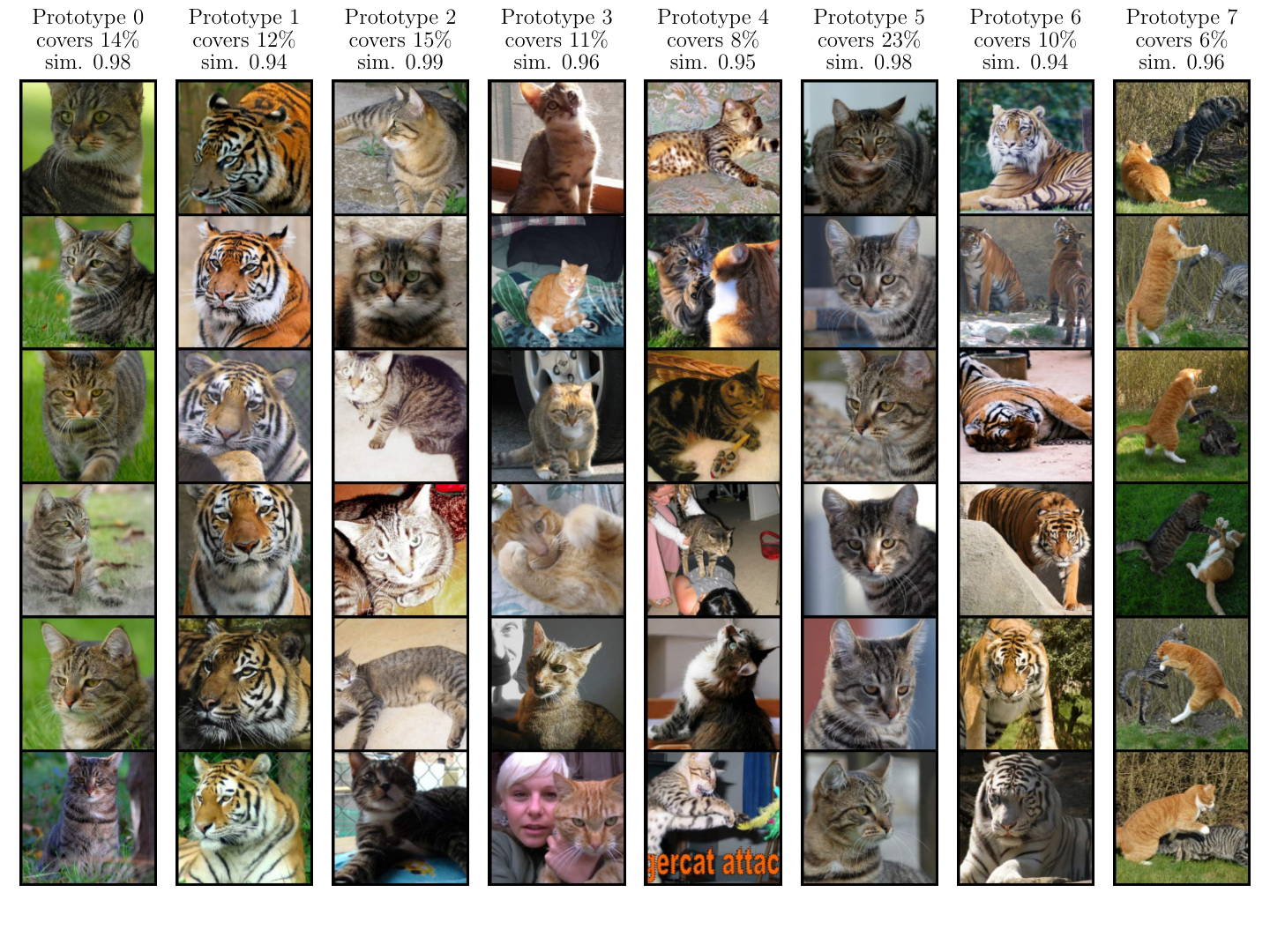}
                \includegraphics[width=0.43\linewidth]{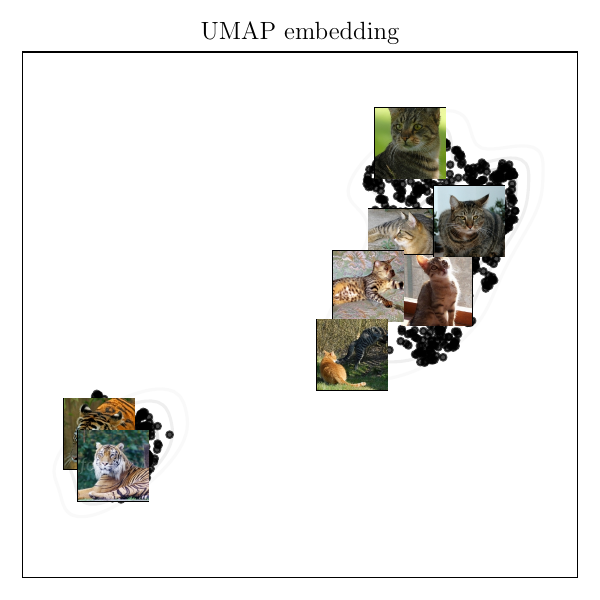}
                \captionof{figure}{
                Revealing wrong object samples in the ImageNet dataset using (eight) prototypes.
                Besides example images for each prototype, we also show a UMAP embedding.
                We further provide information on how many samples a prototype ``covers'', measured by the number of instances closest to the prototype in training set, and how similar they are to the overall mean in terms of cosine similarity.
                    (\emph{top}): For the ImageNet class ``lacewing'' (VGG-16, layer \texttt{features.28}), there are also samples of Leopard Lacewing butterflies in the training data.
                    (\emph{bottom}): For the ImageNet class ``tiger cat'' (ResNet-18, last \texttt{BasicBlock} layer), there are also samples of tigers in the training data.
                }
                \label{fig:app:add_proto:data_quality:wrong_labels:lacewing_tiger_cat}
    \end{figure*}

\begin{figure*}[t] 
            \centering
                \includegraphics[width=0.56\linewidth]{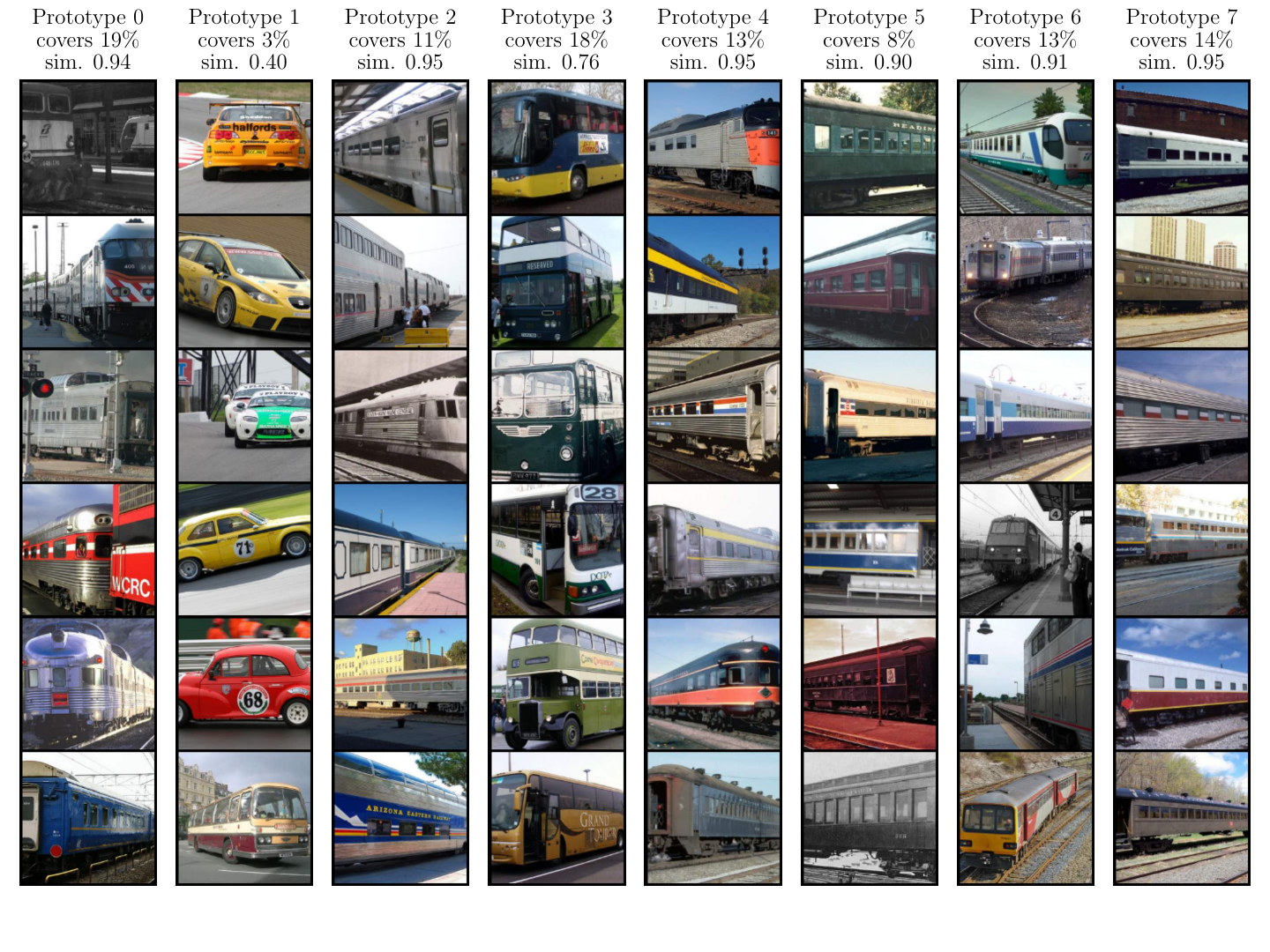}
                \includegraphics[width=0.43\linewidth]{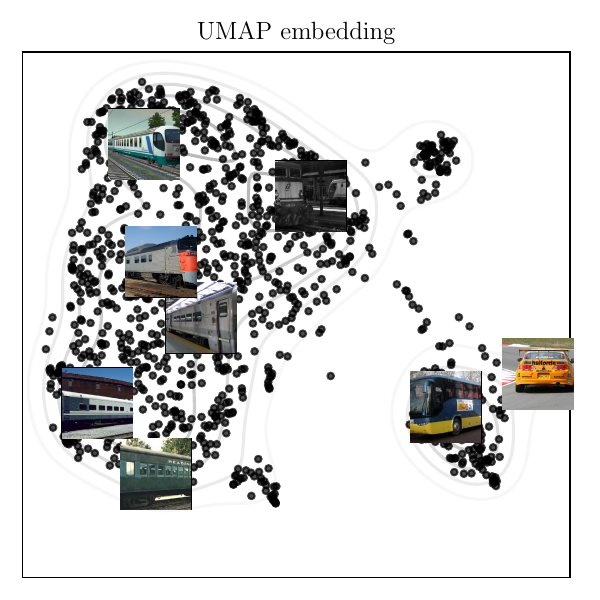}
                \includegraphics[width=0.56\linewidth]{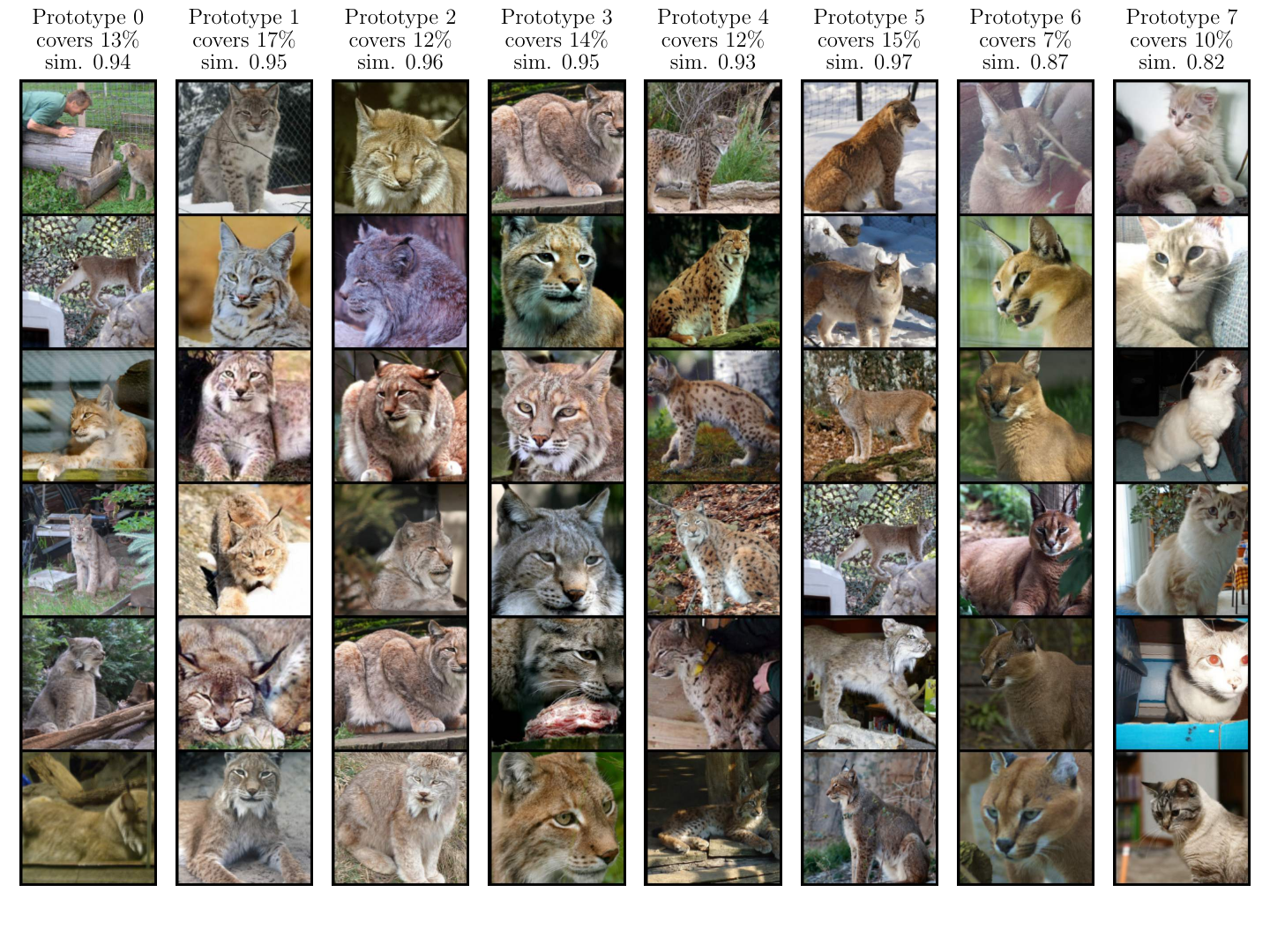}
                \includegraphics[width=0.43\linewidth]{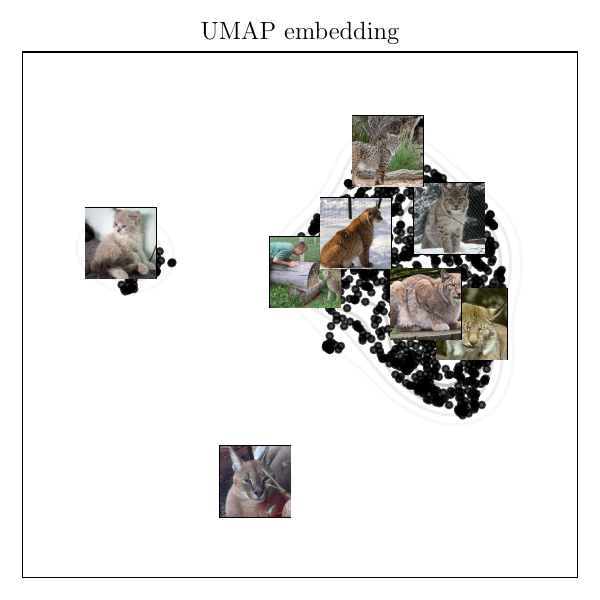}
                \captionof{figure}{
                Revealing wrong object samples in the ImageNet dataset using (eight) prototypes.
                We further provide information on how many samples a prototype ``covers'', measured by the number of instances closest to the prototype in training set, and how similar they are to the overall mean in terms of cosine similarity.
                Besides example images for each prototype, we also show a UMAP embedding.
                    (\emph{top}): For the ImageNet class ``passenger (railroad) car'' (VGG-16, layer \texttt{features.28}), there are also samples of buses and cars in the training data.
                    (\emph{bottom}): For the ImageNet class ``lynx, catamount'' (EfficientNet-b0, last \texttt{features} block), there are also samples of  Blue Lynx Ragdoll cats (prototype 7) in the training data. Note the distinct cluster for catamounts (prototype 6)  compared to lynx.
                }
                \label{fig:app:add_proto:data_quality:wrong_labels:passenger_lynx}
    \end{figure*}

    By studying prototypes,
    we found several classes, where wrong objects are included unintendedly, thus receiving a wrong label.
    As ImageNet is a collection of images generated through keyword search (the class labels) of image databases,
    ambiguous class labels might result in the retrieval of unwanted or wrong images.
    For example, we found instances where pictures of  Leopard Lacewing butterflies were mistakenly assigned to the ``lacewing'' class, a completely different insect species, as shown in Figure~\ref{fig:app:add_proto:data_quality:wrong_labels:lacewing_tiger_cat} (\emph{top}).
    Similarly,
    we found tigers in the ``Tiger Cat'' class, as depicted in Figure~\ref{fig:app:add_proto:data_quality:wrong_labels:lacewing_tiger_cat} (\emph{bottom}).
    More examples are shown in Figure~\ref{fig:app:add_proto:data_quality:wrong_labels:passenger_lynx} where cars and buses are included in the ``passenger (railroad) car'' class, and 
    Blue Lynx Ragdoll cats for the ``lynx, catamount'' class.
    In these two figures,
    we show eight prototypes for VGG, ResNet and EfficientNet models, with an additional UMAP~\cite{mcinnes2018umap} embedding illustrating the distribution of prediction strategies, \ie, concept relevance vectors.
    

    \subsubsection{Spotting Correlating Features} 
\begin{figure*}[t] 
            \centering
                \includegraphics[width=0.95\linewidth]{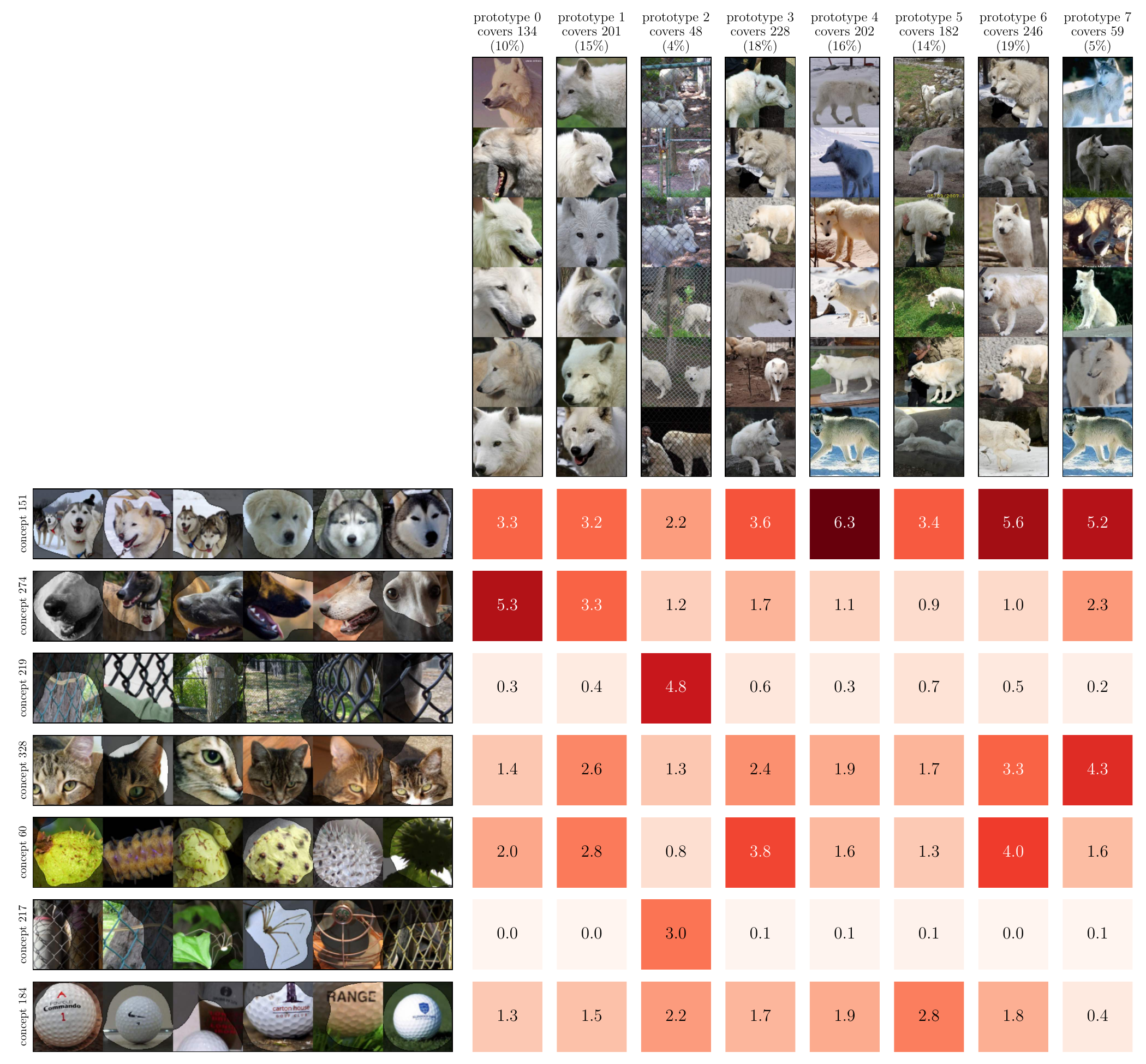}
                \captionof{figure}{
                Revealing correlating features in the ImageNet dataset using (eight) prototypes of a VGG-16 in layer \texttt{features.28}.
                For each prototype, we show relevant concepts and their corresponding relevance scores (\%).
                We further provide information on how many samples a prototype ``covers'', measured by the number of instances closest to the prototype in training set.
                For the class ``white wolf'', prototype 2 deviates from the other prototypes by a high relevance on ``fence'' concepts.
                }
                \label{fig:app:add_proto:data_quality:correlation:white_wolf}
    \end{figure*}

\begin{figure*}[t] 
            \centering
                \includegraphics[width=0.95\linewidth]{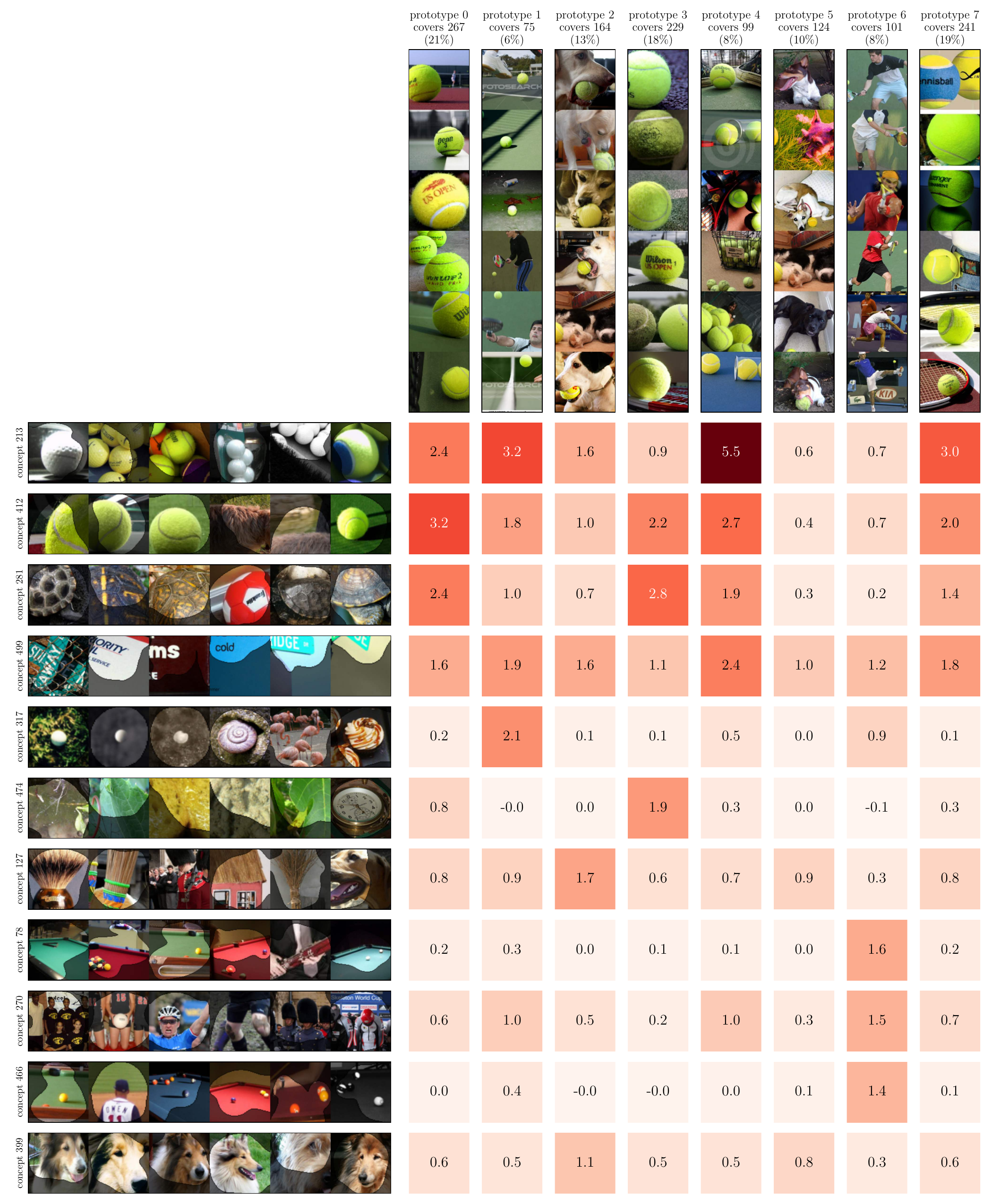}
                \captionof{figure}{
                Revealing correlating features in the ImageNet dataset using (eight) prototypes of a VGG-16 in layer \texttt{features.26}.
                For each prototype, we show relevant concepts and their corresponding relevance scores (\%).
                We further provide information on how many samples a prototype ``covers'', measured by the number of instances closest to the prototype in training set.
                For the class ``tennis ball'', prototype 2 (and 5) deviate from the other prototypes by relevances on ``dog'' concepts.
                Notably prototype 6 also deviates, with a focus on concepts related to persons.
                }
                \label{fig:app:add_proto:data_quality:correlation:tennis}
    \end{figure*}

\begin{figure*}[t] 
            \centering
                \includegraphics[width=0.95\linewidth]{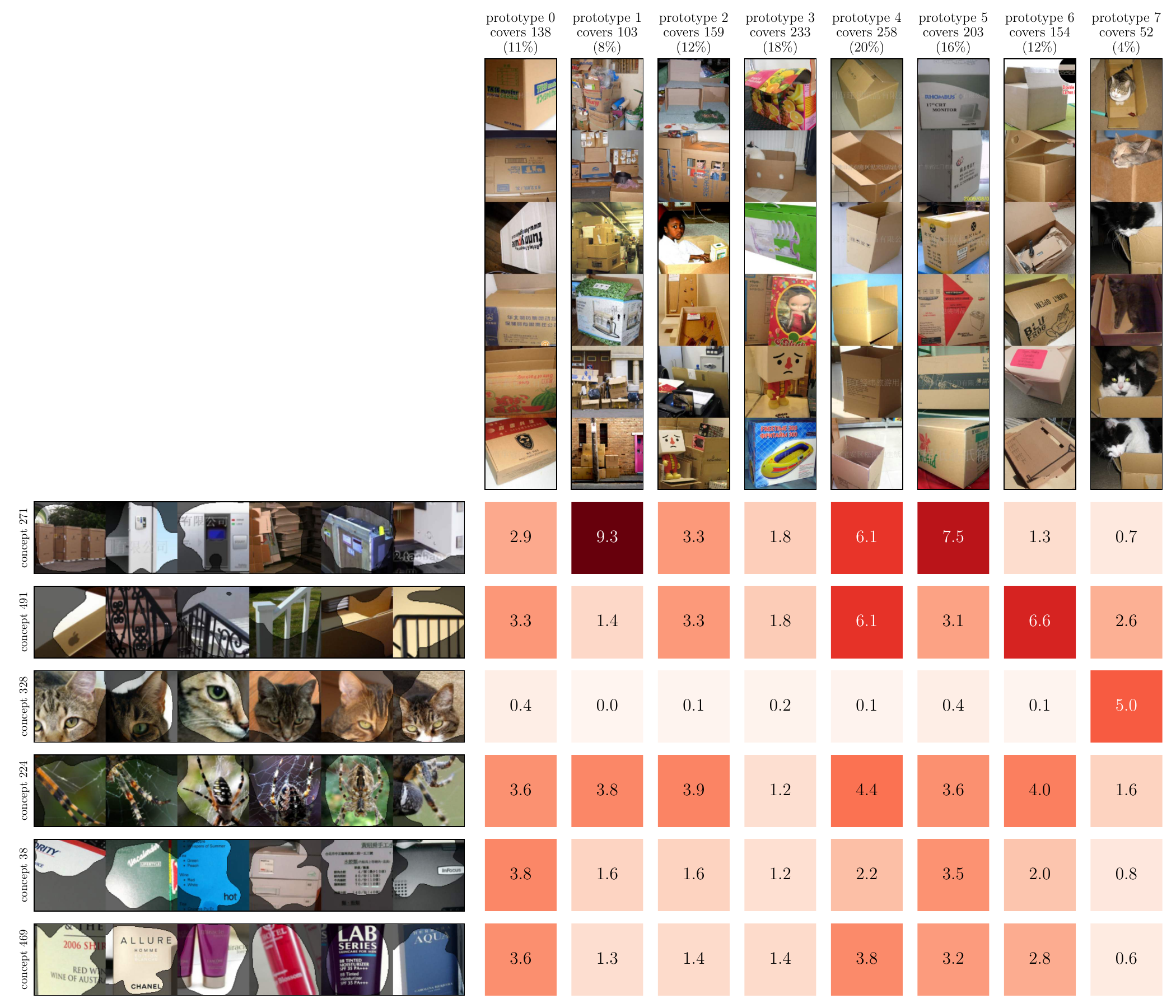}
                \captionof{figure}{
                Revealing correlating features in the ImageNet dataset using (eight) prototypes of a VGG-16 in layer \texttt{features.28}.
                For each prototype, we show relevant concepts and their corresponding relevance scores (\%).
                We further provide information on how many samples a prototype ``covers'', measured by the number of instances closest to the prototype in training set.
                For the class ``carton'', prototype 7 deviates from the other prototypes by a high relevance on ``cat'' concepts.
                Note, that concept 328 refers to spider webs (thin lines), which is likely to be triggered for the writings on the outside of cartons, or the overlaid watermark seen in, \eg, prototype 4. 
                }
                \label{fig:app:add_proto:data_quality:correlation:carton}
    \end{figure*}

\begin{figure*}[t] 
            \centering
                \includegraphics[width=0.95\linewidth]{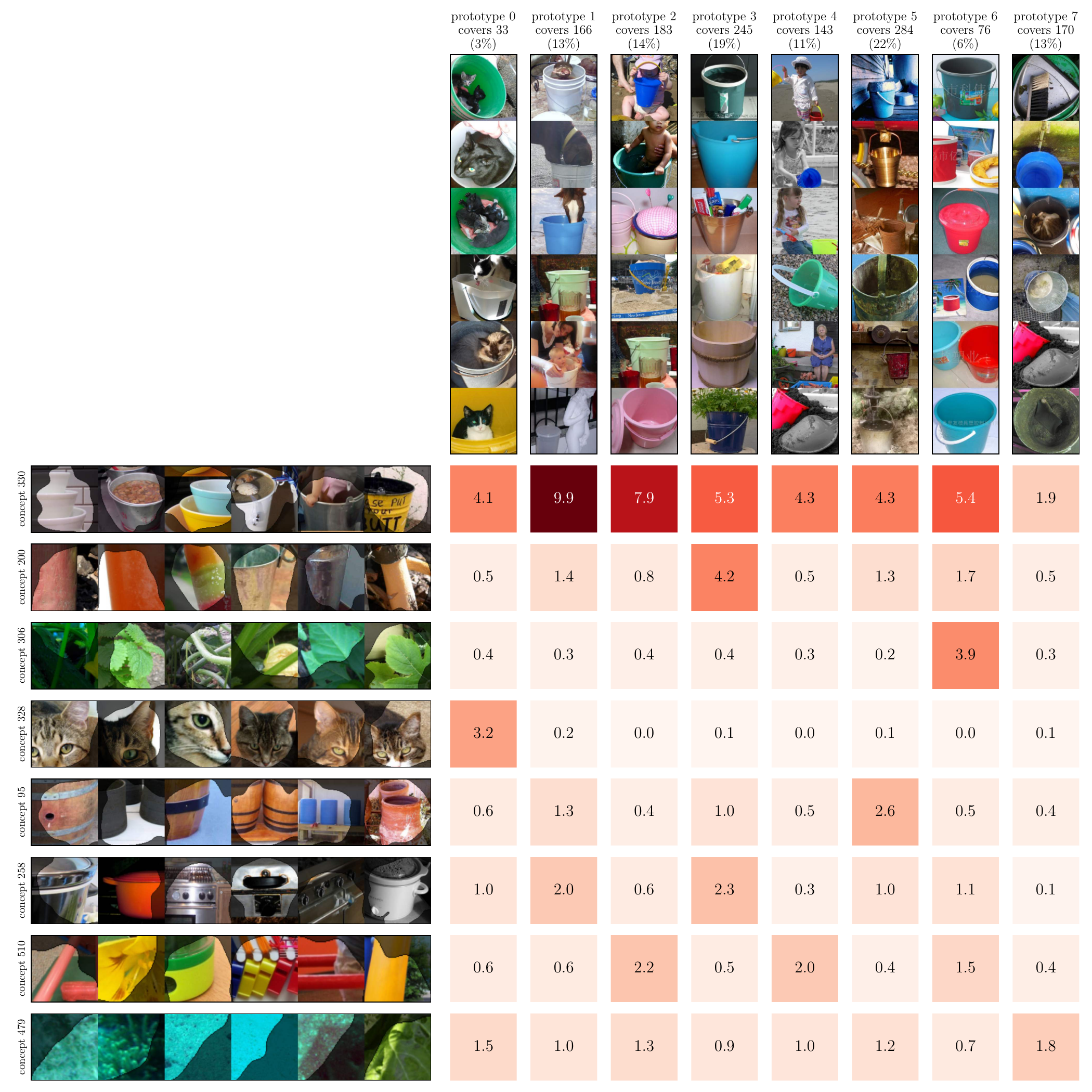}
                \captionof{figure}{
                Revealing correlating features in the ImageNet dataset using (eight) prototypes of a VGG-16 in layer \texttt{features.28}.
                For each prototype, we show relevant concepts and their corresponding relevance scores (\%).
                We further provide information on how many samples a prototype ``covers'', measured by the number of instances closest to the prototype in training set.
                For the class ``bucket'', prototype 0 deviates from the other prototypes by a high relevance on ``cat'' concepts.
                Further note prototype 6 and the Chinese watermark artifact on the example images.
                }
                \label{fig:app:add_proto:data_quality:correlation:bucket}
    \end{figure*}

    Another noteworthy set of data artifacts are correlating features found within the ImageNet training dataset. 
    These features include white wolfs behind fences (Figure~\ref{fig:app:add_proto:data_quality:correlation:white_wolf}),
    dogs with tennis balls (Figure~\ref{fig:app:add_proto:data_quality:correlation:tennis}), and 
    cats in cartons or buckets (Figures~\ref{fig:app:add_proto:data_quality:correlation:carton} and~\ref{fig:app:add_proto:data_quality:correlation:bucket}, respectively). 
    By being able to understand the concepts that are relevant (characteristic) for each prototype,
    we can further validate our observations.
    We thus show for all prototypes the concept relevance scores for the set of most relevant concepts (according to a VGG-16 in layer \texttt{features.28}).
    Concretely,
    the set of concepts is given by retrieving the top-2 concepts for each prototype. 
    Further,
    concepts are visualized using reference samples as given by the RelMax technique~\cite{achtibat2023attribution}.
    With RelMax, we visualize concepts with samples, where the concept was most relevant \wrt the prediction outcome, illustrating how the concept is \emph{used} by the model.
    To improve clarity,
    each reference sample is cropped to the for the concept relevant part, and all irrelevant parts are masked by a semi-transparent black mask.


    \subsubsection{Diagnosing for Poor Data Quality} 

\begin{figure*}[t] 
            \centering
                \includegraphics[width=0.95\linewidth]{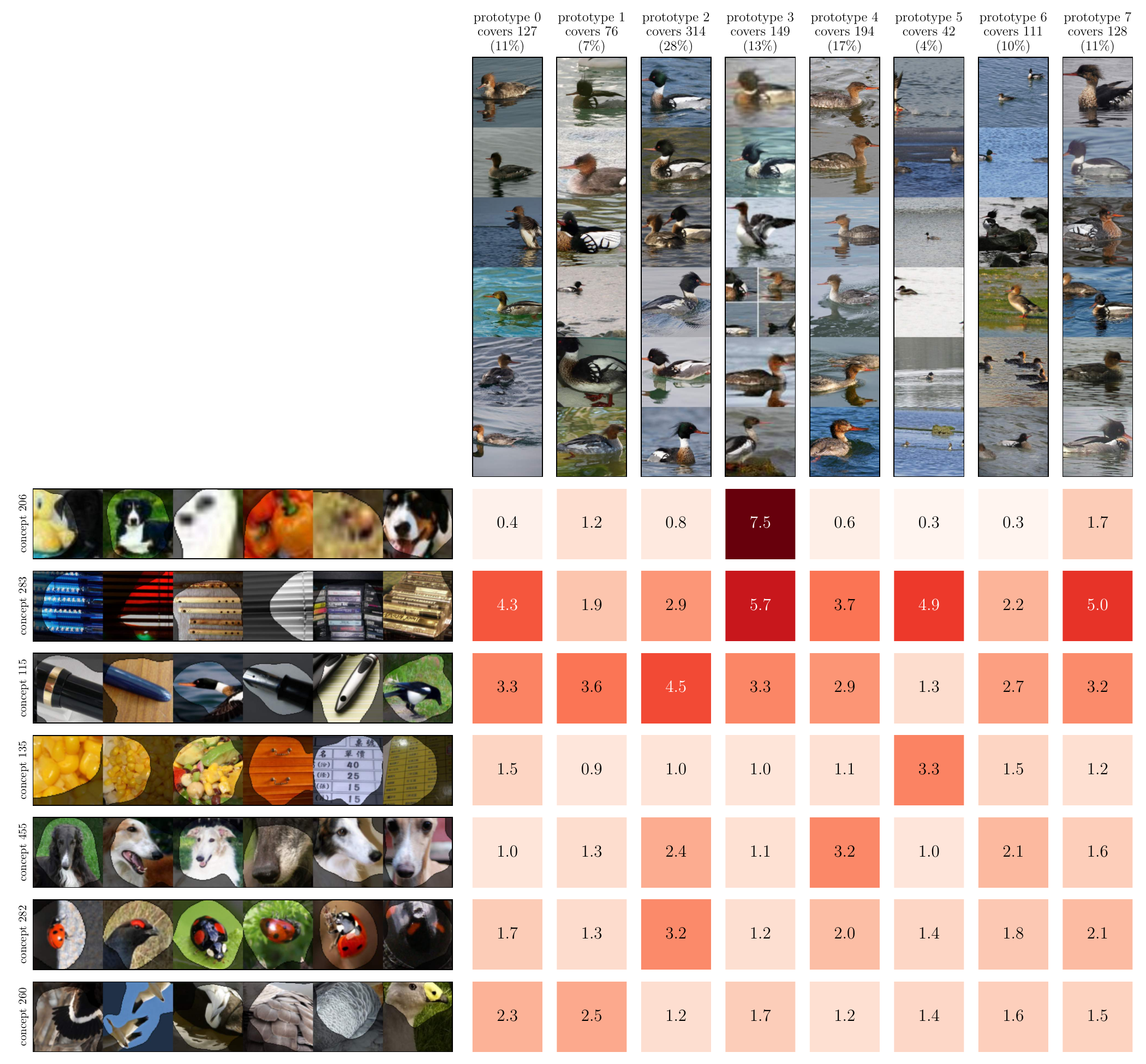}
                \captionof{figure}{
                Revealing correlating features in the ImageNet dataset using (eight) prototypes of a VGG-16 in layer \texttt{features.28}.
                For each prototype, we show relevant concepts and their corresponding relevance scores (\%).
                We further provide information on how many samples a prototype ``covers'', \ie, are closest to the prototype in training set.
                For the class ``red-breasted merganser'', prototype 3 deviates from the other prototypes by a high relevance on ``blur'' concepts.
                }
                \label{fig:app:add_proto:data_quality:correlation:merganser}
    \end{figure*}

\begin{figure*}[t] 
            \centering
                \includegraphics[width=0.95\linewidth]{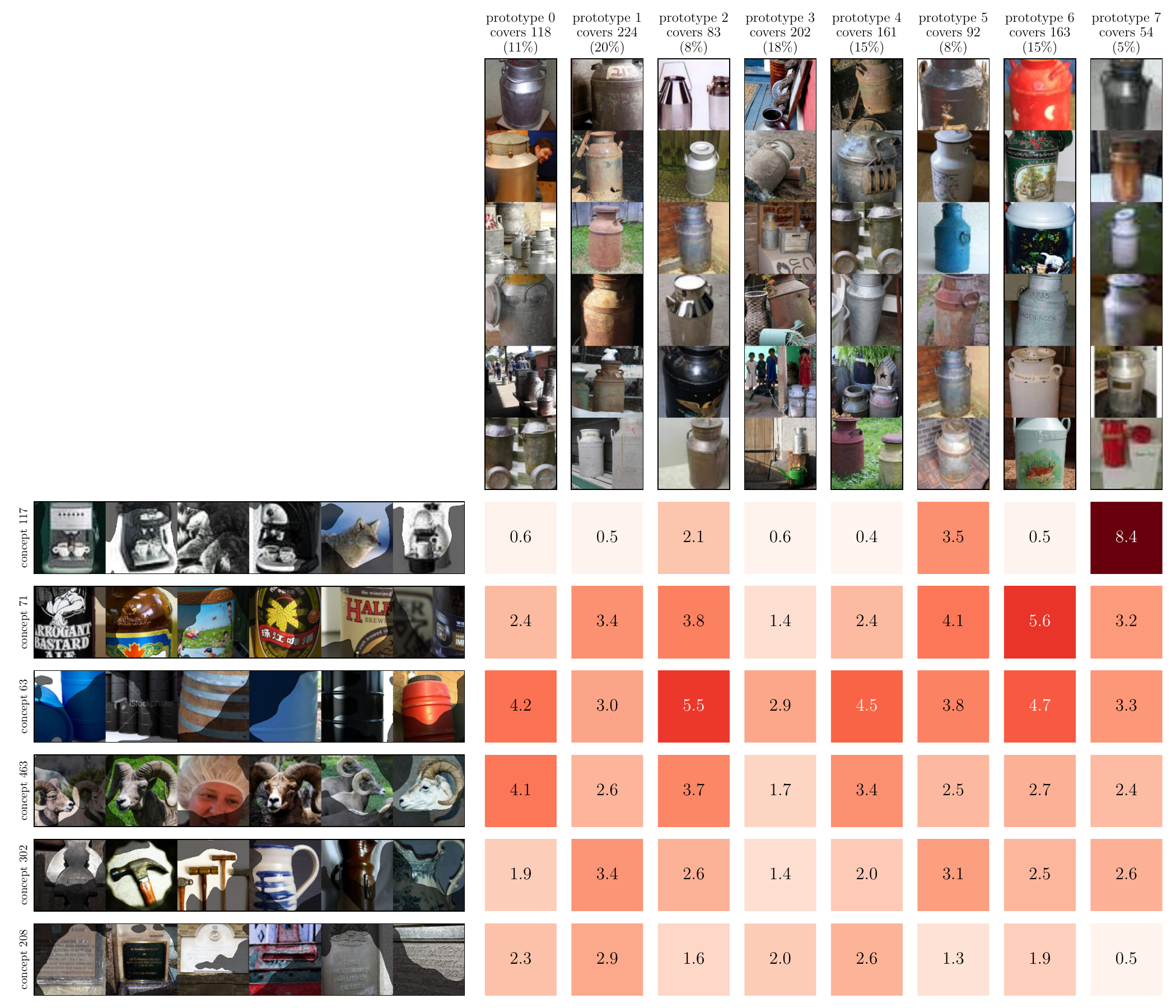}
                \captionof{figure}{
                Revealing data quality issues in the ImageNet dataset using (eight) prototypes of a ResNet-18 in the last \texttt{BasicBlock} layer.
                For each prototype, we show relevant concepts and their corresponding relevance scores (\%).
                We further provide information on how many samples a prototype ``covers'', \ie, are closest to the prototype in training set. 
                For the class ``milk can'', prototypes 5 and 7 deviate from the other prototypes by a high relevance on a ``blur'' concept (concept 117).
                }
                \label{fig:app:add_proto:data_quality:correlation:milk}
    \end{figure*}

\begin{figure*}[t] 
            \centering
                \includegraphics[width=0.95\linewidth]{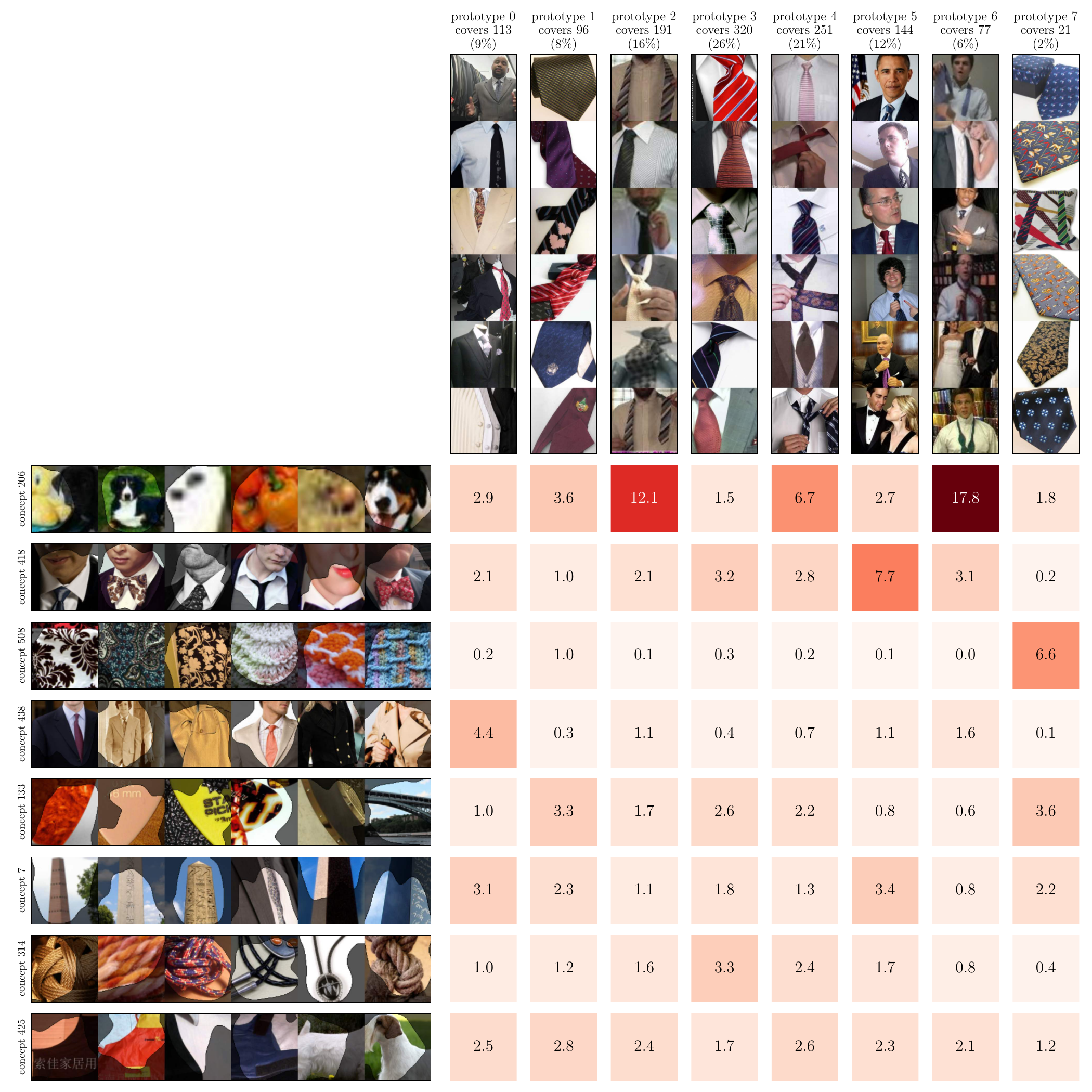}
                \captionof{figure}{
                Revealing data quality issues in the ImageNet dataset using (eight) prototypes of a VGG-16 in layer \texttt{features.28}.
                For each prototype, we show relevant concepts and their corresponding relevance scores (\%).
                We further provide information on how many samples a prototype ``covers'', \ie, are closest to the prototype in training set.
                For the class ``Windsor tie'', prototype 2 and 6 deviate from the other prototypes by a high relevance on ``blur'' concepts.
                }
                \label{fig:app:add_proto:data_quality:correlation:tie}
    \end{figure*}

        By studying prototypes and their characteristic concepts we further were able to reveal issues of poor data quality.
        This includes, \eg, 
        prototypes where a ``blur'' artifact is relevant, as found in the ImageNet classes of ``red-breasted merganser'' (birds), ``milk can'' and ``Windsor tie'', 
        as shown in Figures~\ref{fig:app:add_proto:data_quality:correlation:merganser}, \ref{fig:app:add_proto:data_quality:correlation:milk} and \ref{fig:app:add_proto:data_quality:correlation:tie} for VGG-16 and ResNet models.
        This dedicated ``blur'' concepts result from a large set of training images having poor resolution.
        
\begin{figure*}[t] 
            \centering
                \includegraphics[width=0.95\linewidth]{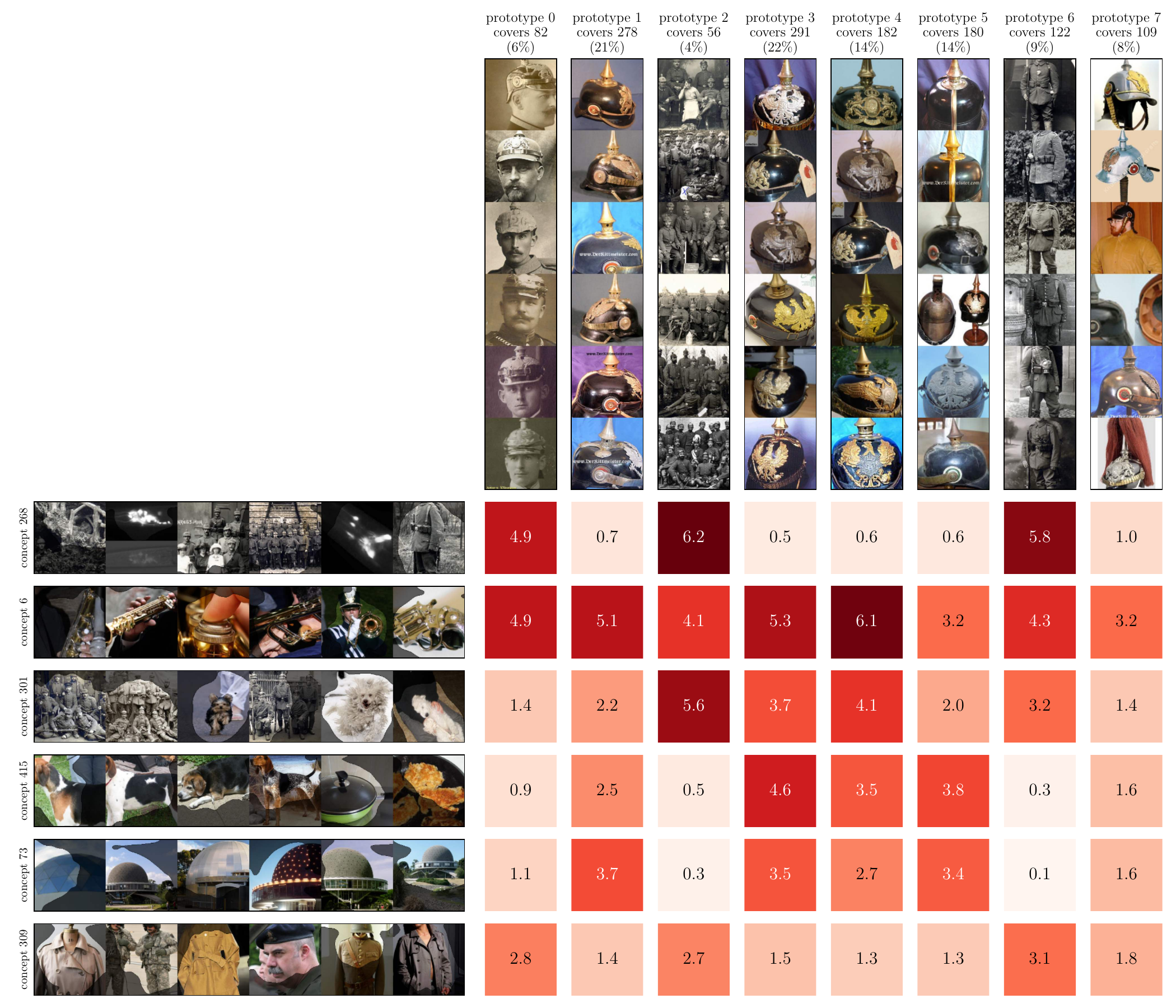}
                \captionof{figure}{
                Revealing data quality issues in the ImageNet dataset using (eight) prototypes of a ResNet-18 in the last \texttt{BasicBlock} layer.
                For each prototype, we show relevant concepts and their corresponding relevance scores (\%).
                We further provide information on how many samples a prototype ``covers'', \ie, are closest to the prototype in training set.
                For the class ``pickelhaube'', prototype 6 deviates from the norm by a high relevance on ``uniform'' concepts, instead of the pickelhaube.
                Notably,
                concept 73 has also (almost) no relevance, which corresponds to a dome-like form (corresponding to the shape of the pickelhaube).
                This is also sensible as in the example images, the heads are cropped out --- due to data augmentation. 
                Also prototype 2, which corresponds to group pictures, is associated with high relevances on uniform concepts.
                It is further to note that prototypes 0, 2 and 6 show a high relevances on a ``gray color'' concept, indicating that the model has learned to detect gray-scale pictures.
                }
                \label{fig:app:add_proto:data_quality:correlation:pickelhaube} 
    \end{figure*}

        Further,
        in the class of ``pickelhaube'' we found that the model has learned to detect the class object not only because of the pickelhaube, but also in the absence of the object because of the uniform or face features, as shown in Figure~\ref{fig:app:add_proto:data_quality:correlation:pickelhaube}.
        Here,
        prototype 6 captures military men in uniform where the head (and thus also the pickelhaube) is cropped out, leading the model to use also alternative features.
        This illustrates the effect data augmentation techniques (including, \eg, random cropping) can have on the model behavior.
    

\renewcommand\thefigure{D.\arabic{figure}}    
\renewcommand\thetable{D.\arabic{table}}
\setcounter{figure}{0}
\setcounter{table}{0}
\renewcommand\theequation{D.\arabic{equation}}
\setcounter{equation}{0}
    \clearpage
    \clearpage
    \section{Model (Prediction) Validation}
    \label{app:add_proto}
    In this section,
    we present additional examples and clarifications regarding the use of prototypes for model (prediction) validation.

    \subsection{Covariance Matrix Understanding}
    \label{app:add_proto:covar}
        A high likelihood of a prediction with concept relevance vector $\boldsymbol{\nu}$ belonging to prototype $i$ of class $k$ is given for a high probability density value $p_i^k(\boldsymbol{\nu})$, \ie,
        \begin{equation}
            \label{eq:app:add_proto:covar:pdf}
                p_i^k(\boldsymbol{\nu}) = \frac{1}{(2\pi)^\frac{n}{2} \det( \sig^k_i)^\frac{1}{2}}e^{-\frac{1}{2} \left(\boldsymbol{\nu} - \mean^k_i\right)^\top \left(\sig^k_i\right)^{-1} \left(\boldsymbol{\nu} - \mean^k_i\right)}\,.
        \end{equation}
        Here,
        the 
        probability density
        $p_i^k(\boldsymbol{\nu})$
        depends on the term
        \begin{equation}
            \label{eq:app:add_proto:covar:maha}
                \delta^k_i(\boldsymbol{\nu}) = \left(\boldsymbol{\nu} - \mean^k_i\right)^\top \left(\sig^k_i\right)^{-1} \left(\boldsymbol{\nu} - \mean^k_i\right)\,.
        \end{equation}

        For simplicity,
        we assume that we choose one prototype and one class, simplifying notation to
    
        \begin{equation}
            \label{eq:app:add_proto:covar:maha_simple}
                \delta = \left(\boldsymbol{\nu} - \mean\right)^\top \sig^{-1} \left(\boldsymbol{\nu} - \mean\right) = \boldsymbol\Delta^\top \sig^{-1} \boldsymbol\Delta
        \end{equation}
        with $\boldsymbol\Delta = \boldsymbol{\nu} - \mean$.
        Then,
        we have
        \begin{equation}
            \label{eq:app:add_proto:covar:maha_simple_extended}
                \delta = \Delta_1 \Sigma_{11}^{-1} \Delta_1 + \Delta_1 \Sigma_{12}^{-1} \Delta_2 + \dots + \Delta_m \Sigma_{mm}^{-1} \Delta_m\,.
        \end{equation}
        Here,
        contributions result from intra-concept deviations, \ie, $\Delta_i \Sigma_{ii}^{-1} \Delta_i \geq 0$ (as $\Sigma_{ii}^{-1} \geq 0$, because $\Sigma$ and $\Sigma^{-1}$ are positive semi-definite matrices),
        and inter-concept deviations $\Delta_i \Sigma_{ij}^{-1} \Delta_j$ (with $i \neq j$).
        Contrary to only examining the difference vector $\boldsymbol\Delta = \boldsymbol{\nu} - \mean$,
        $\delta$ also includes information from the covariance matrix and allows to investigate contributions from inter-concept deviations.

    \subsection{Increasing the Number of Prototypes}
    \label{app:add_proto:prot_number}
\begin{figure}[t]
            \centering
                \includegraphics[width=0.99\linewidth]{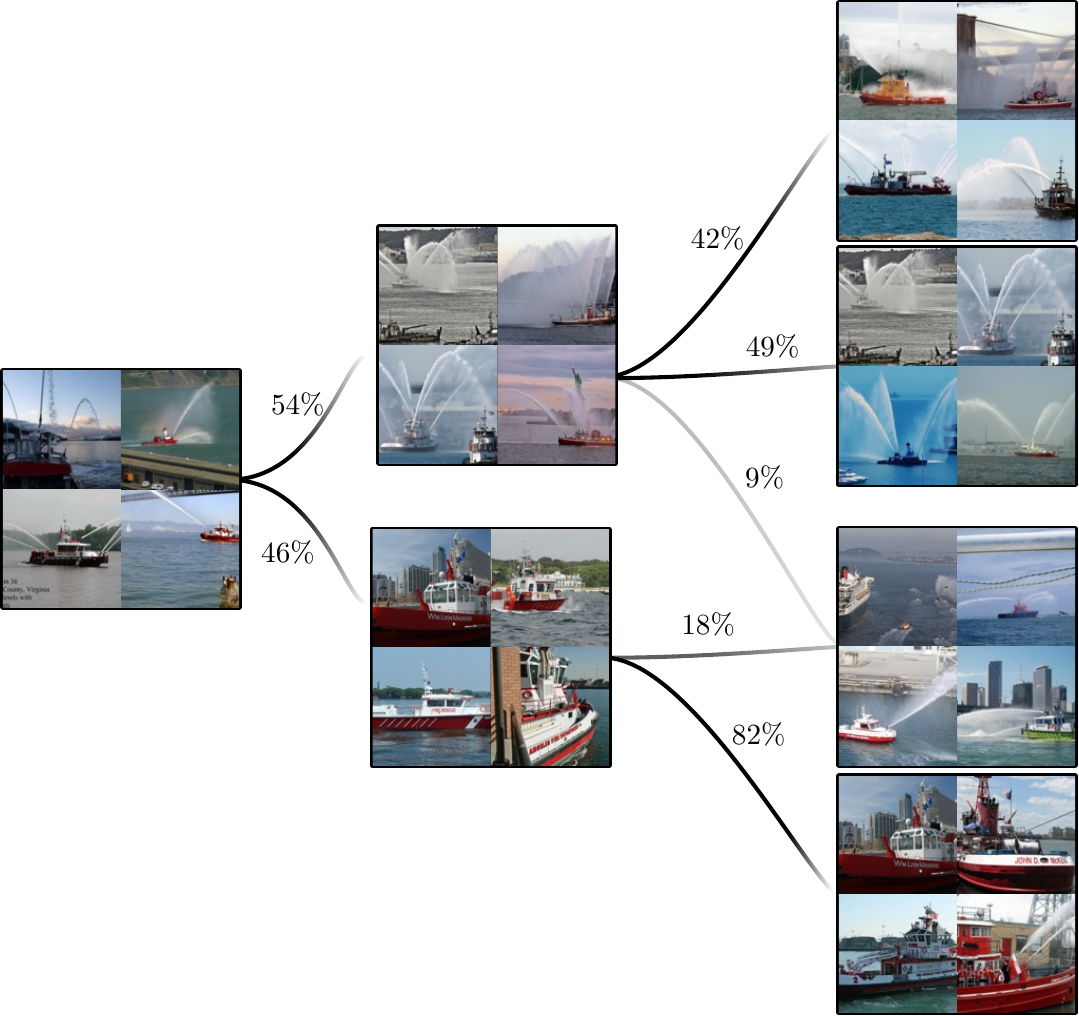}
                \captionof{figure}{
                Qualitative example for changing the number of prototypes to one (\emph{left}), two (\emph{middle}) and four (\emph{right}).
                We show the prototypes for the ImageNet class ``fireboat'' resulting for a VGG-16 model and layer \texttt{features.28}.
                Whereas the single prototype depicts fireboats spraying (little) water, 
                increasing the number leads to more distinct prototypes, \eg, fireboats with \emph{and} without water fountains of varying strength.
                Note that we also depict the amount of samples (closest to a prototype) that transfer from one prototype to the other.
                }
                \label{fig:app:prot_number:fireboat}
        \end{figure}

    We in the following provide qualitative examples of resulting prototypes when increasing their number (used to fit the \gls{gmm}).
    The first example is shown in Figure~\ref{fig:app:prot_number:fireboat},
    where we visualize the emerging prototypes for ImageNet class ``fireboat'' setting the prototype number to one, two and four.
    Whereas one prototype only visualizes fireboats spraying water,
    we reveal that the model has learned to differentiate between fireboats spraying and not spraying water by increasing prototype numbers.
    
\begin{figure}[t]
            \centering
                \includegraphics[width=0.99\linewidth]{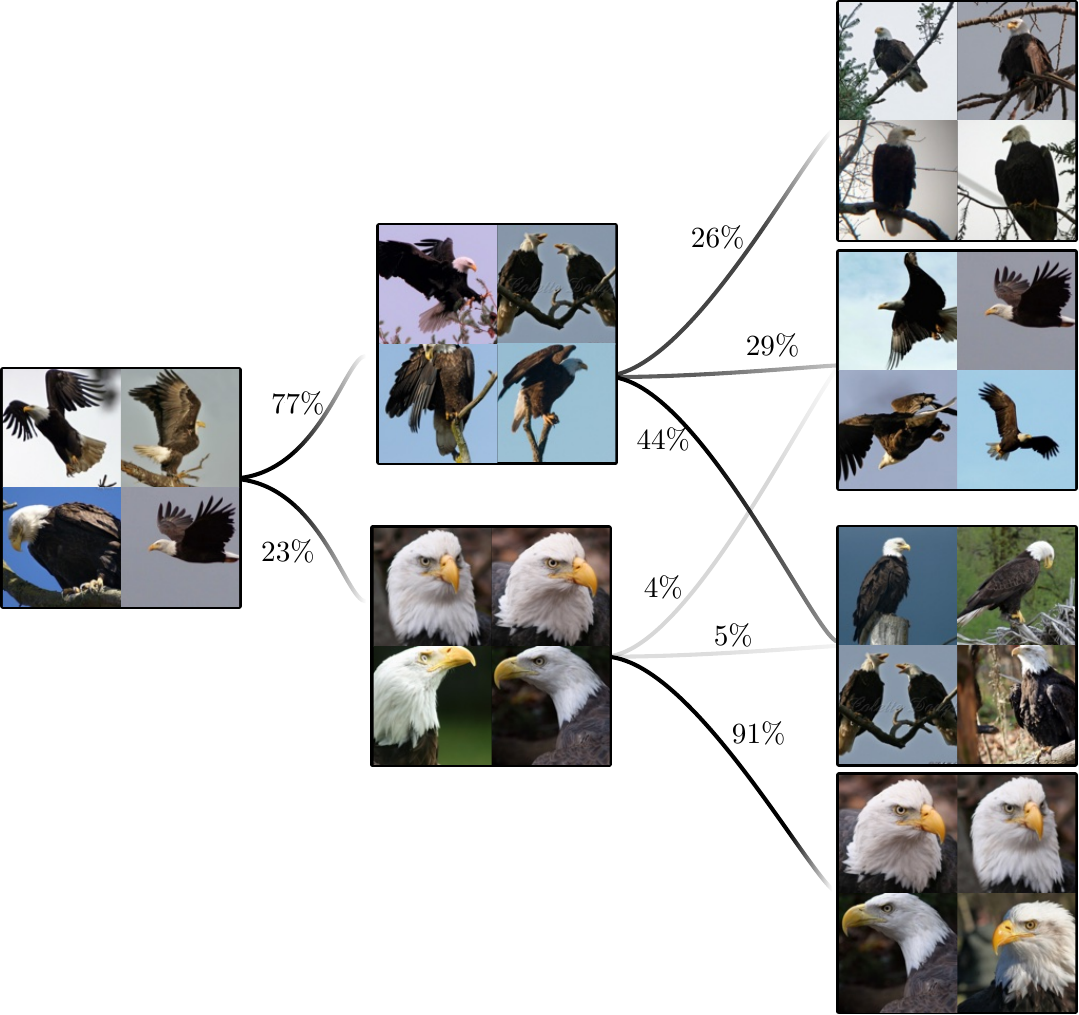}
                \captionof{figure}{
                Qualitative example for changing the number of prototypes to one (\emph{left}), two (\emph{middle}) and four (\emph{right}).
                We show the prototypes for the ImageNet class ``American eagle'' resulting for a VGG-16 model and layer \texttt{features.28}.
                Whereas the single prototype depicts eagles flying and sitting on branches, 
                increasing the number leads to more distinct prototypes, \eg, close-ups, flying and sitting.
                Note that we also depict the amount of samples (closest to a prototype) that transfer from one prototype to the other.
                }
                \label{fig:app:prot_number:eagle}
        \end{figure}

\begin{figure}[t]
            \centering
                \includegraphics[width=0.99\linewidth]{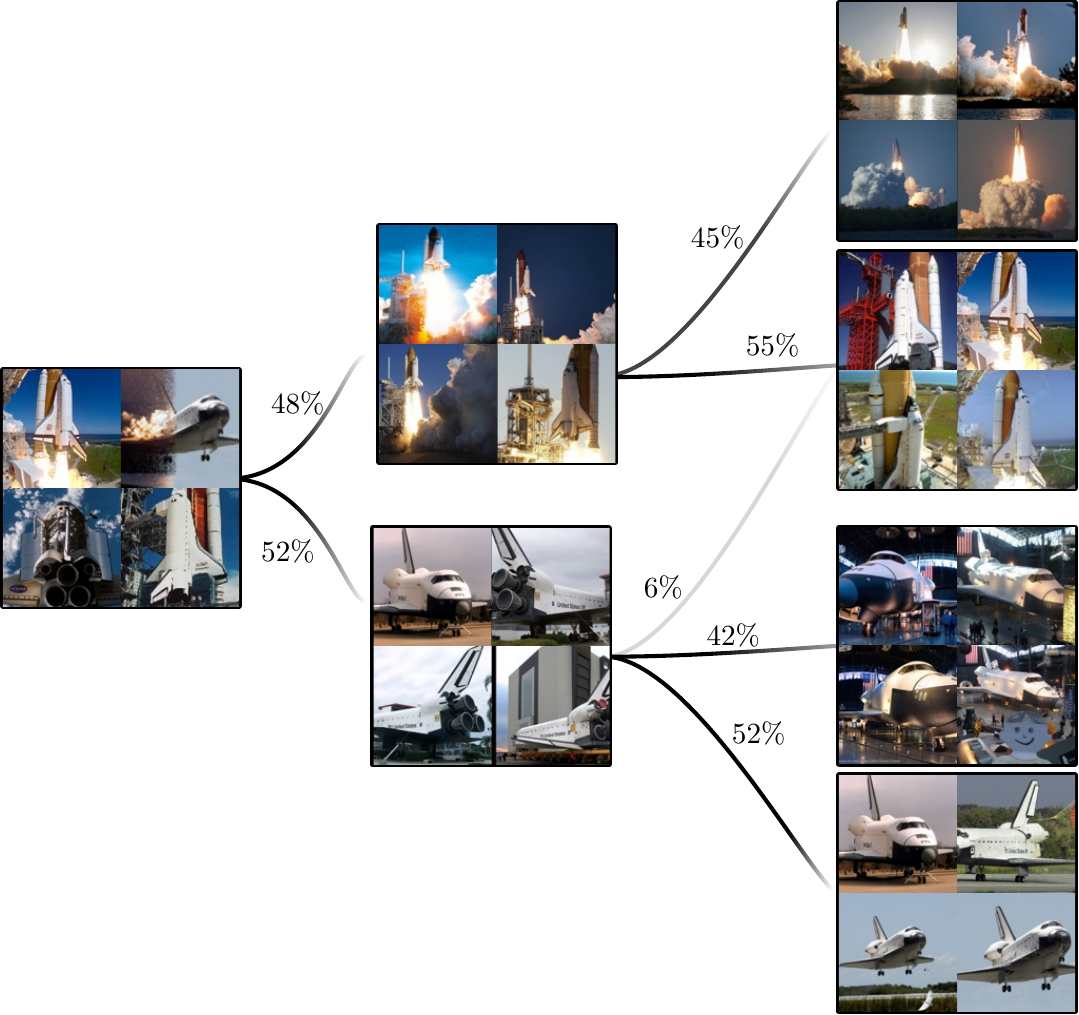}
                \captionof{figure}{
                Qualitative example for changing the number of prototypes to one (\emph{left}), two (\emph{middle}) and four (\emph{right}).
                We show the prototypes for the ImageNet class ``space shuttle'' resulting for a VGG-16 model and layer \texttt{features.28}.
                Whereas the single prototype depicts various positions of a space shuttle, 
                increasing the number leads to more distinct prototypes, \eg, launching with a tail of fire, standing in exhibitions, landing on the landing site.
                Note that we also depict the amount of samples (closest to a prototype) that transfer from one prototype to the other.
                }
                \label{fig:app:prot_number:spaceshuttle}
        \end{figure}

    A second example is shown in Figure~\ref{fig:app:prot_number:eagle}, 
    where we visualize the emerging prototypes for ImageNet class ``American eagle'' when varying the prototype number.
    Whereas one prototype visualizes eagles sitting on a branch or flying,
    we reveal that the model has learned to differentiate between flying eagles, sitting eagles and eagle heads by increasing prototype numbers.
        A third example is shown in Figure~\ref{fig:app:prot_number:spaceshuttle},
        where we visualize the emerging prototypes for ImageNet class ``space shuttle'' when varying the prototype number.
        By increasing prototype numbers,
        we reveal that the model has learned to differentiate between starting space ships (with dust clouds and fire), flying space ships and space ships in a halls.

    \subsection{Revealing Spurious Behavior}
    \label{app:spurious_behavior}

\begin{figure}[t]
            \centering
                \includegraphics[width=0.99\linewidth]{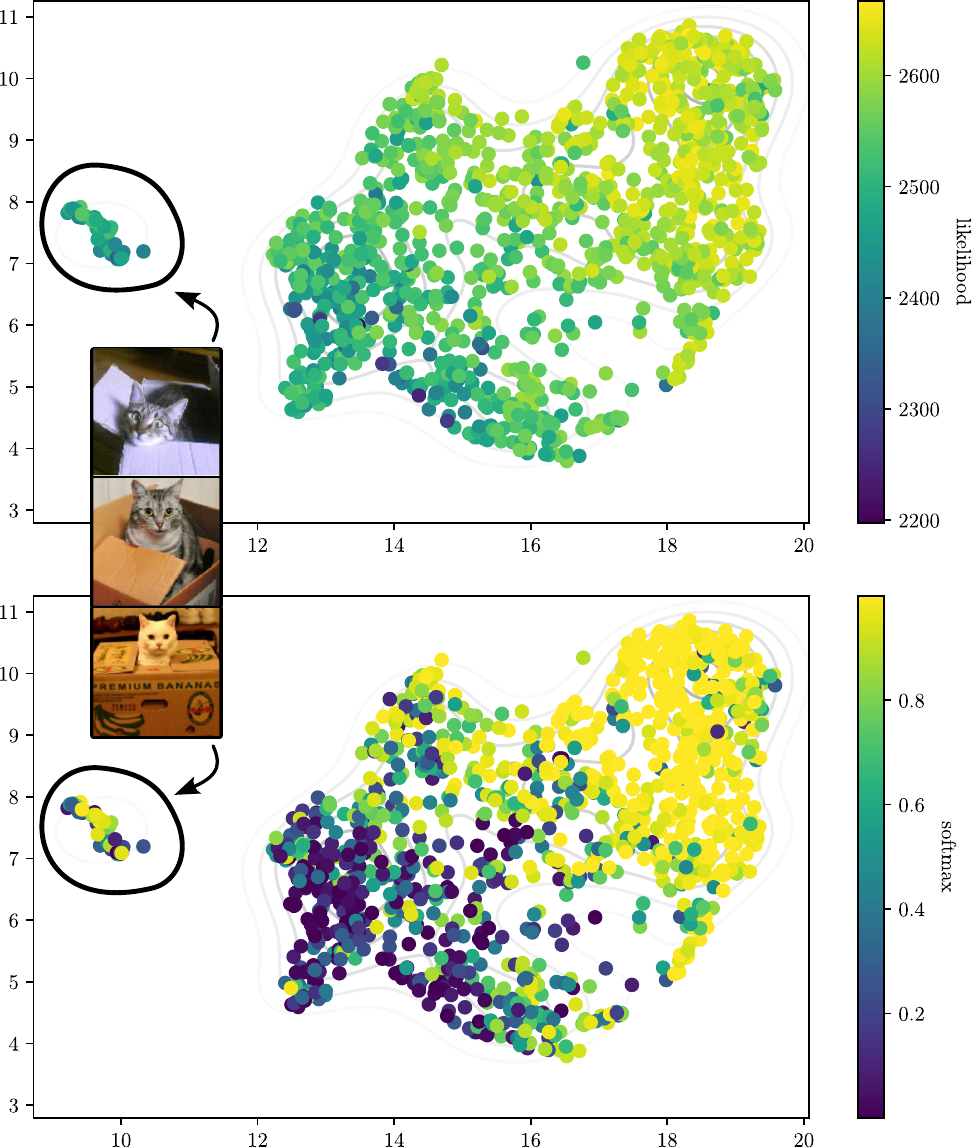}
                \captionof{figure}{
                UMAP embeddings of the ImageNet class ``carton'' using concept relevances of a VGG-16 model and layer \texttt{features.28}.
                (\emph{top}): For each prediction, we plot the class log-likelihood when modeling the distribution with one Gaussian.
                (\emph{bottom}): For each prediction, we plot the softmax value for the ``carton'' class.
                }
                \label{fig:app:spurious:carton_embedding}
        \end{figure}

    In Section~\ref{sec:exp:spurious}, we found a cluster for the ImageNet ``carton'' class, in which cats were depicted sitting in cartons.
    As shown in Figure~\ref{fig:app:spurious:carton_embedding},
    predictions in the cat cluster also have a high softmax probability score for the ``carton'' class.
    Thus, not only are cat features highly relevant (\wrt ``carton'') for these samples (as shown in Figure~\ref{fig:exp:sanity:use_case:spurious}),
    the ``carton'' output probability score is high as well. 
    
    \subsection{Out-of-Distribution Detection}
    \label{app:ood_detection}
    We present additional results for the \gls{ood} detection experiment in Section~\ref{sec:exp:outlier_detection}.
    Concretely,
    the results for all models trained on CIFAR-10 are shown in Table~\ref{tab:app:exp:ood:cifar} and for CUB-200 in Table~\ref{tab:app:exp:ood:imagenet}.
    Again we compare the methods of MSP~\cite{hendrycks2016baseline}, Energy~\cite{liu2020energy} and Mahalanobis~\cite{lee2018simple} with variants of \gls{ours} based on the log-likelihood measure as defined in Equation~\eqref{eq:methods:log_likelihood}, and Euclidean distance.
    For ImageNet,
    we perform \gls{ood} detection on the first 50 of the 1000 ImageNet classes.
    Note,
    that for \gls{ours},
    we only compute \glspl{gmm} with one prototype, and leave optimization \wrt prototype number and feature layer for future work.
        \begin{table*}[t] 
        \centering
        \caption{\gls{ood} detection results for (VGG$|$ResNet$|$EfficientNet) models trained on \textbf{CIFAR-10}. Higher \gls{auc} scores are better.}
        \begin{tabular}{@{\hspace{1em}}l@{\hspace{1em}}c@{\hspace{1em}}c@{\hspace{1em}}c@{\hspace{1em}}c@{\hspace{1em}}c@{\hspace{1em}}}
        \toprule
        {} &                 LSUN &                 iSUN &             Textures &                 SVHN &  Average \\
        \midrule
MSP~\cite{hendrycks2016baseline} & $90.9\,|\,83.4\,|\,86.0$ & $87.3\,|\,79.2\,|\,87.5$ & $90.9\,|\,81.9\,|\,83.1$ & $90.4\,|\,82.0\,|\,83.0$ & $85.5$ \\
Energy~\cite{liu2020energy} & $94.3\,|\,91.1\,|\,91.5$ & $91.2\,|\,84.2\,|\,92.3$ & $94.0\,|\,86.2\,|\,86.2$ & $91.4\,|\,85.7\,|\,85.9$ & $89.5$ \\
Mahalanobis~\cite{lee2018simple} & $49.0\,|\,43.0\,|\,76.3$ & $57.1\,|\,61.2\,|\,84.5$ & $81.5\,|\,55.7\,|\,71.5$ & $68.9\,|\,53.0\,|\,74.9$ & $64.7$ \\
\gls{ours}-E (ours) & $88.2\,|\,76.0\,|\,81.7$ & $81.7\,|\,75.9\,|\,88.4$ & $91.0\,|\,73.9\,|\,78.7$ & $89.4\,|\,65.7\,|\,80.4$ & $80.9$ \\
\gls{ours}-GMM (ours) & $94.3\,|\,83.7\,|\,86.1$ & $88.7\,|\,80.2\,|\,91.1$ & $95.8\,|\,81.6\,|\,83.0$ & $94.8\,|\,83.0\,|\,84.2$ & $87.2$ \\
        \bottomrule
        \end{tabular}
        \label{tab:app:exp:ood:cifar}
        \end{table*}
        \begin{table*}[t] 
        \centering
        \caption{\gls{ood} detection results for (VGG$|$ResNet$|$EfficientNet) models trained on \textbf{ImageNet}. 
        Higher \gls{auc} scores are better.}
        \begin{tabular}{@{\hspace{1em}}l@{\hspace{1em}}c@{\hspace{1em}}c@{\hspace{1em}}c@{\hspace{1em}}c@{\hspace{1em}}c@{\hspace{1em}}}
        \toprule
        {} &                 iSUN &                 Places365 &             Textures &                 CIFAR-10 &  Average \\
        \midrule

MSP~\cite{hendrycks2016baseline} & $96.3\,|\,93.2\,|\,99.1$ & $96.7\,|\,95.9\,|\,99.5$ & $96.0\,|\,95.3\,|\,99.0$ & $93.5\,|\,88.9\,|\,97.4$ & $95.9$ \\
Energy~\cite{liu2020energy} & $99.9\,|\,99.9\,|\,99.8$ & $99.8\,|\,99.8\,|\,99.8$ & $99.7\,|\,99.5\,|\,99.6$ & $99.6\,|\,99.3\,|\,98.6$ & $99.6$ \\
Mahalanobis~\cite{lee2018simple} & $38.8\,|\,98.7\,|\,99.6$ & $88.8\,|\,94.5\,|\,98.3$ & $86.3\,|\,90.5\,|\,99.2$ & $19.6\,|\,99.0\,|\,98.8$ & $84.3$ \\
\gls{ours}-E (ours) & $98.7\,|\,99.2\,|\,99.2$ & $98.3\,|\,98.1\,|\,98.3$ & $99.3\,|\,99.0\,|\,99.7$ & $99.3\,|\,98.8\,|\,98.0$ & $98.8$ \\
\gls{ours}-GMM (ours) & $98.3\,|\,98.9\,|\,99.5$ & $99.3\,|\,99.0\,|\,98.9$ & $99.7\,|\,99.3\,|\,99.8$ & $99.1\,|\,98.7\,|\,98.6$ & $99.1$ \\
        \bottomrule
        \end{tabular}
        \label{tab:app:exp:ood:imagenet}
        \end{table*}

    \subsection{Studying (Dis-)Similarities Across Classes}
    \label{app:exp:class_similaritites}

    In our experiments, prototypes emerge as tools for exploring both similarities and distinctions in concept utilization across various classes.

            \begin{figure}[t] 
            \centering
                \includegraphics[width=0.97\linewidth]{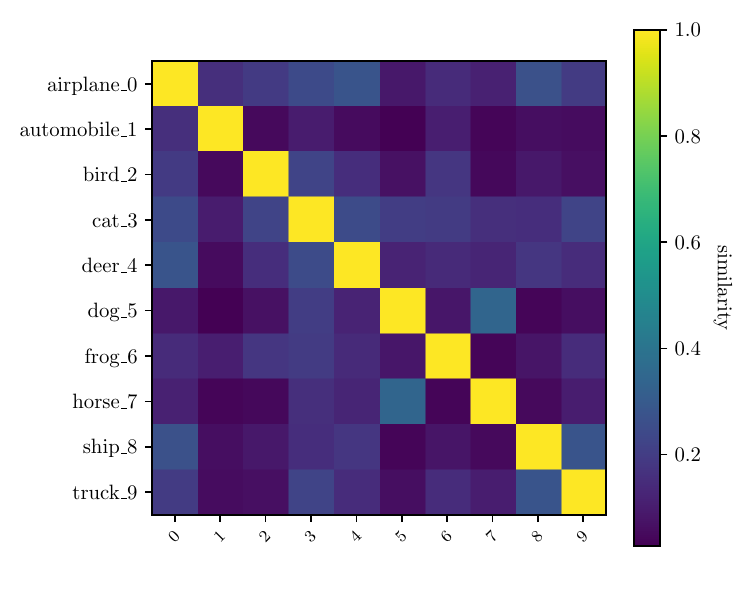}
                \captionof{figure}{
                Class similarity matrix between class prototypes (one per class) for the CIFAR-10 dataset and VGG-16 on layer \texttt{features.28}.
                }
                \label{fig:app:add_proto:class_similarity:cifar10_vgg}
        \end{figure}
                \begin{figure}[t] 
            \centering
                \includegraphics[width=0.97\linewidth]{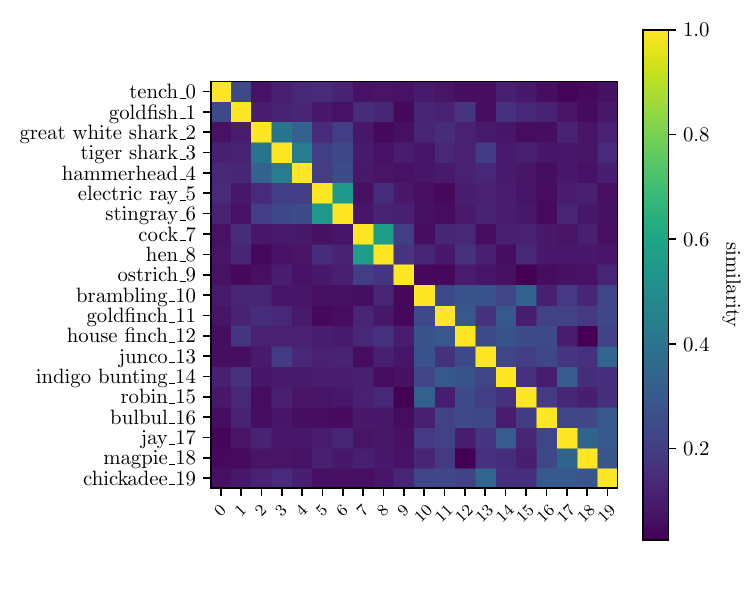}
                \captionof{figure}{
                Class similarity matrix between class prototypes (one per class) for the ImageNet dataset (first 20 classes) and EfficientNet-B0 on the last convolutional \texttt{features} block.
                }
                \label{fig:app:add_proto:class_similarity:imagenet_efficient}
        \end{figure}
                \begin{figure}[t] 
            \centering
                \includegraphics[width=0.97\linewidth]{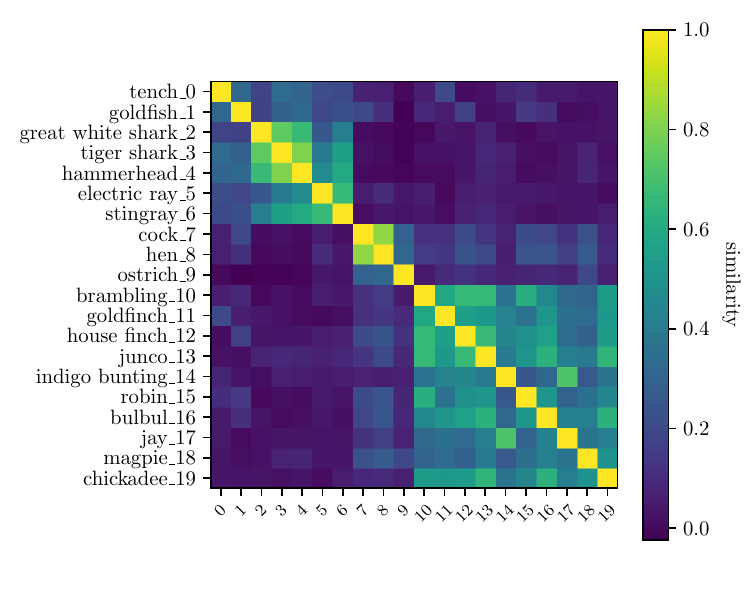}
                \captionof{figure}{
                Class similarity matrix between class prototypes (one per class) for the ImageNet dataset (first 20 classes) and ResNet-18 on the last convolutional \texttt{BasicBlock}.
                }
                \label{fig:app:add_proto:class_similarity:imagenet_resnet}
        \end{figure}
            \begin{figure*}[t] 
            \centering
                \includegraphics[width=0.99\linewidth]{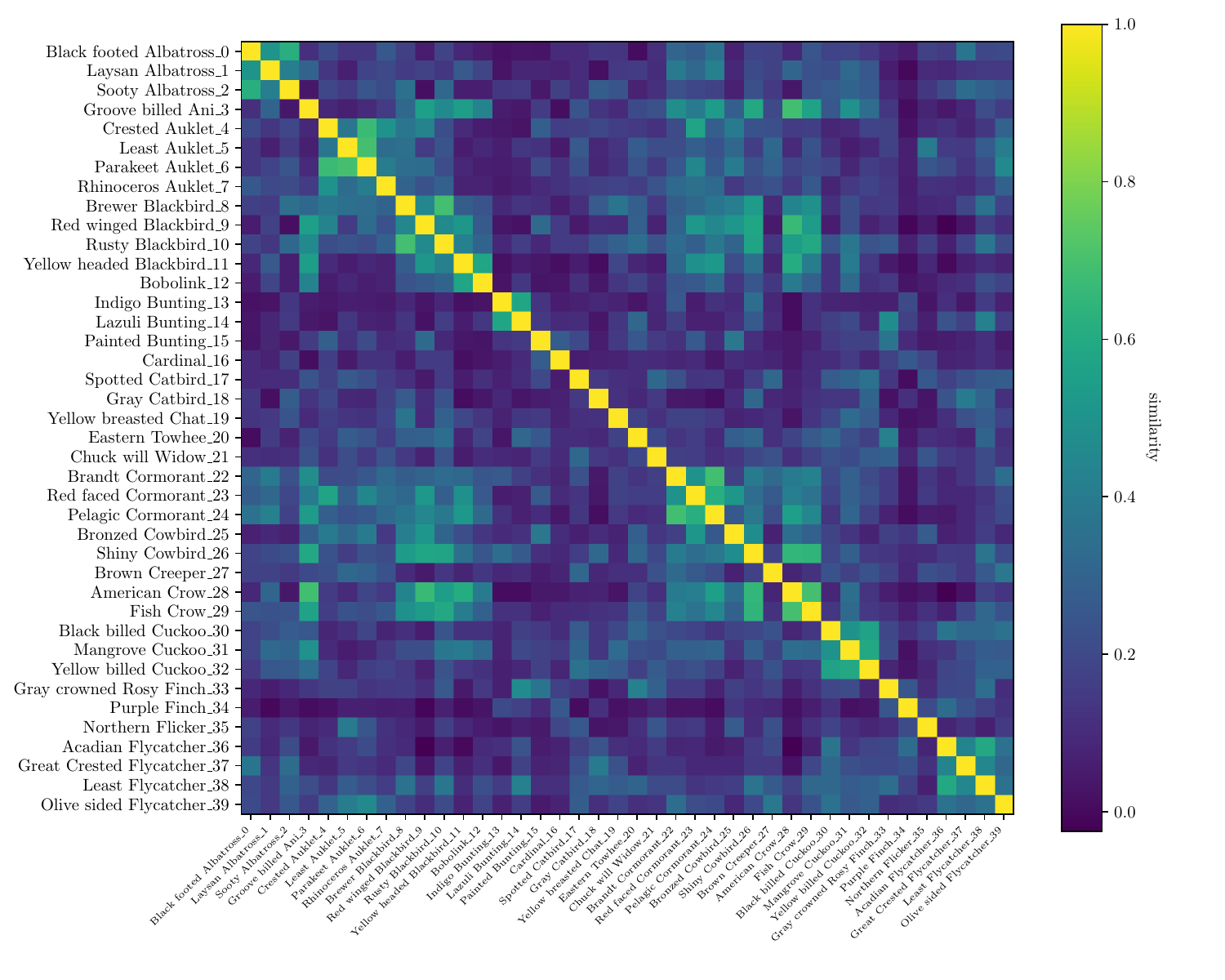}
                \captionof{figure}{
                Class similarity matrix between class prototypes (one per class) for the CUB-200 dataset (first 40 classes) and VGG-16 on layer \texttt{features.28}.
                }
                \label{fig:app:add_proto:class_similarity:cub_vgg}
        \end{figure*}
    
    For a global understanding of the model,
    we present matrices depicting inter-class similarities (using cosine similarity between prototypes) for a VGG-16 trained on CUB-200 (Figure~\ref{fig:app:add_proto:class_similarity:cub_vgg}), CIFAR-10 (Figure~\ref{fig:app:add_proto:class_similarity:cifar10_vgg}), ResNet on ImageNet (Figure~\ref{fig:app:add_proto:class_similarity:imagenet_resnet}) and EfficientNet on ImageNet (Figure~\ref{fig:app:add_proto:class_similarity:imagenet_efficient}).

        \begin{figure}[t]
            \centering
                \includegraphics[width=0.99\linewidth]{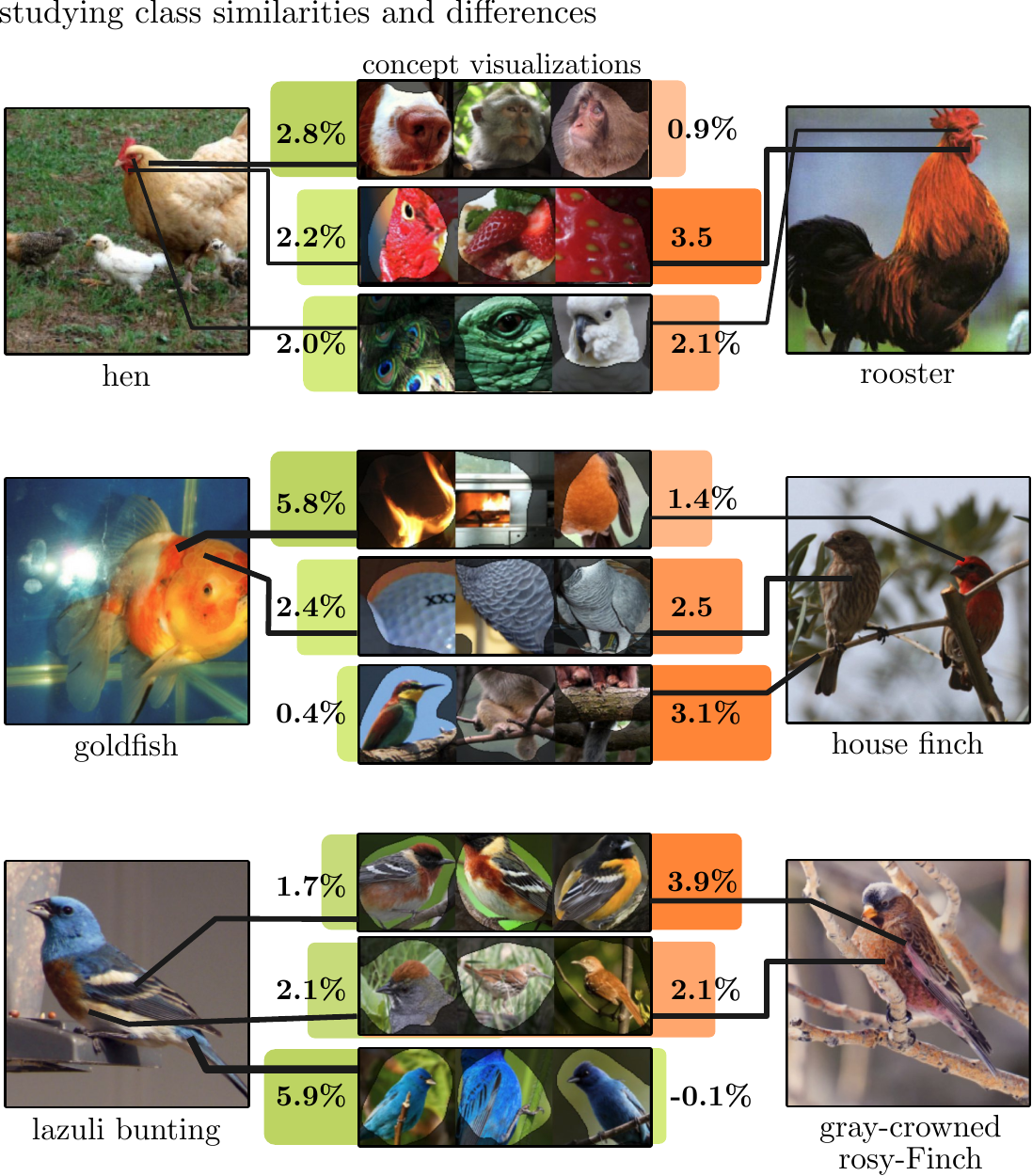}
                \captionof{figure}{Studying class (dis-)similarities across classes in terms of concepts found in layer \texttt{features.28} for a VGG-16 model.
                (\emph{top}): ImageNet ``hen'' compared to ``rooster'' prototype.
                Whereas the ``eye'' concept is used for both equally, the ``red color'' concept is more important for the rooster class.
                For the hen, the brown-gray color is more important.
                (\emph{middle}): ImageNet ``goldfish'' compared to ``House Finch'' prototype.
                Whereas for both, the ``scales'' texture is important, 
                the orange color is much more important for the goldfish. On the other side,
                the ``animal on branch'' concept is only used for the House Finch.
                (\emph{bottom}): CUB-200 ``Lazuli Bunting'' compared to ``Gray-crowned Rosy-Finch'' prototype.
                Wheras for both ``brown color'' is important, ``black-white wings with brown color'' is more relevant for the Finch.
                A ``blue color'' concept is only relevant for the Lazuli Bunting.
                }
                \label{fig:app:class_similarities}
        \end{figure} 
    
Moreover, we delve into detailed analyses of layer \texttt{features.28} for a VGG-16 model. 
Here, we illustrate the similarities and differences between classes by choosing three characteristic concepts.
For instance,
when comparing the ImageNet classes ``hen'' and ``rooster'' as in Figure~\ref{fig:app:class_similarities} (\emph{top}),
we observe an equal reliance on the ``eye'' concept.
However, the rooster places greater importance on the ``red color'' compared to the hen,
where the brown-gray color is more significant.

Similarly, when comparing between the ImageNet classes ``goldfish'' and ``House Finch'' as shown in Figure~\ref{fig:app:class_similarities} (\emph{middle}), both exhibit a reliance on the ``scales'' texture.
However, the goldfish emphasizes the orange color, while the House Finch uniquely employs the ``animal on branch'' concept.

In another instance, when comparing CUB-200 classes ``Lazuli Bunting'' and ``Gray-crowned Rosy-Finch'', both prioritize ``brown color''.
However, the Finch places greater importance on ``black-white wings with brown color'',
whereas a ``blue color'' concept plays a significant role (only) for the Lazuli Bunting.

    \subsection{Comparing Single Predictions with Prototypes}
    \label{sec:app:comparing_with_prototype}
    As detailed in Section~\ref{sec:exp:prediction_validation} and Figure~\ref{fig:introduction:local_to_glocal},
    we can compare a single prediction and its explanation with a prototype to understand how ordinary a prediction is in a more objective manner. 
    Specifically,
    we can compare differences in how and which concepts are used.
    These differences can be quantitatively measured via the log-likelihood as in Equation~\eqref{eq:methods:log_likelihood},
    enabling for an automatic detection of outliers, or to assign predictions to the closest prototypical prediction strategy.

\begin{figure*}[t]
            \centering
                \includegraphics[width=0.98\linewidth]{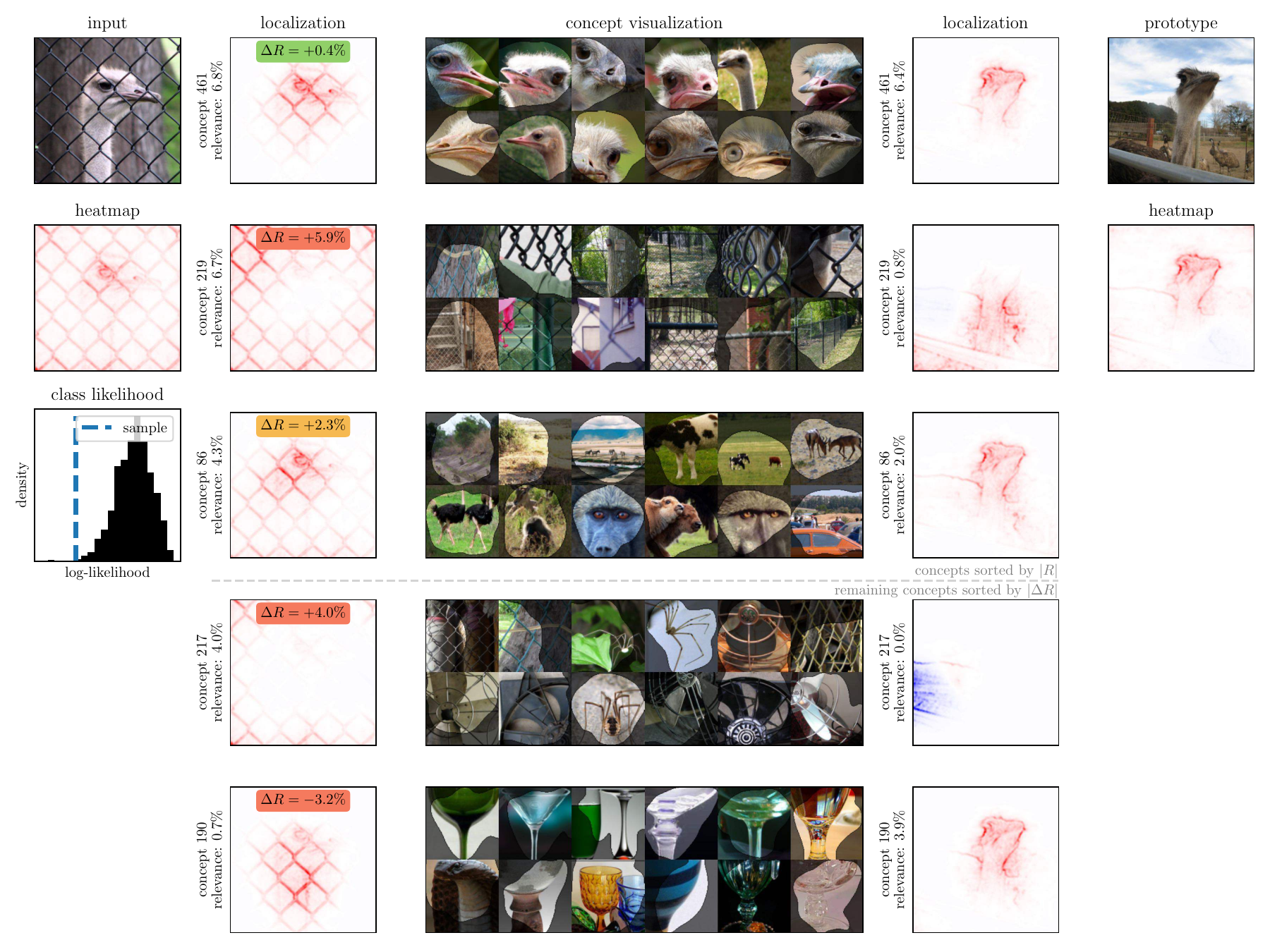}
                \captionof{figure}{
                Comparing a single prediction (\emph{left}) with a prototype (\emph{right}) allows to validate predictions in more objective manner.
                Here, an ostrich (ImageNet class ``ostrich'') is correctly predicted by a VGG-16 model,
                but is detected as an outlier based on the likelihood measure as defined in Equation~\eqref{eq:methods:log_likelihood} (illustrated in the plot in the first column and third row).
                In this example, we compare against the most similar prototype (of six) based on concept relevance scores $R$ from layer \texttt{features.28}.
                By comparing the used concepts,
                we can understand that the model strongly uses fence features for the outlier prediction, which are not apparent for the prototype.
                Compared to the prototype,
                concepts regarding the head's shape are missing.
                }
                \label{fig:app:local_validation:ostrich}
        \end{figure*}

\begin{figure*}[t]
            \centering
                \includegraphics[width=0.98\linewidth]{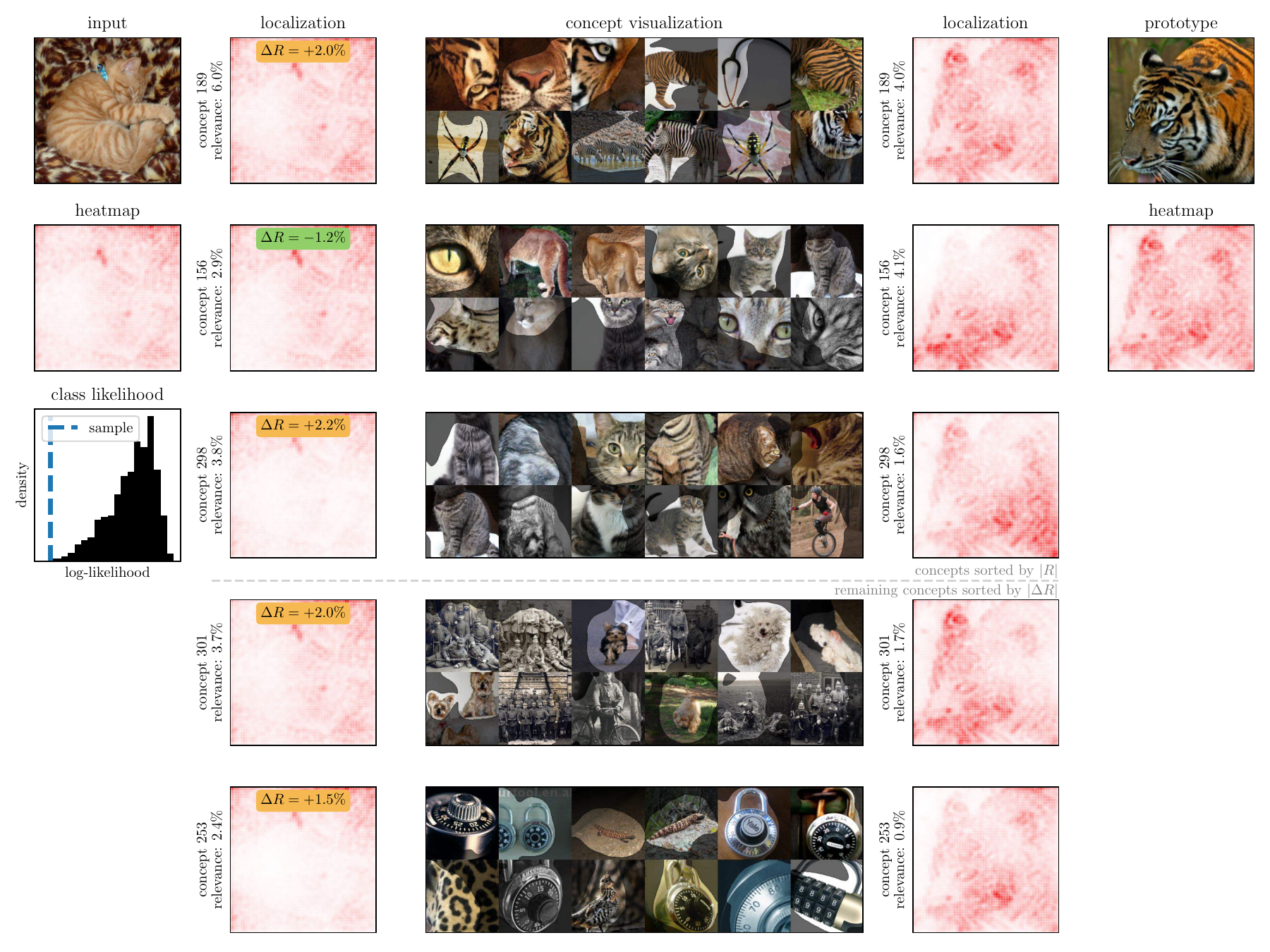}
                \captionof{figure}{
                Comparing a single prediction (\emph{left}) with a prototype (\emph{right}) allows to validate predictions in more objective manner.
                Here, a Tiger Cat (ImageNet class ``Tiger Cat'') is correctly predicted by a ResNet-18 model,
                but is detected as an outlier based on the likelihood measure as defined in Equation~\eqref{eq:methods:log_likelihood} (illustrated in the plot in the first column and third row).
                In this example, we compare against the most similar prototype (of six) based on concept relevance scores $R$ from the last \texttt{BasicBlock}.
                The test sample shows an orange Tiger Cat lying on a blanket with jaguar fur pattern.
                Interestingly,
                the most similar prototype is not a Tiger Cat, but a tiger, as already observed in Figure~\ref{fig:app:add_proto:data_quality:wrong_labels:lacewing_tiger_cat} (\emph{bottom}) due to a high number of tiger samples in the training dataset with class label ``Tiger Cat''.
                By comparing the used concepts,
                we can understand that the sample and prototype are similar in a ``stripes-texture'' concept.
                However, the sample deviates in a stronger use of cat- and jaguar-like concepts (concepts 298 and 253).
                Underrepresented are concepts associated with cat eyes and heads (concept 156), which is sensible, as the eyes of the cat in the test prediction are closed.
                }
                \label{fig:app:local_validation:cat}
        \end{figure*}

\begin{figure*}[t]
            \centering
                \includegraphics[width=0.98\linewidth]{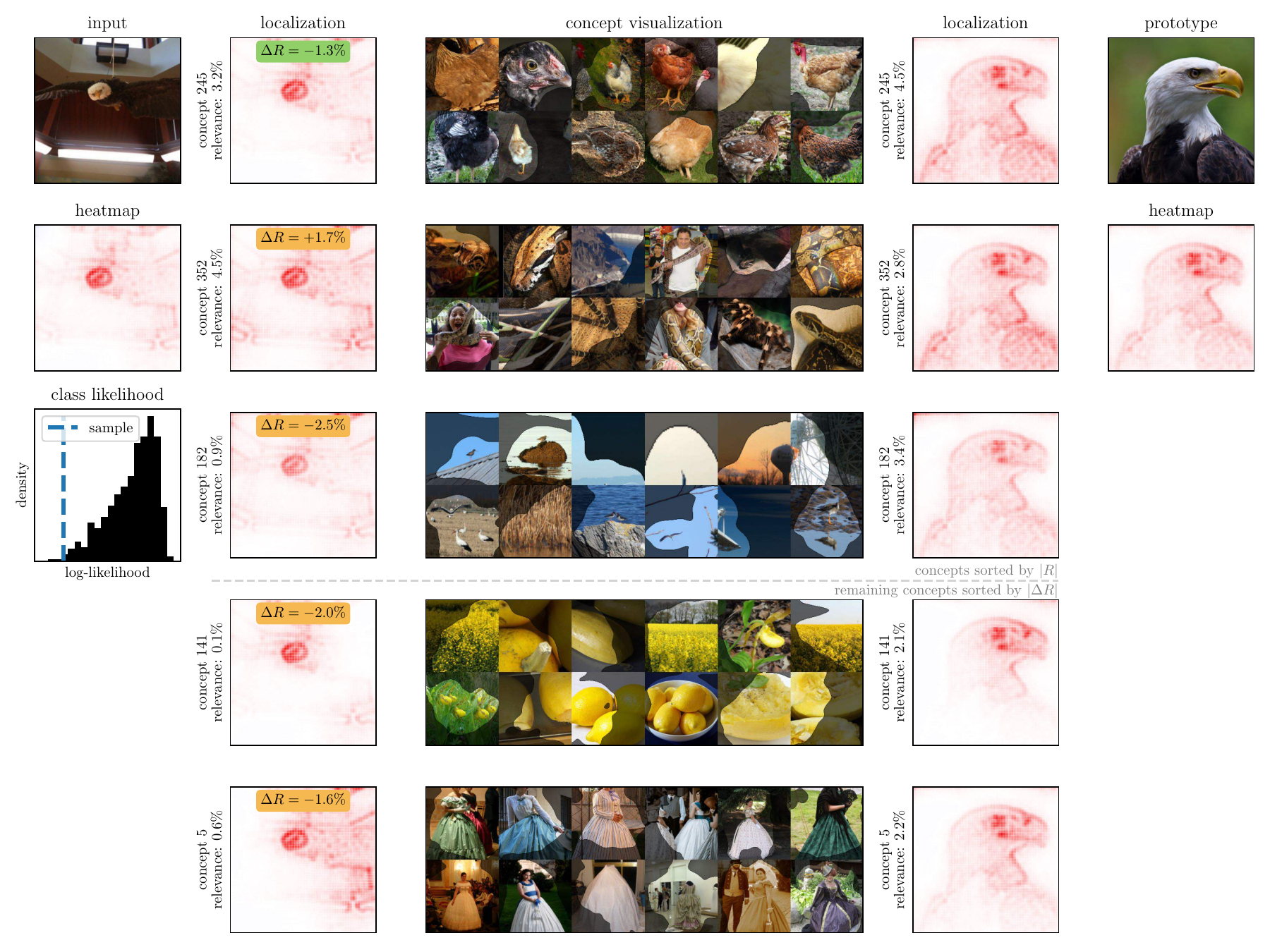}
                \captionof{figure}{
                Comparing a single prediction (\emph{left}) with a prototype (\emph{right}) allows to validate predictions in more objective manner.
                Here, a bald eagle (ImageNet class ``bald eagle'') is correctly predicted by a ResNet-18 model,
                but is detected as an outlier based on the likelihood measure as defined in Equation~\eqref{eq:methods:log_likelihood} (illustrated in the plot in the first column and third row).
                In this example, we compare against the most similar prototype (of six) based on concept relevance scores $R$ from the last \texttt{BasicBlock}.
                Whereas concepts related to the feathers (concepts 245 and 352) are relevant for both,
                concepts related to the environment (concept 182) or yellow beak (concept 141) are missing.
                }
                \label{fig:app:local_validation:eagle}
        \end{figure*}

\begin{figure*}[t]
            \centering
                \includegraphics[width=0.98\linewidth]{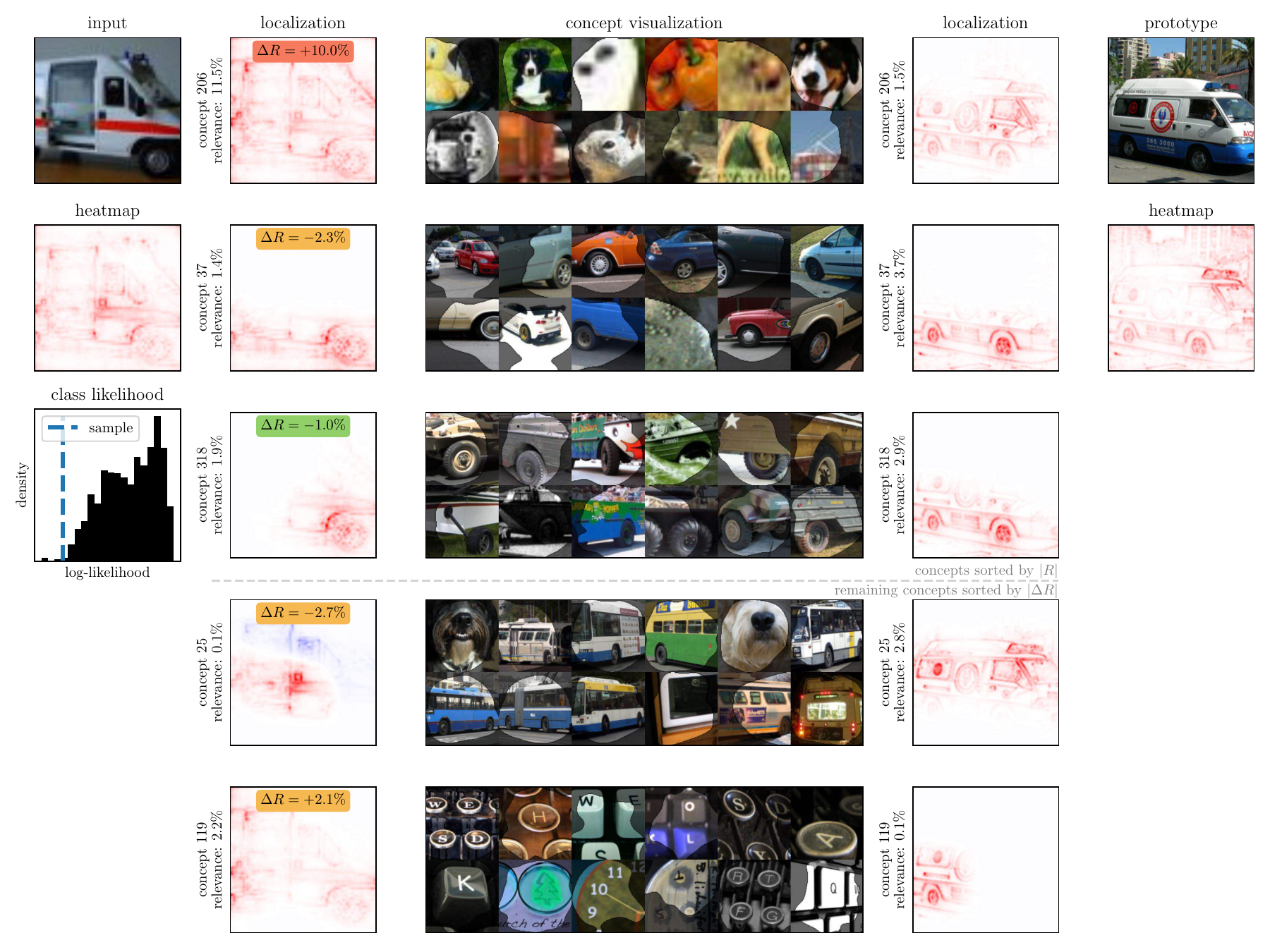}
                \captionof{figure}{
                Comparing a single prediction (\emph{left}) with a prototype (\emph{right}) allows to validate predictions in more objective manner.
                Here, an ambulance car (ImageNet class ``ambulance'') is correctly predicted by a VGG-16 model,
                but is detected as an outlier based on the likelihood measure as defined in Equation~\eqref{eq:methods:log_likelihood} (illustrated in the plot in the first column and third row).
                In this example, we compare against the most similar prototype (of six) based on concept relevance scores $R$ from layer \texttt{features.28}.
                By comparing the used concepts,
                we can understand that the model strongly uses features related to blur (concept 206) for the outlier prediction, which are not apparent for the prototype.
                Compared to the prototype,
                concepts regarding side of buses with black windows (concept 25) are missing.
                }
                \label{fig:app:local_validation:ambulance}
        \end{figure*}

    We present examples for validating single predictions using prototypes with samples from the ImageNet dataset in Figures~\ref{fig:app:local_validation:ostrich},~\ref{fig:app:local_validation:cat},~\ref{fig:app:local_validation:eagle} and~\ref{fig:app:local_validation:ambulance}.
    Here, 
    we show correct predictions that have a low log-likelihood compared to samples of the same class in the training set, as illustrated in the plot in the first column and third row of each figure.
    Each sample is compared to the most similar prototype (of six in total).
    We further show three concepts that are most relevant for either prediction or prototype,
    followed by two concepts where the differences in terms of concept relevance scores between sample and prototype are largest.

    In the first example in Figure~\ref{fig:app:local_validation:ostrich},
    an ostrich is correctly predicted by a VGG-16 model.
    The ostrich's head and part of the neck is visible, mimicking the prototype.
    However,
    the test prediction differs in a fence, behind which the ostrich is depicted.
    By studying the concept relevance scores,
    we can understand that the fence was highly relevant for the prediction,
    which is not the case for the prototype.
    Further, concepts regarding the head's shape are missing.
    
    A second example is shown in Figure~\ref{fig:app:local_validation:cat},
    where a Tiger Cat is correctly predicted by a ResNet-18 model.
    Here, the test sample shows an orange Tiger Cat lying (sleeping) on a blanket with jaguar fur pattern.
    Notably,
    the most similar prototype is not a Tiger Cat, but a tiger, as already observed in Figure~\ref{fig:app:add_proto:data_quality:wrong_labels:lacewing_tiger_cat} (\emph{bottom}) due to a high number of tiger samples in the training dataset with class label ``Tiger Cat''.
    For both sample and prototype,
    a ``stripes-texture'' concept is relevant,
    however,
    the test sample deviates in a stronger use of cat- and jaguar-like concepts (concepts 298 and 253).
    On the other side,
    underrepresented are concepts associated with cat eyes and heads (concept 156), 
    which can be expected, as the eyes of the cat in the test prediction are closed.

        The third example is shown in Figure~\ref{fig:app:local_validation:eagle},
        where a bald eagle is correctly predicted by a ResNet-18 model.
        Here the eagle is depicted as an exhibit hanging in a room,
        whereas the prototype shows a close-up picture of an eagle's head.
        While concepts related to the feathers (concepts 245 and 352) are relevant for both,
        concepts related to the environment (concept 182) or yellow beak (concept 141) are missing.
        
        A last example is shown in Figure~\ref{fig:app:local_validation:ambulance},
        where an ambulance is correctly predicted by a VGG-16 model.
        Here,
        the test input has very low resolution and shows an ambulance car from the side.
        The prototype shows an ambilance car from the side as well, but in higher resolution.
        By comparing the used concepts,
        we can understand that the model strongly uses features related to blur (concept 206) for the outlier prediction, which are not apparent for the prototype.
        Further, compared to the prototype,
        concepts regarding side of buses with black windows (concept 25) are missing.
        
\end{document}